%% file: ZIG_main.tex
\documentclass[12pt, article
]{imsart}

\RequirePackage[OT1]{fontenc}
\RequirePackage{amsthm,amsmath} %,natbib
\usepackage{natbib}
\RequirePackage[colorlinks,citecolor=blue,urlcolor=blue]{hyperref}
\hypersetup{
    linktocpage=true,
    colorlinks=true,
    bookmarks=true,
    citecolor=blue,
    urlcolor=blue,
    linkcolor=blue,
    citebordercolor={1 0 0},
    urlbordercolor={1 0 0},
    linkbordercolor={.7 .8 .8},
    breaklinks=true,
    pdfpagelabels=true,
    }
\RequirePackage{hypernat}
\usepackage{amssymb}
\usepackage{multirow}
\usepackage{capt-of}
\usepackage{amsmath}
\usepackage{algorithm}
\usepackage[noend]{algpseudocode}
\usepackage{graphicx}
\usepackage{lineno}
\graphicspath{ {./figures/} }

\usepackage{booktabs,caption,fixltx2e}
\usepackage[flushleft]{threeparttable}
\usepackage{subcaption}
\usepackage{relsize}

\usepackage[compact]{titlesec}
\titlespacing{\section}{1pt}{*0.5}{*0.1}
\titlespacing{\subsection}{0.7pt}{*0.3}{*0.1}
\titlespacing{\subsubsection}{0pt}{*0}{*0}

\usepackage[dvipsnames]{xcolor}

\newcommand{\modify }{\color{Black}  }  %violet
\newcommand{\revise}{\color{Black}}
\makeatletter
\def\BState{\State\hskip-\ALG@thistlm}
\makeatother

\input{mac}
\begin{document}
	
\thispagestyle{empty}		
	\begin{frontmatter}
		\title{Matrix factorization and prediction for high
dimensional co-occurrence count data via shared parameter alternating zero inflated Gamma model % in cancer classification
}
		%Title on one or two lines
		%\newline without capitals, except after a colon
		%}} }
		
		\vglue 5mm

		\begin{aug}
			%\author{\fnms{Xukun} \snm{Li}\ead[label=e1]{ xukun@ksu.edu}}
			%,
			\author{\fnms{Taejoon } \snm{Kim$^1$}\ead[label=e1]{taejoon.kim@csueastbay.edu}}
and
\author{\fnms{Haiyan} \snm{Wang$^{2a}$}\corref{} \ead[label=e2]{hwang@ksu.edu}}
%and
%\author{\fnms{Santosh} \snm{Ghimire$^3$} \ead[label=e3]{santoshghimire@ioe.edu.np} }
			\affiliation{{\small Department of Statistics, Kansas State
					University, 101 Dickens Hall, Manhattan, KS 66506}
				%\printead{e2}
			}

			\address{
               $^1$  Department of Statistics and Biostatistics, California State University East Bay\\
               $^2$ Department of Statistics, Kansas State University\\
              % $^3$ Department of Applied Sciences and Chemical Engineering, Pulchowk Campus, Tribhuvan University
				% }
			}

			\runauthor{Kim \& Wang
}
		\end{aug}
		\footnotetext{\textit{$^*$ Running head: Alternating ZIG for Matrix factorization and prediction}}
		
		\footnotetext{\textit{$^{a}$ Author to whom correspondence may be
				addressed. Email: hwang@ksu.edu}}
		%\medskip
		%\footnotetext{\textit{$^a$ Affiliation   }}
		
		\runtitle{Matrix factorization and prediction using alternating ZIG}

\setcounter{page}{1}

\vspace{0.2in}
		\begin{abstract}
{\modify
High-dimensional sparse matrix data frequently arise in various applications. A notable example is the weighted word-word co-occurrence count data, which summarizes the weighted frequency of word pairs appearing within the same context window. This type of data typically contains highly skewed non-negative values with an abundance of zeros. Another example is the co-occurrence of item-item or user-item pairs in e-commerce, which also generates high-dimensional data. The objective is to utilize this data to predict the relevance between items or users. In this paper, we assume that items or users can be represented by unknown dense vectors. The model treats the co-occurrence counts as arising from zero-inflated Gamma random variables and employs cosine similarity between the unknown vectors to summarize item-item relevance. The unknown values are estimated using the shared parameter alternating zero-inflated Gamma regression models (SA-ZIG). Both canonical link and log link models are considered. Two parameter updating schemes are proposed, along with an algorithm to estimate the unknown parameters. Convergence analysis is presented analytically. Numerical studies demonstrate that the SA-ZIG using Fisher scoring without learning rate adjustment may fail to find the maximum likelihood estimate. However, the SA-ZIG with learning rate adjustment performs satisfactorily in our simulation studies.}

\end{abstract}

	\end{frontmatter}
	
\input{chapter3.tex}

\bibliographystyle{chicago} %abbrvnat
\bibliography{references}

\end{document}

%% file: mac.tex
\makeatletter
\@addtoreset{equation}{section}

\makeatother

\relax

\newcommand{\vb}{{\bf b}}
\newcommand{\vc}{{\bf c}}

\newcommand{\vt}{{\bf t}}

\newcommand{\vw}{{\bf w}}
\newcommand{\vx}{{\bf x}}
\newcommand{\vy}{{\bf y}}

\newcommand{\vU}{{\bf U}}

\newcommand{\vY}{{\bf Y}}

\newcommand{\wtb}{{\widetilde b}}
\newcommand{\wtvb}{\mbox{\boldmath $\wtb$}}

\newcommand{\wte}{{\widetilde e}}

\newcommand{\wtw}{{\widetilde w}}
\newcommand{\wtvw}{\mbox{\boldmath $\wtw$}}

\newcommand{\wttheta}{{\widetilde \theta}}
\newcommand{\wtvtheta}{\mbox{\boldmath $\wttheta$}}

\newcommand{\vtheta}{\mbox{\boldmath $\theta$}}

\newcommand{\bay}{\begin{array}}
\newcommand{\eay}{\end{array}}

\newcommand{\Var}{\mbox{Var}}

\newcommand{\bqa}{\begin{eqnarray*}}
\newcommand{\eqa}{\end{eqnarray*}}
\newcommand{\bqan}{\begin{eqnarray}}
\newcommand{\eqan}{\end{eqnarray}}
\newcommand{\bqt}{\begin{quote}}
\newcommand{\eqt}{\end{quote}}
\newcommand{\bt}{\begin{tabbing}}
\newcommand{\et}{\end{tabbing}}
\newcommand{\bit}{\begin{itemize}}
\newcommand{\eit}{\end{itemize}}
\newcommand{\ben}{\begin{enumerate}}
\newcommand{\een}{\end{enumerate}}
\newcommand{\beq}{\begin{equation}}
\newcommand{\eeq}{\end{equation}}
\newcommand{\bdefi}{\begin{definition}}
\newcommand{\edefi}{\end{definition}}
\newcommand{\bpro}{\begin{proposition}}
\newcommand{\epro}{\end{proposition}}
\newcommand{\blem}{\begin{lemma}}
\newcommand{\elem}{\end{lemma}}
\newcommand{\bth}{\begin{theorem}}
\newcommand{\bco}{\begin{corollary}}
\newcommand{\eco}{\end{corollary}}
\newcommand{\bdes}{\begin{description}}
\newcommand{\edes}{\end{description}}

\newtheorem{definition}{Definition}[section]
\newtheorem{theorem}[definition]{Theorem}

%% file: chapter3.tex
% +--------------------------------------------------------------------+
% | Sample Chapter 2
% +--------------------------------------------------------------------+

%\cleardoublepage

% +--------------------------------------------------------------------+
% | Replace "This is Chapter 2" below with the title of your chapter.
% | LaTeX will automatically number the chapters.
% +--------------------------------------------------------------------+
%  keywords={Matrix decomposition;Vectors;Optimization;Algorithm design and analysis;Data analysis;Signal processing algorithms;Semantics;Data mining;dimensionality reduction;multivariate data analysis;nonnegative matrix factorization (NMF)},

\section{Introduction}
{\modify

Matrix factorization is a fundamental technique in linear algebra and data science, widely used for dimensionality reduction, data compression, and feature extraction. Recent research expands its use in various fields, including recommendation systems (e.g., collaborative filtering), bioinformatics, and signal processing. Researchers are actively pursuing new types of factorizations.
In their effort to discover new factorizations and provide a unifying structure, \cite{ref6}
list 53 systematically derived matrix factorizations arising from the generalized Cartan decomposition. Their results apply to invertible matrices and generalizations of orthogonal matrices in classical Lie groups.

Nonnegative matrix factorization (NMF) is particularly useful when dealing with non-negative data, such as in image processing and text mining. \cite{ref5} survey existing NMF methods and their variants, analyzing their properties and applications. \cite{saberimovahed2024} present a comprehensive survey of NMF, focusing on its applications in feature extraction and feature selection. \cite{Wang_zhang2013} summarize theoretical research on NMF from 2008 to 2013, categorizing it into four types and analyzing the principles, basic models, properties, algorithms, and their extensions and generalizations.

There are many advances aimed at developing more efficient algorithms tailored to specific applications. One prominent application of matrix factorization is in recommendation systems, particularly for addressing the cold-start problem. In recommender systems, matrix factorization models decompose user-item, user-user, or item-item interaction matrices into lower-dimensional latent spaces, which can then be used to generate recommendations.

\cite{Koren} summarizes the literature on recommender systems and proposes a multifaceted collaborative filtering model that integrates both neighborhood and latent factor approaches. \cite{Zhang_Liu2015} develop a matrix factorization model to generate recommendations for users in a social network. \cite{Fernandez-Tobias-et_al2019} proposes a matrix factorization model for cross-domain recommender systems by extracting items from three different domains and finding item similarities between these domains to improve item ranking accuracy.

\cite{Puthiya_et_al2020} develop a matrix factorization model that combines user-item rating matrices and item-side information matrices to develop soft clusters of items for generating recommendations. \cite{Nguyen_et_al2016} propose a matrix factorization that learns latent representations of both users and items using gradient-boosted trees. \cite{Panda_Ray2022} provide a systematic literature review on approaches and algorithms to mitigate cold-start problems in recommender systems.

%
%There are many advances aiming to develope more efficient algorithms tailored to specific applications.
%One of the prominent applications of matrix factorization is in recommendation systems, particularly to address the cold-start problem.
%  In recommender systems, matrix factorization models decompose user-item, user-user, or item-item interaction
%matrix into lower-dimensional latent space, which can then be used to generate
%recommendations. \cite{Koren} summarized the literature on recommender systems and proposed a multifaceted collaborative filtering model that integrates both neighborhood and latent factor approaches.
%\cite{Zhang_Liu2015} developed a matrix factorization model to generate recommendations for users in a social network.
%\cite{Fernandez-Tobias-et_al2019} proposed a matrix factorization model for cross-domain
%recommender system by extracting the items from three different
%domains and find item similarities between these domains to improve the item ranking accuracy.
%\cite{Puthiya_et_al2020} developed a matrix factorization model that combines
%user-item rating matrix and item-side information matrix to develop soft clusters of items
%for generating recommendations.  \cite{Nguyen_et_al2016} proposed a matrix factorization that learns latent representations of
%both users and items with gradient boosted trees. \cite{Panda_Ray2022} provided a systematic
%literature review on approaches and algorithms to mitigate cold start
%problems in recommender systems.

Matrix factorization is also used in natural language processing (NLP) in recent years. Word2Vec by \citeauthor{mikolov2013A} (\citeyear{mikolov2013A}a, \citeyear{mikolov2013B}b) marks a milestone in NLP history. Although no clear matrices are presented in their study, Word2Vec models the co-occurrence of words and phrases using latent vector representations via a shallow neural network model. Another well-known example of matrix factorization in NLP is the word representation with global vectors (GloVe) by \cite{pennington2014glove}. They model the co-occurrence count matrix of words using cosine similarity of latent vector representations via alternating least squares.
%least squares principle for optimization.
However, practical data may be heavily skewed, making the sum or mean squared error unsuitable as an objective function. To address this, \cite{pennington2014glove} utilize weighted least squares to reduce the impact of skewness in the data.  Manually creating the weight is difficult to work well for real data. In GloVe training, the algorithm is set to run a fixed number of iterations without other convergence check. Here, we consider using the likelihood principle to model the skewed data matrix.
}

%Matrix factorization is also used in natural language processing (NLP) in recent years. Word2Vec by
%\citeauthor{mikolov2013A} (2013a) and \citeauthor{mikolov2013B} (2013b) is a milestone in NLP history. Even though no clear matrices are presented in their study, Word2Vec models the co-occurrence of words and phrases using latent vector representations via a shallow neural network model.
%Word representation with global vectors by \cite{pennington2014glove} is another well-known example of matrix factorization developed for NLP. The co-occurrence count matrix of words was modeled by cosine similarity of latent vector representations via alternating least squares.
%}
%These studies primarily use least squares principle to achieve optimization. The practical data may be heavily skewed, which makes sum or mean squared error unsuitable for objective function. \cite{pennington2014glove} utilized weighted least squares to reduce the impact of skewness in the data. Here we would like to consider using the likelihood principle to model the skewed data matrix.

In this paper, our goal is to model {\modify non-negative} continuous {\modify sparse matrix} data from skewed distributions with an abundance of zeros but {\modify lacking} covariate information. We are particularly interested in zero-inflated Gamma observations, {\modify often referred to} as semi-continuous data {\modify due to the presence of many zeros and the highly skewed distribution of positive observations.} Examples of such data include insurance claim data, household expenditure data, and precipitation data (see
\cite{Mills:2013}).
\cite{Wei-etal:2020} {\modify study and simulate} data using actual deconvolved calcium imaging data, {\modify employing a zero-inflated Gamma model} to accommodate spikes of observed inactivity and traces of calcium signals in neural populations. \cite{Nobre-etal:2017} {\modify examine the} amount of leisure time spent on physical activity and explanatory variables such as gender, age, education level, and annual per capita family income. {\modify They find that the zero-inflated Gamma model is preferred over the multinomial model. Beyond Gamma distribution, Weibull distribution can also model skewed non-negative data. Unfortunately, when shape parameter is unknown, Weibull distribution is not a member of the exponential family. Further, the alternating update is not suitable because its sufficient statistics is not linear in the observed data. Different estimation procedure will need to be developed if Weibull distribution is used. The Gamma distribution not only is a member of the exponential family, its sufficient statistic is linear in the observed data. This gives solid theoretical ground for alternating update to find a solution. Section \ref{section:convergence} explains more details in this regard. }

%In this paper, our goal is to model continuous data from skewed distribution with abundance of zeros but lack of covariate information. We are particularly interested in zero inflated Gamma observations, which are sometimes referred as semi-continuous data because there are a lot of zeros and the distribution of non-zero observations are highly skewed continuous data.
%Examples of such data include insurance claim data, household expenditure data, precipitation data, etc. (see \cite{Mills:2013}). \cite{Wei-etal:2020} studied and simulated data using the actual deconvolved calcium imaging data using zero inflated Gamma to accommodate spike of observed inactivity and trace of calcium signals in neural populations. \cite{Nobre-etal:2017} studied amount of leisure time spent on physical activity and the explanatory variables including gender, age, education level, and annual per capita family income. The zero inflated Gamma model was preferred over multinomial model.

In all the aforementioned examples, there are explicitly observed covariates or factors. Furthermore, the two parts of the model parameters are
pendicular to each other, {\modify allowing model estimation by fitting} two separate regressions: binomial regression and Gamma regression. {\modify Unfortunately, such models} cannot be applied to user-item or co-occurrence count matrix data {\modify arising} from many practical applications. Examples include user-item or item-item co-occurrence data from online shopping platforms and co-occurring word-word pairs in sequences of texts.
One reason {\modify is the absence of observed covariates}. Additionally, the mechanism that leads to the observed data entries in the data matrix for the binomial and Gamma parts may be of the same nature, {\modify making it} more appropriate to utilize shared parameters in both parts of the model. Therefore, in this paper, we consider shared parameter modeling of zero-inflated Gamma data using alternating regression. We consider two different link functions for the Gamma part: canonical link and log link.

%In all of the aforementioned examples, there are explicitly observed covariates or factors. Further, the two parts of the model parameters are perpendicular to each other such that model estimation can be done by simply fitting two separate regressions: binomial regression and Gamma regression. Such models, unfortunately, can not be applied to the user-item or cooccurrence count matrix data that arise from many practical applications. Examples include user-item or item-item cooccurence data from online shopping platforms, cooccurring word-word pairs in sequences of texts, etc.
%One reason is because there are no  covariates observed. Additionally, the mechanism that leads to the observed data entries in the data matrix for the binomial and the Gamma part may be of the same nature, in which case it is more appropriate to utilize shared parameter in both parts of the model. Therefore, in this paper, we consider shared parameter modeling of zero inflated Gamma data using alternating regression. We will consider two different link functions for the Gamma part: canonical link and log link.

{\modify

We believe that our study is the first that utilizes a shared parameter likelihood for zero-inflated skewed distribution to conduct matrix factorization. The alternating least squares (ALS) shares similar spirit as our SA-ZIG in terms of alternately updating parameters.  Most of matrix factorization methods in the literature involving ALS are adding an assumption without realizing it.
In order for ALS to be valid, the data need to have constant variance. This is because the objective function behind the ALS procedure is the mean squared error which gives equal emphasis to all observations regardless of how big or small their variations are. If the variations for different observations are drastically different, it is unfair to treat the residuals equally.
The real data in the high dimensional sparse co-occurrence matrix are often very skewed with a lot of zeros and do not have constant variance. The contribution of this paper is the SA-ZIG model that models the positive co-occurrence data with Gamma distribution and attributes the many zeros in the data as from a Bernoulli distribution. Shared parameters are used in both the Bernoulli and Gamma parts of the model.
The latent row and column vector representations in matrix decomposition can be thought of as missing values that have some distributions relying on a smaller set of parameters. Estimating the vector representation for the rows relies the joint likelihood of the observed matrix data and the missing vector representation for the columns.  Due to missingness,
the estimation of row vector representations is transformed to using conditional likelihood of the observed data given the column vector representations, and vice versa. This alternating update in the end gives the maximum likelihood estimate if the sufficient statistic for the column vector representation is linear in the observed data and the row vector representations. Both ALS and our SA-ZIG rely on this assumption to be valid.

The remainder of this paper is structured as follows: Section \ref{zig_basic} outlines the fundamental framework of the ZIG model. Section \ref{section:ZIGcanonicallink} focuses on parameter estimation within the SA-ZIG model using the canonical link. Section \ref{section:ZIGloglink} addresses parameter estimation for the scenario involving the log link in the Gamma regression component. Convergence analysis is presented in Section \ref{section:convergence}. Section \ref{section:lr_adjustment} details the SA-ZIG algorithms incorporating learning rate adjustments. Section \ref{simulation} presents the experimental studies. Finally, Section \ref{section:conclusion} concludes the paper by summarizing the research findings, contributions, and limitations.}

%
%The remaining part of the paper is organized as follows. Section \ref{zig_basic} lays out the basic structure of the ZIG model. Section \ref{section:ZIGcanonicallink} is devoted to parameter estimation under the
%ZIG model with canonical link. Section \ref{section:ZIGloglink} deals with parameter estimation for the case with log link in Gamma regression part. Convergence analysis is discussed in
%Section \ref{section:convergence}.
%The ZIG algorithms with learning rate adjustment are given in Section \ref{section:lr_adjustment}.
%Experimental studies are given in Section \ref{simulation}. Section \ref{section:conclusion} concludes the paper by summarizing the research findings, contributions, and limitations of the paper.}

\section{Shared parameter alternating zero-inflated Gamma regression}  \label{zig_basic}
Suppose the observed data are $\{y_{ij}, i=1, \ldots, n, \, j=1, \ldots, n\}$ whose distribution depends on some unknown Bernoulli random variable $g_{i j}$ and a Gamma random variable such that
\bqa
g_{i j} = \begin{cases}0, & \text{with probability} \quad 1-p_{i j} \\ 1, & \text{with probability} \quad p_{i j}\end{cases}, \qquad
y_{i j} = \left\{\begin{array}{cc}0, & \text{if} \quad g_{i j}=0 \\ \text { Gamma observation, } & \text{if} \quad g_{i j}=1 \end{array} \right. ,
\eqa
The mechanism behind the observed data may be a result of combined contribution from some covariates or factors but none was observed.

 We can write the probability mass function (pmf) for the Bernoulli random variable and probability density function (pdf) for the Gamma random variable as follows:
\bqa
p\left(g_{i j}\right) &=& p_{i j}^{g_{i j}} \cdot\left(1-p_{i j}\right)^{1-g_{i j}}, \\
f\left(y_{i j} \mid g_{i j}=1\right) &=& \frac{1}{\Gamma\left(v_{i}\right)} \cdot\left(\frac{v_{i} y_{i j}}{\mu_{i j}}\right)^{v_{i}} \cdot \frac{1}{y_{i j}} \cdot \exp \left(-\frac{v_{i} y_{i j}}{\mu_{i j}}\right).
\eqa
Product of the two functions generates the joint distribution of the Bernoulli and Gamma random variable
\bqa
f\left(y_{i j}, g_{i j}\right) &=& p\left(g_{i j}\right) \cdot f\left(y_{i j} \mid g_{i j}\right).
\eqa
By summing up the two possibilities of the Bernoulli random variable, we obtain the pdf of the zero inflated Gamma (ZIG) distribution:
\bqan \label{pdf}
f\left(y_{i j}\right)
%&=&
\!=\!
 \sum_{x=0}^{1} p\left(g_{i j}\!=\! x\right) \cdot f\left(y_{i j} \mid g_{i j}\!=\!x\right)
=
% \\ &=&
\left(1\!-\!p_{i j}\right)^{I\left(y_{i j}=0\right)} \cdot\left[p_{i j} \,\, f\left(y_{i j} \mid g_{i j}\!=\!1\right)\right]^{I\left(y_{i j}>0\right)}.
\eqan

For the Bernoulli part, we use the logit link function to connect the probability $p_{ij}$ to the effects of some unknown covariates or factors as follows:
\bqan
\label{formula:ZIG-logitlink-eta-ij}
\eta_{i j} = \log \frac{p_{i j}}{1-p_{i j}}= \vw_{i}^{\top} \wtvw_{j} + b_{i}+ \wtb_{j},
\eqan
where $\vw_i$ and $b_i$ are unknown parameters related to row $i$ while $\wtvw_j$ and $\wtb_j$ are unknown parameters related column $j$. Assume $\vw_i$ and $\wtvw_j$ are $d$-dimensional vectors and $b_i$, $\wtb_j$ are scalars.
For the Gamma observations,
$$\log f\left(y_{i j} \mid y_{i j}>0\right)=-\log \Gamma\left(v_{i j}\right)+v_{i j} \log \left(\frac{v_{i j}}{\mu_{i j}}\right)+\left(v_{i j}-1\right) \log y_{i j}-\frac{v_{i j} y_{i j}}{\mu_{i j}}.$$
 If using the canonical link, the mean of the response variable %and $\vtheta$ and $\wtvtheta$
 are connected to the unknown parameters through the following formula,
\bqan
\label{formula:ZIG-canonicallink}
%\tau_{i j}=
g(\mu_{i j}) = -\mu_{i j}^{-1} = \vw_{i}^{\top} \wtvw_{j} + e_{i}+ \wte_{j},
\eqan
where $e_i$ and $\wte_j$ are scalars.
With the canonical link, the likelihood function and the score equations are both function of the sufficient statistic. As the sufficient statistic carries all information about the unknown parameters, we can restrict our attention to the sufficient statistic without losing any information. This link function, however, could have difficulty in the estimation process. The natural parameter space is $\{\mu_{i j}: \mu_{i j} > 0\}$. The right hand side of equation (\ref{formula:ZIG-canonicallink}) may yield an estimate of $\mu_{i j}$ that is outside of this parameter space.

Another link function that is popularly used for Gamma distribution is the log link $g(\mu) = \log(\mu)$ which gives the log linear model:
\bqan
\label{formula:ZIG-loglink-tau-ij}
%\tau_{i j} =
g(\mu_{i j}) = \log(\mu_{i j}) = \vw_{i}^{\top} \wtvw_{j} + e_{i}+ \wte_{j}.
\eqan
This link eliminated the non-negativity problem and the model parameters have better interpretation than the canonical link. For each unit increment in $\wtw_{jk}$ or $w_{i k}$, the mean increases multiplicatively by $\exp(w_{i k})$ or by $\exp(\wtw_{j k})$ respectively. Even though the parameters enjoy better interpretation, the estimation could also become a problem in the sense that the dot product on the right hand side of (\ref{formula:ZIG-loglink-tau-ij}) could become large leading to $\mu_{i j}$ being infinity and hence the estimation algorithm diverges.

In both links, the common dot product $\vw_{i}^{\top} \wtvw_{j}$ is used to reflect the fact that the cosine similarity is the key driving force behind the observed data in the table. For example, in natural language processing (NLP), $\vw_i$ and $\wtvw_j$ each represents a word in a dense vector. Each one of them contains both linguistic information and word usage information in it. The observations are distance weighted cooccurrence counts that are linked to relevance between the words and the relevance can be captured with the cosine similarity. In the example of item-item cooccurrence matrix, $\vw_i$ and $\wtvw_j$ represent the hidden product information of the items including characteristics, properties, functionality, popularity, users' ratings on them, etc. The dot product again tells how the two items are relevant to each other. Sometimes the cooccurrence matrix was derived from a time series such as a sequence of watched movies in time order from a customer. In this case, the cooccurrence count tells how often the two items were considered in a similar time frame because the counts were weighted based on the position separation of the two items in the sequence. The relevance of these items summarized by the cosine similarity could reflect how the two products are alike in their property, functionality, etc. Therefore, the dot product serves as major contributor for the observed weighted count.

In both link functions, the intercepts are allowed to be different from that in the logistic model for flexibility. In some classical statistical models such as in \cite{Moulton-et-al:2002}, shared parameter modeling is used by assuming one part of model parameters to be proportional to the other. Such proportionally assumed parameters are meaningful for the case where covariates are observed. In our case, both $\vw$ and $\wtvw$ are unknown. Different pairs of $\vw$ and $\wtvw$ could give the same dot product. Adding an additional proportionality parameter {\modify only makes} the model even more non-identifiable. This is why we believe it is better to put flexibility in the intercepts instead of using the proportionality parameters.

Due to shared parameters being used, the parameters from both the Gamma regression part and the Logistic regression part should be estimated simultaneously. There are past studies that use separate set of parameters. \cite{Mills:2013} conducted hypothesis testing to compare two groups via mixture of zero-inflated Gamma and zero-inflated log-normal models. However, the parameters from the two model components were modelled separately. In a totally applied setting, \cite{Moulton-et-al:2002} considered normal and log-Gamma mixture model for HIV RNA data using shared parameters assuming two parts of the model are proportional to each other. They proposed such application and simulation, but no inference was given. The shared parameter modeling was employed in some literature to achieve some model parsimony. For example, \cite{Wu-Carroll:1988} used shared parameters for both a random effects linear model and a probit-modeled censoring process. \cite{Ten-et-al:1998} used a similar approach for simultaneous modeling of a mean structure and informative censoring. \cite{Albert-et-al:1997} used shared parameters to model the intensity of a Poisson process and a binary measure of severity.

Denote
$\vtheta_i = (\vw_i^{\top}, b_i, e_i)^{\top}$, $\wtvtheta_i = (\wtvw_i^{\top}, \wtb_i, \wte_i)^{\top}$, and
 $\vtheta = (\vtheta_1, \ldots, \vtheta_n)^{\top}$, $\wtvtheta = (\wtvtheta_1, \ldots, \wtvtheta_n)^{\top}$.
We {\modify estimate} $\vtheta$ and $\wtvtheta$ alternately. In this estimation scheme, the likelihood function can be treated as a function of either $\vtheta$ or $\wtvtheta$ but not concurrently at a same time. When estimating $\vtheta$, the likelihood is treated as a function of $\vtheta$ while $\wtvtheta$ stays fixed. Reversely, when we estimate the $\wtvtheta$, the likelihood is treated as a function of $\wtvtheta$ while $\vtheta$ is treated as fixed.
This resembles block coordinate descent in which one part of parameters is updated while holding the remaining part of parameters as fixed values.  For clarity, we use separate notations $\ell_i$ and $\widetilde{\ell}_j$ to refer to these two occasions, i.e.,
\bqa
\ell_i = \ell_i(\vtheta_i; \wtvtheta) = \sum_{j=1}^n \ell_{ij}(\vtheta_i; \wtvtheta_j), \qquad
\widetilde{\ell}_j = \widetilde{\ell}_j(\wtvtheta_j; \vtheta) = \sum_{i=1}^n \ell_{i j}(\wtvtheta_j; \vtheta_i),
\eqa
where $\ell_{ij}(\vtheta_i; \wtvtheta_j)$ and $\ell_{i j}(\wtvtheta_j; \vtheta_i)$ are both equal to $\log(f(y_{ij}))$ but one is treated as a function of $\vtheta_i$ while the other one is treated as a function of $\wtvtheta_j$.

Note that $\ell_i$ and $\widetilde{\ell}_j$ simply represent the log likelihood of a row or a column of the data matrix. The $\ell_i$ can be thought of as the log likelihood function of $\vtheta_i$ for data in the $i^{th}$ row of the co-occurence matrix and the $\widetilde{\ell}_j$ is the log likelihood function of $\wtvtheta_j$ for data in the $j^{th}$ column of the matrix. Each log likelihood functions could be split into two components: one corresponds to the Bernoulli part and the other corresponds to the Gamma part
\bqa
\ell_{i} = \ell_{i}(\vtheta_i ; \wtvtheta) = %\ell_{i}(\vtheta_i \text{ while $\wtvtheta$ is known }) =
 \ell_{i}^{(1)}+\ell_{i}^{(2)}, \quad
 \widetilde{\ell}_{j} = \widetilde{\ell}_{j}(\wtvtheta_j ; \vtheta)
%= \widetilde{\ell}_{j}(\wtvtheta_j \text{ while $\vtheta$ is known })
= \widetilde{\ell}_{j}^{(1)}+\widetilde{\ell}_{j}^{(2)},
 \eqa
 where
 \bqa
\ell_{i}^{(1)} &=& \ell_{i}^{(1)}(\vtheta_i ; \wtvtheta)
= %\ell_i^{(1)} (\vtheta_i \text{ while $\wtvtheta$ is known }) \\
%\quad &=&
\sum_{j=1}^{n}\left\{I\left(y_{i j}=0\right) \cdot \log \left(1-p_{i j}\right)+I\left(y_{i j}>0\right) \cdot \log p_{i j}\right\}, \\
\ell_{i}^{(2)} &=& \ell_{i}^{(2)}(\vtheta_i ; \wtvtheta)
%= \ell_i^{(2)} (\vtheta_i \text{ while $\wtvtheta$ is known })
= \sum_{j=1}^{n} I\left(y_{i j}>0\right) \cdot \log f\left(y_{i j} \mid y_{i j}>0\right),
\\
\widetilde{\ell}_{j}^{(1)} &=& \widetilde{\ell}_{j}^{(1)}(\wtvtheta_j ; \vtheta)
=
%\widetilde{\ell}_j^{(1)} (\widetilde{\vtheta} \text{ while $\vtheta$ is known }) \\
\nonumber
%&=&
\sum_{i=1}^{n}\left\{I\left(y_{i j}=0\right) \cdot \log \left(1-p_{i j}\right)+I\left(y_{i j}>0\right) \cdot \log p_{i j}\right\}, \\
\label{formula:ellj-tilde-2}
\widetilde{\ell}_{j}^{(2)} &=& \widetilde{\ell}_{j}^{(2)}(\wtvtheta_j ; \vtheta) =
%\widetilde{\ell}_j^{(2)} (\wtvtheta_j \text{ while $\vtheta$ is known }) \\
\nonumber
%&=&
 \sum_{i=1}^{n} I\left(y_{i j}>0\right) \cdot \log f\left(y_{i j} \mid y_{i j}>0\right).
\eqa
Assuming that the observations $y_{i j}$'s are independent of each other conditional on unobserved $\vtheta_i$ and $\wtvtheta_j$, $i=1, \ldots, n, j=1, \ldots, n$, the overall loglikelihood function from all observations can be written as
\bqa
%&& L=\prod_{i=1}^{n} L_{i}; \qquad
\ell(\vtheta, \wtvtheta) %=\log(L)
=\sum_{i=1}^{n} \ell_{i} (\vtheta, \wtvtheta)=\sum_{i} (\ell_i^{(1)} + \ell_i^{(2)});
\qquad
\widetilde{\ell}(\wtvtheta ; \vtheta)= \sum_{j=1}^n \widetilde{\ell}_j(\wtvtheta_j ; \vtheta)=\sum_{j} (\widetilde{\ell}_j^{(1)} + \widetilde{\ell}_j^{(2)}).
%\text{where }
%\ell_{i}=\sum_{j'=1}^{n} \ell_{i j'}; \quad \text{with} \quad \ell_{i j'} = \log(f(y_{i j'})).
\eqa

The alternating regression deals with two separate log likelihood functions $\ell(\vtheta ; \wtvtheta)$ and $\widetilde{\ell}(\wtvtheta ; \vtheta)$ respectively.
%\bqa
%&& \ell(\vtheta ; \wtvtheta)=\sum_{i} \sum_{j} \ell_{i j}= \sum_{i} \sum_{j} (\ell_{i j}^{(1)}+\ell_{i j}^{(2)}) = \sum_{i} (\ell_i^{(1)} + \ell_i^{(2)}), \\
%&& \widetilde{\ell}(\wtvtheta ; \vtheta)=\sum_{i} \sum_{j} \widetilde{\ell}_{i j}= \sum_{i} \sum_{j} (\widetilde{\ell}_{i j}^{(1)}+\widetilde{\ell}_{i j}^{(2)}) = \sum_{j} (\widetilde{\ell}_j^{(1)} + \widetilde{\ell}_j^{(2)}).
%\eqa
% Elements under the summation of this log likelihood

% This log likelihood can be split into two components.
% \bqa
% \ell_{i}(\vtheta_i ; \wtvtheta) &=&\ell_{i}^{(1)}+\ell_{i}^{(2)}, \\
% \ell_{i}^{(1)} &=& \sum_{j=1}^{n}\left\{I\left(y_{i j}=0\right) \cdot \log \left(1-p_{i j}\right)+I\left(y_{i j}>0\right) \cdot \log p_{i j}\right\}, \\
% \ell_{i}^{(2)} &=& \sum_{j=1}^{n} I\left(y_{i j}>0\right) \cdot \log f\left(y_{i j} \mid y_{i j}>0\right).
% \eqa
The $\ell_i^{(1)}$ and $\widetilde{\ell}_j^{(1)}$ part is the traditional log likelihood for binary logistic regression. The $\ell_i^{(2)}$ and $\widetilde{\ell}_j^{(2)}$ part is the Gamma log likelihood restricted to only positive observations. If the two parts do not share common parameters, then the estimation can be done separately. % as in \cite{Mills:2013}.
%In our case, however,
However, they
share some common parameters $\vw_i$ and $\wtvw_j$. In the next two sections, we {\modify describe} the parameter estimations.

\section{ZIG model with canonical links}
\label{section:ZIGcanonicallink}

In this section, we consider the case with canonical links, i.e., the Bernoulli part {\modify uses} logit function and the Gamma part {\modify uses} negative inverse link. Using the canonical link with generalized linear models enjoys the benefit that the score equations are a function of sufficient statistics. Here, we consider parameter estimation under the canonical links. The negative inverse link has difficulty to interpret model parameters and also has some restrictions in terms of the support of the link function, which does not match the positive value of Gamma distribution. More details can be seen as we introduce the model.

Recall that the the log likelihood and the link function for the logistic part on the $i^{th}$ row of data are
\bqa
&& \ell_{i}^{(1)} = \sum_{j=1}^{n}\left\{I\left(y_{i j}=0\right) \cdot \log \left(1-p_{i j}\right)+I\left(y_{i j}>0\right) \cdot \log p_{i j}\right\}, \\
&& \eta_{i j} = \log \frac{p_{i j}}{1-p_{i j}}= \vw_{i}^{\top} \wtvw_{j} + b_{i} + \wtb_{j},
\eqa
where $\vw_i$ and $b_i$ are unknown parameters while treating $\wtvw_j$ and $\wtb_j$ as fixed.
The log likelihood and the negative inverse link function for the Gamma part on the $i^{th}$ row of data are
\bqan \label{li2}
&& \ell_{i}^{(2)} = \sum_{j=1}^{n} I\left(y_{i j}>0\right) \cdot \log f\left(y_{i j} \mid y_{i j}>0\right), \\ \nonumber
&& \quad \text{with } \log f\left(y_{i j} \mid y_{i j}>0\right)=-\log \Gamma\left(v_{i j}\right)+v_{i j} \log \left(\frac{v_{i j}}{\mu_{i j}}\right)+\left(v_{i j}-1\right) \log y_{i j}-\frac{v_{i j} y_{i j}}{\mu_{i j}}, \\
&& \tau_{i j} = g(\mu_{i j}) = -\mu_{i j}^{-1} = \vw_{i}^{\top} \wtvw_{j} + e_{i}+ \wte_{j}.
\eqan
The log likelihood function for $i^{th}$ row of data is
\bqa
\ell_{i} &=&\ell_{i}^{(1)}+\ell_{i}^{(2)}.
\eqa
To get partial derivatives, we write $\ell_i^{(1)}$ as follows
\bqan \label{li1}
\ell_{i}^{(1)} &=& \sum_{j=1}^{n}\left(1-g_{i j}\right) \cdot \log \left(1-p_{i j}\right)+g_{i j} \cdot \log p_{i j} %\\
%&=& \sum_{j=1}^{n} g_{i j} \cdot \log \frac{p_{i j}}{1-p_{i j}}+\log \left(1-p_{i j}\right) \\
%&=& \sum_{j=1}^{n} g_{i j} \cdot \eta_{i j}+\log \frac{1}{1+\exp \left(\eta_{i j}\right)} \\
%&=&
=
\sum_{j=1}^{n} g_{i j} \cdot \eta_{i j}-\log \left\{1+\exp \left(\eta_{i j}\right)\right\}.
\eqan
The inverse logit and its partial derivative w.r.t. $\vw_i$ are
%Note that $p_{i j}$ is success probability from Logistic regression, the following formulae hold:
\bqa
p_{i j} =
\frac{\exp \left(\vw_{i}^{\top} \wtvw_{j}+ b_{i}+\wtb_{j}\right)}{1+\exp \left(\vw_{i}^{\top} \wtvw_{j}+b_{i}+\wtb_{j}\right)}, \qquad \frac{\partial p_{i j}}{\partial \vw_{i}}=  p_{i j}\left(1-p_{i j}\right) \wtvw_{j}.  %\\
%\frac{\partial \eta_{i j}}{\partial \vw_{i}} &=& \wtvw_{j}, \qquad \frac{\partial \eta_{i j}}{\partial b_{i}}=1, \qquad \frac{\partial p_{i j}}{\partial b_{i}} = p_{i j}\left(1-p_{i j}\right), \\
%\frac{\partial p_{i j}}{\partial \vw_{i}} &=& \frac{\exp \left(\eta_{i j}\right) \cdot \wtvw_{j} \cdot \left\{1+\exp \left(\eta_{i j}\right)\right\}- \exp \left(2\eta_{i j}\right) \cdot \wtvw_{j}}{\left\{1+\exp \left(\eta_{i j}\right)\right\}^{2}}
%% &=& p_{i j}  \cdot \wtvw_{j}- p_{i j}^{2} \wtvw_{j} \\
%= p_{i j}\left(1-p_{i j}\right) \wtvw_{j}.
\eqa
Therefore, the first order partial derivatives can be summarized as
\bqan
\begin{gathered}
\label{formula:ZIG-canonicallink-1stpartial-ll1}
\frac{\partial \ell_{i}^{(1)}}{\partial \vw_{i}} = \sum_{j=1}^{n}\left\{g_{i j} \cdot \wtvw_{j}-\frac{\exp \left(\eta_{i j}\right)}{1+\exp \left(\eta_{i j}\right)} \wtvw_{j}\right\}
% &=& \sum_{j=1}^{n}\left\{g_{i j} \wtvw_{j}-p_{i j} \wtvw_{j}\right\} \\
= \sum_{j=1}^{n}\left(g_{i j}-p_{i j}\right) \wtvw_{j}, \\
\frac{\partial \ell_{i}^{(1)}}{\partial b_{i}} = \sum_{j=1}^{n}\left(g_{i j}-p_{i j}\right).
\end{gathered}
\eqan
The negative second order partial derivatives and their expectations are the same and are given below
\bqan
%\nonumber
%&&
 -\frac{\partial^{2} \ell_{i}^{(1)}}{\partial \vw_{i} \partial \vw_{i}^{\top}} %=  -\sum_{j=1}^{n} c_{i} \wtc_{j} \wtvw_{j} \frac{\partial p_{i j}}{\partial \vw_{i}}
%-\sum_{j=1}^{n} \wtvw_{j} p_{i j}\left(1-p_{i j}\right) \wtvw_{j}^{\top}
= \sum_{j=1}^{n} p_{i j}\left(1-p_{i j}\right) \wtvw_j \wtvw_{j}^{\top}= E\left[-\frac{\partial^{2} \ell_{i}^{(1)}}{\partial \vw_{i} \partial \vw_{i}^{\top}}\right] %\\
\label{formula:ZIG-canonicallink-Sww-ll1}
%\Longrightarrow && E\left[-\frac{\partial^{2} \ell_{i}^{(1)}}{\partial \vw_{i} \partial \vw_{i}^{\top}}\right] = \sum_{j=1}^{n} p_{i j}\left(1-p_{i j}\right) \cdot \wtvw_{j} \wtvw_{j}^{\top},
\eqan
\bqan
\begin{gathered}
\label{formula:ZIG-canonicallink-Swb-ll1}
-\frac{\partial^{2} \ell_{i}^{(1)}}{\partial \vw_{i} \partial b_{i}} \!=\! \!\sum_{j=1}^{n} \wtvw_{j} \frac{\partial p_{i j}}{\partial b_{i}} \!=\! \!\sum_{j=1}^{n} \wtvw_{j} p_{i j}\left(1\!-\!p_{i j}\right) =
%\quad
%\Longrightarrow \quad E\left[-\frac{\partial^{2} \ell_{i}^{(1)}}{\partial \vw_{i} \partial b_{i}}\right] \!=\! \sum_{j=1}^{n} \wtvw_{j} p_{i j}\left(1\!-\!p_{i j}\right), \\
%\frac{\partial^{2} \ell_{i}^{(1)}}{\partial b_{i}^{2}} = -\sum_{j=1}^{n} p_{i j}\left(1-p_{i j}\right) \quad
%\Longrightarrow  \quad
E\left[-\frac{\partial^{2} \ell_{i}^{(1)}}{\partial b_{i}^{2}}\right]\\
%=\sum_{j=1}^{n} p_{i j}\left(1-p_{i j}\right).
-\frac{\partial^{2} \ell_{i}^{(1)}}{\partial b_{i}^{2}} = \sum_{j=1}^{n} p_{i j}\left(1-p_{i j}\right)
= E\left[-\frac{\partial^{2} \ell_{i}^{(1)}}{\partial b_{i}^{2}}\right]
\end{gathered}
\eqan

%\bqan
%\begin{gathered}
%\label{formula:ZIG-canonicallink-Swb-ll1}
%\frac{\partial^{2} \ell_{i}^{(1)}}{\partial \vw_{i} \partial b_{i}} \!=\! -\!\sum_{j=1}^{n} \wtvw_{j} \frac{\partial p_{i j}}{\partial b_{i}} \!=\! -\!\sum_{j=1}^{n} \wtvw_{j} p_{i j}\left(1\!-\!p_{i j}\right) \quad
%\Longrightarrow \quad E\left[-\frac{\partial^{2} \ell_{i}^{(1)}}{\partial \vw_{i} \partial b_{i}}\right] \!=\! \sum_{j=1}^{n} \wtvw_{j} p_{i j}\left(1\!-\!p_{i j}\right), \\
%\frac{\partial^{2} \ell_{i}^{(1)}}{\partial b_{i}^{2}} = -\sum_{j=1}^{n} p_{i j}\left(1-p_{i j}\right) \quad
%\Longrightarrow  \quad E\left[-\frac{\partial^{2} \ell_{i}^{(1)}}{\partial b_{i}^{2}}\right]=\sum_{j=1}^{n} p_{i j}\left(1-p_{i j}\right).
%\end{gathered}
%\eqan
Now consider the second component of the log likelihood $\ell_i^{(2)}$ from the $i$th row of data
\bqan
\ell_{i}^{(2)} &=& \sum_{j=1}^{n} I\left(y_{i j}>0\right) \cdot \log f\left(y_{i j} \mid y_{i j}>0\right)%=\sum_{j=1}^{n} g_{i j} \cdot \log f\left(y_{i j} \mid y_{i j}>0\right)
\nonumber\\
%&=& \sum_{j=1}^{n} g_{i j} \cdot\left\{-\log \Gamma\left(v_{i j}\right)+v_{i j} \log v_{i j}+v_{i j} \log \left(-\vw_{i}^{\top} \wtvw_{j}-e_{i}-\wte_{j}\right)\right. \nonumber\\
%&& \left.\quad+\left(v_{i j}-1\right) \log y_{i j}+v_{i j} y_{i j}\left(\vw_{i}^{\top} \wtvw_{j}+e_{i}+\wte_{j}\right)\right\} \nonumber\\
&=& \sum_{j=1}^{n} g_{i j} \cdot\left\{v_{i j} \log \left(-\vw_{i}^{\top} \wtvw_{j}-e_{i}-\wte_{j}\right)+v_{i j} y_{i j}\left(\vw_{i}^{\top} \wtvw_{j}+e_{i}+\wte_{j}\right)\right. \label{formula:ll2} \\
&&\left.\quad-\log \Gamma\left(v_{i j}\right)+v_{i j} \log v_{i j}+\left(v_{i j}-1\right) \log y_{i j}\right\}. \nonumber
\eqan
The first order partial derivatives for the $\ell_{i}^{(2)}$ are
\bqan
\begin{gathered}
\label{formula:ZIG-canonicallink-1stpartial-ll2}
\frac{\partial \ell_{i}^{(2)}}{\partial \vw_{i}} \!=\! \sum_{j=1}^{n} g_{i j}\! \left\{\frac{v_{i j} \cdot \wtvw_{j}}{\vw_{i}^{\top} \wtvw_{j}\!+\!e_{i}\!+\!\wte_{j}}+v_{i j} y_{i j}  \wtvw_{j}\right\},
\quad
\frac{\partial \ell_{i}^{(2)}}{\partial e_{i}} \!=\! \sum_{j=1}^{n} g_{i j} \!\left\{\frac{v_{i j}}{\vw_{i}^{\top} \wtvw_{j}\!+\!e_{i}\!+\!\wte_{j}}\!+\!v_{i j} \cdot y_{i j}\right\}.
\end{gathered}
\eqan
The second order partial derivatives and their negative expectations are
\bqan
%\nonumber
%&&\frac{\partial^{2} \ell_{i}^{(2)}}{\partial \vw_{i} \partial \vw_{i}^{\top}} = \sum_{j=1}^{n} g_{i j} \cdot v_{i j} \cdot \wtvw_{j} \cdot\left(-\wtvw_{j}^{\top}\right) /\left(-\vw_{i}^{\top} \wtvw_{j}-e_{i}-\wte_{j}\right)^{2} \\
\label{formula:ZIG-canonicallink-Sww-ll2}
%%\Longrightarrow
%&&
 \left.-\frac{\partial^{2} \ell_{i}^{(2)}}{\partial \vw_{i} \partial \vw_{i}^{\top}}\right|_{g_{ij}=1} =
  \sum_{\substack{j=1 \\ g_{i j}=1}}^{n}
%\sum_{j=1 }^{n} p_{i j}
 \frac{v_{i j} \wtvw_{j} \cdot\left(\wtvw_{j}^{\top}\right)} {\left(\vw_{i}^{\top} \wtvw_{j}+e_{i}+\wte_{j}\right)^{2}}=
 E\left[\left.-\frac{\partial^2 \ell_{i}^{(2)}}{\partial \vw_{i} \partial \vw_{i}^{\top}} \right| g_{i j}=1 \right];
\eqan
\bqan
%\nonumber
%&& \frac{\partial^{2} \ell_{i}^{(2)}}{\partial \vw_{i} \partial e_{i}}=\sum_{j=1}^{n} g_{i j} \cdot v_{i j} \cdot \wtvw_{j} \cdot(-1) /\left(-\vw_{i}^{\top} \wtvw_{j}-e_{i}-\wte_{j}\right)^{2} \\
\label{formula:ZIG-canonicallink-Swe-ll2}
%\Longrightarrow &&
 \left.-\frac{\partial^{2} \ell_{i}^{(2)}}{\partial \vw_{i} \partial e_{i}}\right|_{g_{i j}=1}=\sum_{\substack{j=1 \\ g_{i j}=1}}^{n} \frac{  v_{i j} \wtvw_{j}  }{\left(\vw_{i}^{\top} \wtvw_{j}+e_{i}+\wte_{j}\right)^{2}} =
 E\left[\left.-\frac{\partial^{2} \ell_{i}^{(2)}}{\partial \vw_{i} \partial e_{i}} \right| g_{i j}=1 \right]
 %=v_{i j} \sum_{\substack{j=1 \\g_{i j}=1}}^{n}
%%\sum_{j=1 }^{n} p_{i j}
% \wtvw_{j} /\left(-\vw_{i}^{\top} \wtvw_{j}-e_{i}-\wte_{j}\right)^{2};
\eqan
\bqan
%\nonumber
%&& \frac{\partial^{2} \ell_{i}^{(2)}}{\partial e_{i}^{2}}=\sum_{j=1}^{n} g_{i j} \cdot v_{i j} \cdot(-1) /\left(-\vw_{i}^{\top} \wtvw_{j}-e_{i}-\wte_{j}\right)^{2} \\
\label{formula:ZIG-canonicallink-See-ll2}
%\Longrightarrow &&
 \left.-\frac{\partial^{2} \ell_{i}^{(2)}}{\partial e_{i}^{2}}\right|_{g_{ij}=1} = \sum_{\substack{j=1 \\ g_{i j}=1}}^{n}
%\sum_{j=1 }^{n} p_{i j}
v_{i j}  \left(\vw_{i}^{\top} \wtvw_{j}+e_{i}+\wte_{j}\right)^{-2}
=
 E\left[\left.-\frac{\partial^{2} \ell_{i}^{(2)}}{\partial e_{i}^{2}} \right| g_{i j}=1 \right].
\eqan
All of the aforementioned formulae work with the $i^{th}$ row of data.  When we combine the log likelihood from different rows of data, other rows do not contribute to the partial derivative with respect to $\vtheta_i$. That is,
\bqan \label{partial_canonical1}
&& \frac{\partial \ell}{\partial \vw_{i}}=\frac{\partial \ell_{i}}{\partial \vw_{i}} ; \quad \frac{\partial \ell}{\partial b_{i}}=\frac{\partial \ell_{i}^{(1)}}{\partial b_{i}} ; \quad \frac{\partial \ell}{\partial e_{i}}=\frac{\partial \ell_{i}^{(2)}}{\partial e_{i}} ;
\\ \label{partial_canonical2}
&& \frac{\partial^{2} \ell}{\partial \vw_{i} \partial \vw_{i}^{\top}}=\frac{\partial^{2} \ell_{i}}{\partial \vw_{i} \partial \vw_{i}^{\top}} ; \quad \frac{\partial^{2} \ell}{\partial \vw_{i} \partial b_{i}}=\frac{\partial^{2} \ell_{i}^{(1)}}{\partial \vw_{i} \partial b_{i}} ; \quad \frac{\partial^{2} \ell}{\partial \vw_{i} \partial e_{i}}=\frac{\partial^{2} \ell_{i}^{(2)}}{\partial \vw_{i} \partial e_{i}} ; \\
&& \label{partial_canonical3}
\frac{\partial^{2} \ell}{\partial b_{i}^{2}}=\frac{\partial^{2} \ell_{i}^{(1)}}{\partial b_{i}^{2}} ; \quad \frac{\partial^{2} \ell}{\partial b_{i} \partial e_{i}} = \frac{\partial^{2} \ell_i}{\partial b_{i} \partial e_{i}} = 0 ; \quad \frac{\partial^{2} \ell}{\partial e_{i}^2 } = \frac{\partial^{2} \ell_i^{(2)}}{\partial e_{i}^2}.
\eqan

As a result, the estimation of the components in $\vtheta = (\vtheta_1^\top, \ldots, \vtheta_n^\top)^\top$ does not need to be done simultaneously. Instead, we can cycle through the estimation of $\vtheta_i$, for $i=1, \ldots, n$, one by one iteratively. After $\vtheta_i$ is updated, the estimated value of $\vtheta_1, \vtheta_2, \ldots, \vtheta_i$ {\modify is used to} update $\vtheta_k, k=i+1, \ldots, n$.

The first part of the alternating regression has the following updating equation based on the Fisher scoring algorithm:
\bqan
\label{formula:ZIG-canonicallink-update-it}
\vtheta_{i}^{(t+1)}=\vtheta_{i}^{(t)}+S_{\vtheta_{i}^{(t)}}^{-1} U_{\vtheta_{i}^{(t)}}, \quad i=1, \ldots, n,
\eqan
For each $i$ and $t$, this update requires the value of $\vtheta_i$ and their score equations and information matrix at the $t^{th}$ iteration. One iteration alone here is not taking advantage of the data because retrieving the $i^{th}$ row of data takes significant amount of time when the dimension of the data matrix is huge. Therefore, for each row of data retrieved, it is better to update $\vtheta_i^{(t)}$ a certain number of times. Specifically, the $\vtheta_i^{(t)}$ is updated again and again in a loop of E epochs and the updated values are used to recompute the score equations and information matrix, all of which are used for next epoch's update. After all epochs are completed, $\vtheta_i^{(t)}$ takes the value of $\vtheta_i^{(t, E)}$ at the end of all iterations from E epochs based on the updating formula below:
\bqan
\label{formula:ZIG-canonicallink-update-epoch}
\vtheta_{i}^{(t, k+1)}=\vtheta_{i}^{(t, k)}+S_{\vtheta_{i}^{(t, k)}}^{-1} U_{\vtheta_{i}^{(t, k)}}, \quad k=1, \ldots, E,
\eqan
This inner loop of updates makes good use of the already loaded data to refine the estimate of $\vtheta_i^{(t)}$ so that the end estimate is closer to its MLE when the same current $\wtvtheta^{(c)}$ values are used. Note that the updated $\vtheta_i$ in each epoch from (\ref{formula:ZIG-canonicallink-update-epoch}) {\modify leads} to changes in the score equations and Fisher information matrix, whose update will in turn {\modify result in} better estimate of $\vtheta_i$. Such multiple rounds of updates in (\ref{formula:ZIG-canonicallink-update-epoch}) {\modify reduces} the variation of update in $\vtheta_i^{(t)}$ when we go through many iterations based on equation (\ref{formula:ZIG-canonicallink-update-it}). It effectively reduces the number of times to retrieve the data. Below are the formulae involved in the updating equations:
\bqa
\begin{array}{ccc}
\vtheta_{i}^{(t)} \!=\!\left(\!\vw_{i}^{\top (t)}, b_{i}^{(t)}, e_{i}^{(t)}\right)^{\top}\!; & \, U_{\vtheta_{i}^{(t)}}\! =\! \left.\left[\left(\frac{\partial \ell_{i}}{\partial \vw_{i}}\right)^\top, \frac{\partial \ell_{i}}{\partial b_{i}}, \frac{\partial \ell_{i}}{\partial e_{i}}\right]^{\top}\right|_{\vtheta_{i}^{(t)} }\!;
&
\, S_{\vtheta_i^{(t)}}\! =\! \left[\begin{array}{lll}
S_{\vw_{i}}^{(t)} & S_{\vw_{i} b_{i}}^{(t)} & S_{\vw_{i} e_{i}}^{(t)} \\
S_{b_{i} \vw_{i}}^{(t)} & S_{b_{i}}^{(t)} & S_{b_{i} e_{i}}^{(t)} \\
S_{e_{i} \vw_{i}}^{(t)} & S_{e_{i} b_{i}}^{(t)} & S_{e_{i}}^{(t)}
\end{array}\right]
\end{array}
\eqa
with
% and $\frac{\partial \ell_{i}}{\partial \vw_{i}}, \frac{\partial \ell_{i}}{\partial b_{i}}, \frac{\partial \ell_{i}}{\partial e_{i}}$, and elements of $S_{\vtheta_i^{(t)}}$ are functions of $\vtheta_i^{(t)}, \quad \wtvtheta^{(t)} \!=\! (\wtvtheta_1^{(t)\top}, \ldots, \wtvtheta_n^{(t)\top})^\top$:
%\bqa
%&&  \left.\frac{\partial \ell_{i}}{\partial \vw_{i}}\right|_{\vtheta_{i}^{(t)} } = \frac{\partial \ell_{i}^{(1)}(\vtheta_i^{(t)}; \wtvtheta^{(c)})}{\partial \vw_{i}} + \frac{\partial \ell_{i}^{(2)}(\vtheta_i^{(t)}; \wtvtheta^{(c)})}{\partial \vw_{i}}, \\
%&&  \left.\frac{\partial \ell_{i}}{\partial b_{i}}\right|_{\vtheta_{i}^{(t)} }=\frac{\partial \ell_{i}^{(1)}(\vtheta_i^{(t)}; \wtvtheta^{(c)})}{\partial b_{i}}, \qquad \left.\frac{\partial \ell_{i}}{\partial e_{i}}\right|_{\vtheta_{i}^{(t)} }=\frac{\partial \ell_{i}^{(2)}(\vtheta_i^{(t)}; \wtvtheta^{(c)})}{\partial e_{i}},
%\eqa
\bqa
  \left.\frac{\partial \ell_{i}}{\partial \vw_{i}}\right|_{\vtheta_{i}^{(t)} } \!=\! \sum_{k=1}^2\frac{\partial \ell_{i}^{(k)}(\vtheta_i^{(t)}; \wtvtheta^{(c)})}{\partial \vw_{i}}, \qquad
  \left.\frac{\partial \ell_{i}}{\partial b_{i}}\right|_{\vtheta_{i}^{(t)} }\!=\!\frac{\partial \ell_{i}^{(1)}(\vtheta_i^{(t)}; \wtvtheta^{(c)})}{\partial b_{i}}, \qquad \left.\frac{\partial \ell_{i}}{\partial e_{i}}\right|_{\vtheta_{i}^{(t)} }\!=\!\frac{\partial \ell_{i}^{(2)}(\vtheta_i^{(t)}; \wtvtheta^{(c)})}{\partial e_{i}},
\eqa

where these partial derivatives are the $\frac{\partial \ell_{i}^{(1)}}{\partial \vw_{i}}$, $\frac{\partial \ell_{i}^{(2)}}{\partial \vw_{i}}$, $\frac{\partial \ell_{i}^{(1)}}{\partial b_{i}}$, and $\frac{\partial \ell_{i}^{(2)}}{\partial e_{i}}$ evaluated at $\vtheta_i^{(t)}$ and $\wtvtheta^{(c)}$ using formulae in (\ref{formula:ZIG-canonicallink-1stpartial-ll1}), (\ref{formula:ZIG-canonicallink-1stpartial-ll2}), and
\bqa
&& S_{\vw_{i}}^{(t)}=E\left[-\frac{\partial^{2} \ell_{i}^{(1)}(\vtheta_i^{(t)}; \wtvtheta^{(c)})}{\partial \vw_{i} \partial \vw_{i}^{\top}}\right]+E\left[\left. -\frac{\partial^{2} \ell_{i}^{(2)}(\vtheta_i^{(t)}; \wtvtheta^{(c)})}{\partial \vw_{i} \partial \vw_{i}^{\top}}\right| g_{i j}=1\right] ; \\
&& S_{\vw_{i} b_{i}}^{(t)}=E\left[-\frac{\partial^{2} \ell_{i}^{(1)}(\vtheta_i^{(t)}; \wtvtheta^{(c)})}{\partial \vw_{i} \partial b_{i}}\right]; \qquad S_{b_{i}}^{(t)}=E\left[-\frac{\partial^{2} \ell_{i}^{(1)}(\vtheta_i^{(t)}; \wtvtheta^{(c)})}{\partial b_{i}^{2}}\right] ; \qquad S_{b_{i} e_{i}}^{(t)}=0 ; \\
&& S_{\vw_{i} e_{i}}^{(t)}=E\left[\left. -\frac{\partial^{2} \ell_{i}^{(2)}(\vtheta_i^{(t)}; \wtvtheta^{(c)})}{\partial \vw_{i}  \partial e_{i}}\right|g_{i j}=1 \right]; \qquad
S_{e_{i}}^{(t)} = E\left[\left. -\frac{\partial^{2} \ell_{i}^{(2)}(\vtheta_i^{(t)}; \wtvtheta^{(c)})}{\partial e_{i}^2}\right|g_{i j}=1; \right],
\eqa
where the expectations are based on formulae (\ref{formula:ZIG-canonicallink-Sww-ll1}), (\ref{formula:ZIG-canonicallink-Swb-ll1}), (\ref{formula:ZIG-canonicallink-Sww-ll2}), (\ref{formula:ZIG-canonicallink-Swe-ll2}), (\ref{formula:ZIG-canonicallink-See-ll2}) evaluated at $\vtheta_i^{(t)}$ and $\wtvtheta^{(c)}$ if it is in outer loop of update and using the values of $\vtheta^{(t, k)}$ and $\wtvtheta^{(c)}$ when it is in the inner loop of updates.

This concludes the first part of the alternating ZIG regression. It gives iterative update of $\vtheta$ while $\wtvtheta$ is fixed. After we finish updating all $\vtheta_i$'s, we move on to treat the $\vtheta$ as fixed to estimate $\wtvtheta$.

Starting with $\widetilde{\ell}_j = \widetilde{\ell}_j^{(1)} + \widetilde{\ell}_j^{(2)}$, we consider the partial derivatives with respect to $\wtvtheta_j$ while holding $\vtheta$ fixed. The derivations are similar to those for $\ell_i$ but still we list them for clarity. Note that
\bqa
\frac{\partial p_{i j}}{\partial \wtvw_{j}} &=& \frac{\exp \left(\eta_{i j}\right) \cdot \vw_{i}\left\{1+\exp \left(\eta_{i j}\right)\right\}-\exp \left(2\eta_{i j}\right) \cdot \vw_{i}}{\left\{1+\exp \left(\eta_{i j}\right)\right\}^{2}}
% &=& p_{i j} \cdot \vw_{i}-\left(p_{i j}\right)^{2} \cdot \vw_{i} \\
= p_{i j}\left(1-p_{i j}\right) \vw_{i}, \\
\frac{\partial p_{i j}}{\partial \wtb_{j}} &=& p_{i j}\left(1-p_{i j}\right).
\eqa
The first order partial derivatives regarding $\widetilde{\ell}_j^{(1)}$ are
\bqa
\frac{\partial \widetilde{\ell}_{j}^{(1)}}{\partial \wtvw_{j}} = \sum_{i=1}^{n}\left(g_{i j}-p_{i j}\right) \cdot \vw_{i},
\qquad
%= \sum_{i=1}^{n}\left(g_{i j}-p_{i j}\right) \cdot \vw_{i}, \\
\frac{\partial \widetilde{\ell}_{j}^{(1)}}{\partial \wtb_{j}} = \sum_{i=1}^{n}\left(g_{i j}-p_{i j}\right).
%= \sum_{i=1}^{n}\left(g_{i j}-p_{i j}\right).
\eqa
The negative second derivatives and their expectations are
\bqan \label{formual:ZIG-loglink-partial2nd-ll1-0}
&& -\frac{\partial^{2} \widetilde{\ell}_{j}^{(1)}}{\partial \wtvw_{j} \partial \wtvw_{j}^{\top}}= \sum_{i=1}^{n} \vw_{i} \cdot\left(\frac{\partial p_{i j}}{\partial \wtvw_{j}^{\top}}\right)= \sum_{i=1}^{n} \vw_{i} \vw_{i}^{\top} p_{i j}\left(1-p_{i j}\right) = E\left[-\frac{\partial^{2} \widetilde{\ell}_{j}^{(1)}}{\partial \wtvw_{j} \partial \wtvw_{j}^{\top}}\right], %\\
%\Longrightarrow && E\left[-\frac{\partial^{2} \widetilde{\ell}_{j}^{(1)}}{\partial \wtvw_{j} \partial \wtvw_{j}^{\top}}\right]= \sum_{i=1}^{n} \vw_{i} \vw_{i}^{\top} p_{i j}\left(1-p_{i j}\right);
\eqan %\vspace{-0.5in}
\bqan
&& \label{formual:ZIG-loglink-partial2nd-ll1}
-\frac{\partial^2 \widetilde{\ell}_{j}^{(1)}}{\partial \wtvw_{j} \partial \wtb_{j}}=\sum_{i=1}^{n} \vw_{i} p_{i j}\left(1-p_{i j}\right) =
E\left[-\frac{\partial^{2} \widetilde{\ell}_{j}^{(1)}}{\partial \wtvw_{j} \partial b_{j}^{2}}\right],
%\quad
%\Longrightarrow \quad E\left[-\frac{\partial^{2} \widetilde{\ell}_{j}^{(1)}}{\partial \wtvw_{j} \partial b_{j}^{2}}\right]= \sum_{i=1}^{n} \vw_{i} p_{i j}\left(1-p_{i j}\right);
\\ \label{formual:ZIG-loglink-partial2nd-ll1-2}
&& -\frac{\partial^{2} \widetilde{\ell}_{j}^{(1)}}{\partial \wtb_{j}^{2}}=\sum_{i=1}^{n} p_{i j}\left(1-p_{i j}\right) = E\left[-\frac{\partial^{2} \widetilde{\ell}_{j}^{(1)}}{\partial \wtb_{j}^{2}}\right].
 %\quad
%\Longrightarrow \quad E\left[-\frac{\partial^{2} \widetilde{\ell}_{j}^{(1)}}{\partial \wtb_{j}^{2}}\right]= \sum_{i=1}^{n} p_{i j}\left(1-p_{i j}\right).
\eqan
The first order partial derivatives of the second component are $\widetilde{\ell}_j^{(2)}$.
\bqa
 \frac{\partial \widetilde{\ell}_{j}^{(2)}}{\partial \wtvw_{j}} \!= \! \sum_{i=1}^{n} g_{i j} \!\left\{\frac{v_{i j} \cdot \vw_{i}}{\vw_{i}^{\top} \wtvw_{j} \!+ \!e_{i} \!+ \!\wte_{j}}+v_{i j} y_{i j}  \vw_{i}\right\}, \quad
 \frac{\partial \widetilde{\ell}_{j}^{(2)}}{\partial \wte_{j}} \!=\! \sum_{i=1}^{n} g_{i j} \!\left\{\frac{v_{i j}}{\vw_{i}^{\top} \wtvw_{j} \!+ \!e_{i} \!+ \!\wte_{j}}+v_{i j}  y_{i j}\right\} .
\eqa
The negative second order partial derivatives and their expectations are
\bqa
&& \left.-\frac{\partial^{2} \widetilde{\ell}_{j}^{(2)}}{\partial \wtvw_{j} \partial \wtvw_{j}^{\top}}\right|_{g_{i j}=1}=\sum_{\substack{i=1 \\ g_{i j} =1}}^{n} \frac{  v_{i j}  \vw_{i} \vw_{i}^{\top} }{\left(\vw_{i}^{\top} \wtvw_{j}+e_{i}+\wte_{j}\right)^{2}}=
E\left[\left.-\frac{\partial^{2} \widetilde{\ell}_{j}^{(2)}}{\partial \wtvw_{j} \partial \wtvw_{j}^{\top}}\right| g_{i j}=1\right], \\
&&
\left.-\frac{\partial^{2} \widetilde{\ell}_{j}^{(2)}}{\partial \wtvw_{j} \partial \wte_{j}}\right|_{g_{i j}=1} =
\sum_{\substack{i=1 \\ g_{i j} =1}}^{n}
%\sum_{i=1 }^{n} p_{i j}
 \frac{v_{i j} \vw_{i} }{\left(\vw_{i}^{\top} \wtvw_{j}+e_{i}+\wte_{j}\right)^{2}}=E\left[\left.-\frac{\partial^{2} \widetilde{\ell_{j}^{(2)}}}{\partial \wtvw_{j} \partial \wte_{j}}\right| g_{i j}=1 \right], \\
&&
\left.-\frac{\partial^{2} \widetilde{\ell}_{j}^{(2)}}{\partial \wte_{j}^{2}}\right|_{g_{i j}=1}
=
\sum_{\substack{i=1 \\ g_{i j}=1}}^{n}
%\sum_{i=1 }^{n} p_{i j}
v_{i j} \left(\vw_{i}^{\top} \wtvw_{j}+e_{i}+\wte_{j}\right)^{-2}=
E\left[\left.-\frac{\partial^{2} \widetilde{\ell}_{j}^{(2)}}{\partial \wte_{j}^{2}}\right| g_{i j}=1\right].
\eqa

Therefore, the other side of updating equation for the alternating ZIG regression based on the Fisher scoring algorithm is
\bqa
\wtvtheta_{j}^{(t+1)}=\wtvtheta_{j}^{(t)}\!+\!S_{\wtvtheta_{j}^{(t)}}^{-1} U_{\wtvtheta_{j}^{(t)}}, \quad j=1, \ldots, n; \qquad
\wtvtheta_{j}^{(t, k+1)}=\wtvtheta_{j}^{(t, k)}\!+\!S_{\wtvtheta_{j}^{(t, k)}}^{-1} U_{\wtvtheta_{j}^{(t, k)}}, \quad k=1, \ldots, E.
\eqa
Again, for each $j$ and iteration number $t$, this equation is iterated for a certain number of epochs to get refined estimate of $\wtvtheta_j$ based on current value $\vtheta^{(c)}$ without having to reload the data.
The quantities in the updating equation are listed below.
\bqa
\begin{array}{ccc}
\wtvtheta_{j}^{(t)} =\left(\wtvw_{j}^{\top(t)}, \wtb_{j}^{(t)}, \wte_{j}^{(t)}\right)^{\top}; &
U_{\wtvtheta_{j}^{(t)}} = \left.\left[\begin{array}{c}\frac{\partial \widetilde{\ell}_{j}}{\partial \wtvw_{j}} \\ \frac{\partial \widetilde{\ell}_{j}}{\partial \wtb_{j}},\\ \frac{\partial \widetilde{\ell}_{j}}{\partial \wte_{j}}
\end{array}
\right] \right|_{\wtvtheta_{j}^{(t)}};
&
\quad S_{\wtvtheta_j^{(t)}} = \left[\begin{array}{lll}
S_{\wtvw_{j}}^{(t)} & S_{\wtvw_{j} \wtb_{j}}^{(t)} & S_{\wtvw_{j} \wte_{j}}^{(t)} \\
S_{\wtb_{j} \wtvw_{j}}^{(t)} & S_{\wtb_{j}}^{(t)} & S_{\wtb_{j} \wte_{j}}^{(t)} \\
S_{\wte_{j} \wtvw_{j}}^{(t)} & S_{\wte_{j} \wtb_{j}}^{(t)} & S_{\wte_{j}}^{(t)}
\end{array}\right]
\end{array}
\eqa
with
%\bqa
%&& \left.\frac{\partial \widetilde{\ell}_{j}}{\partial \wtvw_{j}}\right|_{\wtvtheta_{j}^{(t)}}\!=\!\frac{\partial \widetilde{\ell}_{j}^{(1)}(\wtvtheta_j^{(t)}; \vtheta^{(c)})}{\partial \wtvw_{j}} + \frac{\partial \widetilde{\ell}_{j}^{(2)}(\wtvtheta_j^{(t)}; \vtheta^{(c)})}{\partial \wtvw_{j}},  \\
%&& \left. \frac{\partial \widetilde{\ell}_{j}}{\partial \wtb_{j}} \right|_{\wtvtheta_{j}^{(t)}}=\frac{\partial \widetilde{\ell}_{j}^{(1)}(\wtvtheta_j^{(t)}; \vtheta^{(c)})}{\partial \wtb_{j}}, \qquad \left. \frac{\partial \widetilde{\ell}_{j}}{\partial \wte_{j}} \right|_{\wtvtheta_{j}^{(t)}}=\frac{\partial \widetilde{\ell}_{j}^{(2)}(\wtvtheta_j^{(t)}; \vtheta^{(c)})}{\partial \wte_{j}},
%\eqa
\bqa
 \left.\frac{\partial \widetilde{\ell}_{j}}{\partial \wtvw_{j}}\right|_{\wtvtheta_{j}^{(t)}}\!=\!\sum_{k=1}^2 \frac{\partial \widetilde{\ell}_{j}^{(k)}(\wtvtheta_j^{(t)}; \vtheta^{(c)})}{\partial \wtvw_{j}},
\qquad \left. \frac{\partial \widetilde{\ell}_{j}}{\partial \wtb_{j}} \right|_{\wtvtheta_{j}^{(t)}}\!=\!\frac{\partial \widetilde{\ell}_{j}^{(1)}(\wtvtheta_j^{(t)}; \vtheta^{(c)})}{\partial \wtb_{j}}, \qquad \left. \frac{\partial \widetilde{\ell}_{j}}{\partial \wte_{j}} \right|_{\wtvtheta_{j}^{(t)}}\!=\!\frac{\partial \widetilde{\ell}_{j}^{(2)}(\wtvtheta_j^{(t)}; \vtheta^{(c)})}{\partial \wte_{j}},
\eqa
where these partial derivatives are $\frac{\partial \widetilde{\ell}_{j}^{(1)}}{\partial \wtvw_{j}}$, $\frac{\partial \widetilde{\ell}_{j}^{(2)}}{\partial \wtvw_{j}}$, $\frac{\partial \widetilde{\ell}_{j}^{(1)}}{\partial \wtb_{j}}$, and $\frac{\partial \widetilde{\ell}_{j}^{(2)}}{\partial \wte_{j}}$ evaluated at $\wtvtheta_j^{(t)}$ and $\vtheta^{(c)}$, and
\bqa
&& S_{\wtvw_{j}}^{(t)}=E\left[-\frac{\partial^{2} \widetilde{\ell}_{j}^{(1)}(\wtvtheta_j^{(t)}; \vtheta^{(c)})}{\partial \wtvw_{j} \partial \wtvw_{j}^{\top}}\right] + E\left[\left.-\frac{\partial^{2} \widetilde{\ell}_{j}^{(2)}(\wtvtheta_j^{(t)}; \vtheta^{(c)})}{\partial \wtvw_{j} \partial \wtvw_{j}^{\top}}\right| g_{i j}=1 \right] ;\\
&& S_{\wtvw_{j} \wtb_{j}}^{(t)}=E\left[-\frac{\partial^{2} \widetilde{\ell}_{j}^{(1)}(\wtvtheta_j^{(t)}; \vtheta^{(c)})}{\partial \wtvw_{j} \partial \wtb_{j}}\right] ; \quad
S_{\wtb_{j}}^{(t)}=E\left[-\frac{\partial^{2} \widetilde{\ell}_{j}^{(1)}(\wtvtheta_j^{(t)}; \vtheta^{(c)},)}{\partial \wtb_{j}^{2}}\right] ; \quad
S_{\wtb_{j} \wte_{j}}^{(t)}=0 ; \\
&& S_{\wtvw_{j} \wte_{j}}^{(t)}=E\left[\left.-\frac{\partial^{2} \widetilde{\ell}_{j}^{(2)}(\wtvtheta_j^{(t)}; \vtheta^{(c)})}{\partial \wtvw_{j}  \partial \wte_{j}}\right| g_{i j}=1 \right]; \quad
S_{\wte_{j}}^{(t)} = E\left[\left.-\frac{\partial^{2} \widetilde{\ell}_{j}^{(2)}(\wtvtheta_j^{(t)}; \vtheta^{(c)})}{\partial \wte_{j}^2}\right| g_{i j}=1 \right].
\eqa
The expectations involved in above formulae were using $\wtvtheta^{(t)}$ at $t^{th}$ iteration and the current values $\vtheta^{(c)}$ if it is in outer loop of update and using the values of $\wtvtheta^{(t, k)}$ and $\vtheta^{(c)}$ when it is in the inner loop of updates.

All of the formulae above assume the parameter $v_{i j}$ is known. When it is not known, the parameter can be estimated with MLE, bias corrected MLE, or moment estimator. Simulation studies in \cite{Du:2007} suggested that when the sample size is large, all the estimators perform similarly but the moment estimator has advantage of easy to compute. If the sample size is medium, then bias corrected MLE is better.

 The canonical link for Gamma regression part could encounter problem. Recall the canonical link for the Gamma regression is $g(\mu_{i j}) = -\mu_{i j}^{-1} = \vw_i^{\top} \wtvw_j + e_i + \wte_j$. The left hand side of the equation is required to be negative because the support of the Gamma distribution is (0, $\infty)$. However, the right hand side of the equation could freely take any value in $(-\infty, \infty)$. Due to this conflict in natural parameter space and the range of estimated value, the likelihood function sometimes could become undefined because $\log(\mu_{ij}^{-1})$ appears in it but   $\log(\widehat{\mu}_{i j}^{-1})$
  %$\log(\vw_i^{\top} \wtvw_j + e_i + \wte_j)$
  can not be evaluated for negative $\widehat{\mu}_{ij}$
  (see equation (\ref{formula:ll2})).

\section{Using log link for Gamma regression}
\label{section:ZIGloglink}

%As we discussed the limitation of the canonical link for Gamma regression part in Section \ref{section:ZIGcanonicallink}, we utilize the other link, i.e.,
In this section, we consider using the log link to model the Gamma regression part while maintaining logistic regression part. The settings of the two-parts model are similar to those in Section \ref{section:ZIGcanonicallink} except that modifying the link function from canonical to log link changes the score equations and the hessian matrix.

For updating formulae on the model parameters based on Fisher scoring algorithm,
the first order and second order partial derivatives of $\ell_i^{(1)}$ remain the same as in (\ref{formula:ZIG-canonicallink-1stpartial-ll1}) -
%{formula:ZIG-canonicallink-Sww-ll1}
(\ref{formula:ZIG-canonicallink-Swb-ll1}) because the new link function does not appear in the Bernoulli part.
Now consider the second component of the log likelihood $\ell_i^{(2)}$, which was given in (\ref{li2}).
In this section, denote $\tau_{ij} = \vw_i^{\top} \wtvw_j + e_i + \wte_j$ as in previous subsection. Then the log likelihood of the ith row of non-zero observations in cooccurrence matrix, corresponding to Gamma distribution can be expressed as
\bqan
\nonumber
&& \ell_{i}^{(2)} = \sum_{j=1}^{n} I\left(y_{i j}>0\right) \cdot \log f\left(y_{i j} \mid y_{i j}>0\right)=\sum_{j=1}^{n} g_{i j} \cdot \log f\left(y_{i j} \mid y_{i j}>0\right) \\
&& \,\,\,\, \quad = \sum_{j=1}^{n} g_{i j} \!\! \left\{\!-\!\log \Gamma\!\left(\!v_{i j}\!\right)\!+\!v_{i j}\! \log \!v_{i j}\!+\!\left(v_{i j}\!-\!1\right)\! \log y_{i j}\!-\!v_{i j}\tau_{i j}\!-\!v_{i j} y_{i j} \frac{1}{\exp \left(\tau_{i j}\right)}\right\}. \qquad
\label{formula:ZIG-loglink-ll2}
\eqan
%where
%\bqa
%\log f\left(y_{i j} \mid y_{i j}>0\right)
%% &=& -\log T\left(v_{i j}\right)+v_{i j}\left(\log v_{i j}-\left(\omega_{i} T \widetilde{w}_{j}+e_{i}+\tilde{e}_{j}\right)\right)+\left(v_{i j}-1\right) \log y_{i j}-v_{i j} y_{i j} \frac{1}{\exp \left(\omega_{0} \widetilde{w}_{j}+e_{i}+\tilde{e}_{j}\right)} \\
%= -\log \Gamma\left(v_{i j}\right)+v_{i j} \log v_{i j}+\left(v_{i j}-1\right) \log y_{i j}-v_{i j} \cdot \tau_{i j}-v_{i j} y_{i j} \frac{1}{\exp \left(\tau_{i j}\right)}.
%\eqa
The equations below give the first order partial derivatives of $\ell_i^{(2)}$ with respect to the parameters $\vw_i$ and $e_i$.
\bqan
\begin{gathered}
\begin{aligned}
\label{formual:ZIG-loglink-partial1st-ll2}
&\frac{\partial \ell_{i}^{(2)}}{\partial \vw_{i}} = \sum_{j=1}^{n}-g_{i j}  v_{i j}\left\{\frac{\partial \tau_{i j}}{\partial \vw_{i}}+\frac{y_{i j}}{\mu_{i j}}\left(-\frac{\partial \tau_{i j}}{\partial \vw_{i}}\right)\right\}
% &=& \sum_{j=1}^{n}-g_{i j} \cdot v_{i j} \cdot \mu_{i j}^{-1}\left(\mu_{i j}-y_{i j}\right) \cdot \frac{\partial \tau_{i j}}{\partial \vw_{i}} \\
% &=& \sum_{j=1}^{n} g_{i j} \cdot v_{i j} \cdot \mu_{i j}^{-1}\left(y_{i j}-\mu_{i j}\right) \cdot \frac{\partial \tau_{i j}}{\partial \vw_{i}} \\
= \sum_{j=1}^{n} g_{i j}  v_{i j}  \mu_{i j}^{-1}\left(y_{i j}-\mu_{i j}\right)  \wtvw_{j}, \\
&\frac{\partial \ell_{i}^{(2)}}{\partial e_{i}} = \sum_{j=1}^{n} g_{i j}  v_{i j}  \mu_{i j}^{-1}\left(y_{i j}-\mu_{i j}\right) .
\end{aligned}
\end{gathered}
\eqan
To get the second order partial derivatives, first note that %frequently used two partial derivatives are
\bqa
\mu_{i j}=\exp \left(\tau_{i j}\right) \quad \Rightarrow \frac{\partial \mu_{i j}}{\partial \vw_{i}}=\frac{\partial \mu_{i j}}{\partial \tau_{i j}}  \frac{\partial \tau_{i j}}{\partial \vw_{i}}=\mu_{i j}  \widetilde{\vw}_{j}, \quad
\text{ and } \quad \frac{\partial \mu_{i j}}{\partial e_{i}}=\frac{\partial \mu_{i j}}{\partial \tau_{i j}} \frac{\partial \tau_{i j}}{\partial e_{i}}=\mu_{i j}.
\eqa
Then the second order partial derivatives and their negative expectations which are components of the Fisher information matrix can be derived as

\bqan
\label{formual:ZIG-loglink-partial2nd-ll2}
\frac{\partial^{2} \ell_{i}^{(2)}}{\partial \vw_{i} \partial \vw_{i}^{\top}}
%=\sum_{j=1}^{n} g_{i j}  v_{i j}\left\{-\mu_{i j}^{-2}  \mu_{i j}  \wtvw_{j}\left(y_{i j}-\mu_{i j}\right)-\wtvw_{j}\right\} \wtvw_{j}^{\top}
%\\&
%\,\, \qquad \qquad
\!=\! - \!\!\sum_{j=1}^{n} g_{i j}  v_{i j}\! \left[\!\frac{y_{i j}\!-\!\mu_{i j}}{\mu_{i j}}\!+\! 1\! \right] \wtvw_{j} \wtvw_{j}^{\top} \,
%\\ &
\Rightarrow
 %\Longrightarrow
 E\!\left[\!\left. \frac{-\partial^{2} \ell_{i}^{(2)}}{\partial \vw_{i} \partial \vw_{i}^{\top}} \right|g_{i j}\!=\!1 \!\right] \!=\!\! \sum_{j=1}^{n} g_{i j} v_{i j}  \wtvw_{j} \wtvw_{j}^{\top}, \qquad && \label{l2_1}\\
 \frac{\partial^{2} \ell_{i}^{(2)}}{\partial \vw_{i} \partial e_{i}}
%=\sum_{j=1}^{n} g_{i j} \cdot v_{i j} \cdot\left\{-\mu_{i j}^{-2} \cdot \mu_{i j} \cdot 1 \left(y_{i j}-\mu_{i j}\right)-\mu_{i j}^{-1}\left(\mu_{i j} \cdot 1\right)\right\} \wtvw_{j} \\
%& \,\,\,\, \quad \qquad
=-\!\!\sum_{j=1}^{n} g_{i j}  v_{i j}\!\left[\frac{y_{i j}\!-\!\mu_{i j}}{\mu_{i j}}\!+\!1\right] \wtvw_{j}
\,
%\\&
\Longrightarrow E\!\left[\left. -\frac{\partial^{2} \ell_{i}^{(2)}}{\partial \vw_{i} \partial e_{i}} \right| g_{i j}\!=\!1\right]\!=\!\sum_{j=1}^{n} g_{i j}  v_{i j}  \wtvw_{j},\qquad\qquad &&\label{l2_2} \\
 \frac{\partial^{2} \ell_{i}^{(2)}}{\partial e_{i}^{2}}
%=\sum_{j=1}^{n} g_{i j} \cdot v_{i j}\left\{-\mu_{i j}^{-2} \cdot \mu_{i j}\left(y_{i j}-\mu_{i j}\right)-\mu_{i j}^{-1} \cdot \mu_{i j} \cdot 1\right\} \cdot 1 \\
%& \quad \qquad
=-\sum_{j=1}^{n} g_{i j}  v_{i j}\left[\frac{y_{i j}-\mu_{i j}}{\mu_{i j}}+1\right] \,
%\\&
 \Longrightarrow E\left[\left. -\frac{\partial^{2} \ell_{i}^{(2)}}{\partial e_{i}^{2}} \right| g_{i j}=1\right]=\!\sum_{j=1}^{n} g_{i j}  v_{i j}.\qquad \qquad\qquad  &&\label{l2_3}
\eqan

The first part of the alternating regression has the following updating equation based on the Fisher scoring algorithm:
\bqan
\nonumber
&& \vtheta_{i}^{(t+1)}\!=\!\vtheta_{i}^{(t)}+S_{\vtheta_{i}^{(t)}}^{-1} U_{\vtheta_{i}^{(t)}}, \quad i\!=\!1, \ldots, n,  \\ %\qquad
\label{formula:ZIG-loglink-update-epoch}
&& \vtheta_{i}^{(t, k+1)}\!=\!\vtheta_{i}^{(t, k)}+S_{\vtheta_{i}^{(t, k)}}^{-1} U_{\vtheta_{i}^{(t, k)}}, \quad k\!=\!1, \ldots, E.
\eqan
The updating equation in (\ref{formula:ZIG-loglink-update-epoch}) is iterated E epochs, say E equals 20, for the same $i$ and $t$ where $t$ is the iteration number.
%That is, for each iteration of the update, the estimation of $\vtheta_i$ is iteratively updated for 20 epochs.
The multiple epochs here allow the estimate of $\vtheta_i$ to get closer to its MLE for given current value $\wtvtheta^{(c)}$ as the score equations and information matrix were also updated with the estimated parameter value. There is no need to have too many epochs because the parameter $\wtvtheta^{(c)}$ is not the true value yet and still need to be estimated later. Below are the formulae involved in the updating equations:
\bqa
\begin{array}{ccc}
\vtheta_{i}^{(t)} \!=\!\left(\vw_{i}^{\top(t)}, b_{i}^{(t)}, e_{i}^{(t)}\right)^{\top}\!; & \, \left.U_{\vtheta_{i}^{(t)}} \!=\!\left[\left(\frac{\partial \ell_{i}}{\partial \vw_{i}}\right)^\top, \frac{\partial \ell_{i}}{\partial b_{i}}, \frac{\partial \ell_{i}}{\partial e_{i}}\right]^{\top}\right|_{\vtheta_{i}^{(t)}}\!;
&
\, S_{\vtheta_i^{(t)}} \!=\! \left[\begin{array}{lll}
S_{\vw_{i}}^{(t)} & S_{\vw_{i} b_{i}}^{(t)} & S_{\vw_{i} e_{i}}^{(t)} \\
S_{b_{i} \vw_{i}}^{(t)} & S_{b_{i}}^{(t)} & S_{b_{i} e_{i}}^{(t)} \\
S_{e_{i} \vw_{i}}^{(t)} & S_{e_{i} b_{i}}^{(t)} & S_{e_{i}}^{(t)}
\end{array}\right]
\end{array}
\eqa
with
%\vspace{-0.2in}
% and $\frac{\partial \ell_{i}}{\partial \vw_{i}}, \frac{\partial \ell_{i}}{\partial b_{i}}, \frac{\partial \ell_{i}}{\partial e_{i}}$, and elements of $S_{\vtheta_i^{(t)}}$ are functions of $\vtheta_i^{(t)}, \quad \wtvtheta^{(t)} \!=\! (\wtvtheta_1^{(t)\top}, \ldots, \wtvtheta_n^{(t)\top})^\top$:
%\bqa
%&& \left.\frac{\partial \ell_{i}}{\partial \vw_{i}}\right|_{\vtheta_{i}^{(t)}} = \frac{\partial \ell_{i}^{(1)}(\vtheta_i^{(t)}; \wtvtheta^{(c)})}{\partial \vw_{i}} + \frac{\partial \ell_{i}^{(2)}(\vtheta_i^{(t)}; \wtvtheta^{(c)})}{\partial \vw_{i}}, \\
%&& \left. \frac{\partial \ell_{i}}{\partial b_{i}}\right|_{\vtheta_{i}^{(t)}}=\frac{\partial \ell_{i}^{(1)}(\vtheta_i^{(t)}; \wtvtheta^{(c)})}{\partial b_{i}}, \qquad \left.\frac{\partial \ell_{i}}{\partial e_{i}}\right|_{\vtheta_{i}^{(t)}}=\frac{\partial \ell_{i}^{(2)}(\vtheta_i^{(t)}; \wtvtheta^{(c)})}{\partial e_{i}},
%\eqa
\bqa
 \left.\frac{\partial \ell_{i}}{\partial \vw_{i}}\right|_{\vtheta_{i}^{(t)}} \!=\! \sum_{k=1}^2 \frac{\partial \ell_{i}^{(k)}(\vtheta_i^{(t)}; \wtvtheta^{(c)})}{\partial \vw_{i}},
 %\!+\! \frac{\partial \ell_{i}^{(2)}(\vtheta_i^{(t)}; \wtvtheta^{(c)})}{\partial \vw_{i}},
 \quad
 \left. \frac{\partial \ell_{i}}{\partial b_{i}}\right|_{\vtheta_{i}^{(t)}}\!=\!\frac{\partial \ell_{i}^{(1)}(\vtheta_i^{(t)}; \wtvtheta^{(c)})}{\partial b_{i}}, \qquad \left.\frac{\partial \ell_{i}}{\partial e_{i}}\right|_{\vtheta_{i}^{(t)}}\!=\!\frac{\partial \ell_{i}^{(2)}(\vtheta_i^{(t)}; \wtvtheta^{(c)})}{\partial e_{i}},
\eqa
where these partial derivatives are $\frac{\partial \ell_{i}^{(1)}}{\partial \vw_{i}}$, $\frac{\partial \ell_{i}^{(2)}}{\partial \vw_{i}}$, $\frac{\partial \ell_{i}^{(1)}}{\partial b_{i}}$, and $\frac{\partial \ell_{i}^{(2)}}{\partial e_{i}}$ evaluated at $\vtheta_i^{(t)}$ and $\wtvtheta^{(c)}$ using equations %(\ref{formual:ZIG-loglink-partial1st-ll1}),
(\ref{formula:ZIG-canonicallink-1stpartial-ll1}),
 (\ref{formual:ZIG-loglink-partial1st-ll2}), and
\bqa
&& S_{\vw_{i}}^{(t)}=E\left[-\frac{\partial^{2} \ell_{i}^{(1)}(\vtheta_i^{(t)}; \wtvtheta^{(c)})}{\partial \vw_{i} \partial \vw_{i}^{\top}}\right]+E\left[\left. -\frac{\partial^{2} \ell_{i}^{(2)}(\vtheta_i^{(t)}; \wtvtheta^{(c)})}{\partial \vw_{i} \partial \vw_{i}^{\top}}\right| g_{i j}=1\right] ; \\
&& S_{\vw_{i} b_{i}}^{(t)}=E\left[-\frac{\partial^{2} \ell_{i}^{(1)}(\vtheta_i^{(t)}; \wtvtheta^{(c)})}{\partial \vw_{i} \partial b_{i}}\right]; \qquad S_{b_{i}}^{(t)}=E\left[-\frac{\partial^{2} \ell_{i}^{(1)}(\vtheta_i^{(t)}; \wtvtheta^{(c)})}{\partial b_{i}^{2}}\right] ; \qquad S_{b_{i} e_{i}}^{(t)}=0 ; \\
&& S_{\vw_{i} e_{i}}^{(t)}=E\left[\left. -\frac{\partial^{2} \ell_{i}^{(2)}(\vtheta_i^{(t)}; \wtvtheta^{(c)})}{\partial \vw_{i}  \partial e_{i}}\right|g_{i j}=1 \right]; \qquad
S_{e_{i}}^{(t)} = E\left[\left. -\frac{\partial^{2} \ell_{i}^{(2)}(\vtheta_i^{(t)}; \wtvtheta^{(c)})}{\partial e_{i}^2}\right|g_{i j}=1; \right].
\eqa
The expectations involved in above formulae were given in (\ref{formual:ZIG-loglink-partial2nd-ll1-0}) - (\ref{formual:ZIG-loglink-partial2nd-ll1-2}) %(\ref{formual:ZIG-loglink-partial2nd-ll1}),
and (\ref{formual:ZIG-loglink-partial2nd-ll2})-(\ref{l2_3}) except that the parameters are using the current values $\vtheta^{(t)}$ and $\wtvtheta^{(c)}$ at $t^{th}$ iteration if it is in outer loop of update and using the values of $\vtheta^{(t, k)}$ and $\wtvtheta^{(c)}$ when it is in the inner loop of updates.

The aforementioned equations are used in updating $\vtheta$ part of the alternating ZIG regression. Now, consider the other side of the alternating ZIG regression in which updates are done for $\wtvtheta$ while holding $\vtheta$ fixed. First, consider the first order partial derivatives
\bqa
&& \frac{\partial \widetilde{\ell}}{\partial \wtvw_{j}} = \sum_{i=1}^n\left\{\frac{\partial \widetilde{\ell}_{i j}^{(1)}}{\partial \wtvw_{j}}+\frac{\partial \widetilde{\ell}_{i j}^{(2)}}{\partial \wtvw_{j}}\right\} = \frac{\partial \sum_{i=1}^{n} \widetilde{\ell}_{i j}^{(1)}}{\partial \wtvw_{j}}+\frac{\partial \sum_{i=1}^{n} \widetilde{\ell}_{i j}^{(2)}}{\partial \wtvw_{j}}=\frac{\partial \widetilde{\ell}_{j}^{(1)}}{\partial \wtvw_{j}}+\frac{\partial \widetilde{\ell}_{j}^{(2)}}{\partial \wtvw_{j}}.
\eqa
The first order partial derivatives of $\widetilde{\ell}_j^{(1)}$ are
\bqa
\frac{\partial \widetilde{\ell}_{j}^{(1)}}{\partial \wtvw_{j}} = \sum_{i=1}^{n}\left(g_{i j}-p_{i j}\right)  \vw_{i}, \quad
\frac{\partial \widetilde{\ell}_{j}^{(1)}}{\partial \wtb_{j}} = \sum_{i=1}^{n}\left(g_{i j}-p_{i j}\right), \quad
\frac{\partial \widetilde{\ell}_{j}^{(1)}}{\partial \wte_{j}} = 0.
\eqa
The second order partial derivatives
$\frac{\partial^{2} \widetilde{\ell}_{j}^{(1)}}{\partial \wtvw_{j} \partial \wtvw_{j}^{\top}}$,
$\frac{\partial^2 \widetilde{\ell}_{j}^{(1)}}{\partial \wtb_{j} \partial \wte_{j}}$,
$\frac{\partial^{2} \widetilde{\ell}_{j}^{(1)}}{\partial \wtb_{j}^{2}}$
and their expectations are same as in the canonical link case, which are given in equations (\ref{formual:ZIG-loglink-partial2nd-ll1-0}) - (\ref{formual:ZIG-loglink-partial2nd-ll1-2}). Additionally, \bqa
 \frac{\partial^2 \widetilde{\ell}_{j}^{(1)}}{\partial \wtb_{j} \partial \wte_{j}} = 0; \quad
\frac{\partial^{2} \widetilde{\ell}_{j}^{(1)}}{\partial \wte_{j}^{2}} = 0; \quad \Longrightarrow \quad E\left[-\frac{\partial^2 \widetilde{\ell}_{j}^{(1)}}{\partial \wtb_{j} \partial \wte_{j}}\right]=0; \quad E\left[-\frac{\partial^{2} \widetilde{\ell}_{j}^{(1)}}{\partial \wte_{j}^{2}}\right] = 0.
\eqa
%
%
%\bqa
%&& \frac{\partial^{2} \widetilde{\ell}_{j}^{(1)}}{\partial \wtvw_{j} \partial \wtvw_{j}^{\top}} = \sum_{i=1}^{n} \vw_{i} \cdot\left(-\frac{\partial p_{i j}}{\partial \wtvw_{j}^{\top}}\right) = - \sum_{i=1}^{n} \vw_{i} \vw_{i}^{\top} p_{i j}\left(1-p_{i j}\right), \\
%\Longrightarrow && E\left[-\frac{\partial^{2} \widetilde{\ell}_{j}^{(1)}}{\partial \wtvw_{j} \partial \wtvw_{j}^{\top}}\right] = \sum_{i=1}^{n} \vw_{i} \vw_{i}^{\top} p_{i j}\left(1-p_{i j}\right); \\
%&& \frac{\partial^2 \widetilde{\ell}_{j}^{(1)}}{\partial \wtb_{j} \partial \wte_{j}} = 0; \quad
%\frac{\partial^{2} \widetilde{\ell}_{j}^{(1)}}{\partial \wte_{j}^{2}} = 0; \quad \Longrightarrow \quad E\left[-\frac{\partial^2 \widetilde{\ell}_{j}^{(1)}}{\partial \wtb_{j} \partial \wte_{j}}\right]=0; \quad E\left[-\frac{\partial^{2} \widetilde{\ell}_{j}^{(1)}}{\partial \wte_{j}^{2}}\right] = 0;
%\eqa
%\bqa
%&& \frac{\partial^2 \widetilde{\ell}_{j}^{(1)}}{\partial \wtvw_{j} \partial \wtb_{j}}=-\sum_{i=1}^{n} \vw_{i} p_{i j}\left(1-p_{i j}\right) \quad
%\Longrightarrow \quad E\left[-\frac{\partial^{2} \widetilde{\ell}_{j}^{(1)}}{\partial \wtvw_{j} \partial b_{j}^{2}}\right] = \sum_{i=1}^{n} \vw_{i} p_{i j}\left(1-p_{i j}\right); \\
%&& \frac{\partial^{2} \widetilde{\ell}_{j}^{(1)}}{\partial \wtb_{j}^{2}}=-\sum_{i=1}^{n} p_{i j}\left(1-p_{i j}\right) \quad
%\Longrightarrow \quad E\left[-\frac{\partial^{2} \widetilde{\ell}_{j}^{(1)}}{\partial \wtb_{j}^{2}}\right] = \sum_{i=1}^{n} p_{i j}\left(1-p_{i j}\right).
%\eqa
Next, consider the first order partial derivatives of the second component $\widetilde{\ell}_j^{(2)}$
\bqa
\frac{\partial \widetilde{\ell}_{j}^{(2)}}{\partial \wtvw_{j}}=\sum_{i=1}^{n} g_{i j}  v_{i j}  \mu_{i j}^{-1}\left(y_{i j}-\mu_{i j}\right)  \vw_{i}, \quad
\frac{\partial \widetilde{\ell}_{j}^{(2)}}{\partial \wte_{j}}=\sum_{i=1}^{n} g_{i j}  v_{i j}  \mu_{i j}^{-1}\left(y_{i j}-\mu_{i j}\right) .
\eqa
To derive the second order partial derivatives, note that frequently used two terms are
\bqa
\mu_{i j}=\exp \left(\tau_{i j}\right) \quad
\Rightarrow \frac{\partial \mu_{i j}}{\partial \wtvw_{j}}=\frac{\partial \mu_{i j}}{\partial \tau_{i j}} \cdot \frac{\partial \tau_{i j}}{\partial \wtvw_{j}}=\mu_{i j}  \vw_{i} \quad \mbox{and} \quad
\frac{\partial \mu_{i j}}{\partial \tilde{e}_{j}}=\frac{\partial \mu_{i j}}{\partial \tau_{i j}} \cdot \frac{\partial \tau_{i j}}{\partial \wte_{j}}=\mu_{i j} .
\eqa
Using the two terms, the second derivatives can be written as
\bqa
\!\!&&\!\! \frac{\partial^{2} \ell_{i}^{(2)}}{\partial \wtvw_{j} \partial \wtvw_{j}^{\top}}\!=\!-\!\!\sum_{i=1}^{n} g_{i j}  v_{i j} \left[\frac{y_{i j}\!-\!\mu_{i j}}{\mu_{i j}}\!+\!1\right] \vw_{i} \vw_{i}^{\top}\,
%\\&&
 \Rightarrow E\left[\!\left. -\frac{\partial^{2} \ell_{i}^{(2)}}{\partial \wtvw_{j} \partial \wtvw_{j}^{\top}} \right| g_{i j}\!=\!1\!\right]\!=\!-\!\!\sum_{i=1}^{n} g_{i j}  v_{i j}  \vw_{i} \vw_{i}^{\top}, \\
\!\!&&\!\! \frac{\partial^{2} \ell_{i}^{(2)}}{\partial \wtvw_{j} \partial \wte_{j}}=\!-\!\sum_{i=1}^{n} g_{i j} v_{i j}\left[\frac{y_{i j}-\mu_{i j}}{\mu_{i j}}\!+\!1\right] \vw_{i} \quad
\Rightarrow \quad E\left[\left. \!-\!\frac{\partial^{2} \ell_{i}^{(2)}}{\partial \wtvw_{j} \partial \wte_{j}} \right| g_{i j}=1\right]=\sum_{i=1}^{n} g_{i j}  v_{i j}  \vw_{i}, \\
\!\!&&\!\! \frac{\partial^{2} \ell_{i}^{(2)}}{\partial \wte_{j}^{2}}\!=\!-\sum_{i=1}^{n} g_{i j} v_{i j}\left[\frac{y_{i j}-\mu_{i j}}{\mu_{i j}}\!+\!1\right]  \quad
 \Rightarrow \quad E\left[\left. -\frac{\partial^{2} \ell_{i}^{(2)}}{\partial \wte_{j}^{2}} \right| g_{i j}\!=\!1\right]\!=\!\sum_{i=1}^{n} g_{i j}  v_{i j}.
\eqa

Therefore, the updating equations for $\wtvtheta$ %the alternating ZIG regression
based on the Fisher scoring algorithm are
\bqan
\label{ZIG-loglink-update-tilde-it}
&&\wtvtheta_{j}^{(t+1)}=\wtvtheta_{j}^{(t)}+S_{\wtvtheta_{j}^{(t)}}^{-1} U_{\wtvtheta_{j}^{(t)}}, \quad j=1,\ldots,n, \\
\label{ZIG-loglink-update-tilde-epoch}
&&\wtvtheta_{j}^{(t, k+1)}=\wtvtheta_{j}^{(t, k)}+S_{\wtvtheta_{j}^{(t, k)}}^{-1} U_{\wtvtheta_{j}^{(t, k)}}, \quad k=1,\ldots,E.
\eqan
The updating equation (\ref{ZIG-loglink-update-tilde-it}) is in the outer loop of iterations and the iterations in (\ref{ZIG-loglink-update-tilde-epoch}) are in the inner loop of epochs for the same $j$ and $t$. The quantities involved are
\bqa
\begin{array}{ccc}
\wtvtheta_{j}^{(t)} =\left(\wtvw_{j}^{\top (t)}, \wtb_{j}^{(t)}, \wte_{j}^{(t)}\right)^{\top}\!; & \!\!\!\!\quad \left. U_{\wtvtheta_{j}^{(t)}} =\left[\begin{array}{c}\frac{\partial \widetilde{\ell}_{j}}{\partial \wtvw_{j}} \\
 \frac{\partial \widetilde{\ell}_{j}}{\partial \wtb_{j}}\\ \frac{\partial \widetilde{\ell}_{j}}{\partial \wte_{j}}
\end{array}
\right]\right|_{\wtvtheta_{j}^{(t)}} \!;
&
\!\!\!\!\quad S_{\wtvtheta_j^{(t)}} = \left[\begin{array}{lll}
S_{\wtvw_{j}}^{(t)} & S_{\wtvw_{j} \wtb_{j}}^{(t)} & S_{\wtvw_{j} \wte_{j}}^{(t)} \\
S_{\wtb_{j} \wtvw_{j}}^{(t)} & S_{\wtb_{j}}^{(t)} & S_{\wtb_{j} \wte_{j}}^{(t)} \\
S_{\wte_{j} \wtvw_{j}}^{(t)} & S_{\wte_{j} \wtb_{j}}^{(t)} & S_{\wte_{j}}^{(t)}
\end{array}\right]
\end{array}
\eqa
with
%\bqa
%&& \frac{\partial \widetilde{\ell}_{j}}{\partial \wtvw_{j}}\!=\!\frac{\partial \widetilde{\ell}_{j}^{(1)}(\wtvtheta_j^{(t)}; \vtheta^{(t)})}{\partial \wtvw_{j}} + \frac{\partial \widetilde{\ell}_{j}^{(2)}(\wtvtheta_j^{(t)}; \vtheta^{(t)})}{\partial \wtvw_{j}},  \\
%&& \frac{\partial \widetilde{\ell}_{j}}{\partial \wtb_{j}}=\frac{\partial \widetilde{\ell}_{j}^{(1)}(\wtvtheta_j^{(t)}; \vtheta^{(t)})}{\partial \wtb_{j}}, \qquad \frac{\partial \widetilde{\ell}_{j}}{\partial \wte_{j}}=\frac{\partial \widetilde{\ell}_{j}^{(2)}(\wtvtheta_j^{(t)}; \vtheta^{(t)})}{\partial \wte_{j}},
%\eqa
\bqa
 \left.\frac{\partial \widetilde{\ell}_{j}}{\partial \wtvw_{j}}\right|_{\wtvtheta_{j}^{(t)}} \!=\!\sum_{k=1}^2 \frac{\partial \widetilde{\ell}_{j}^{(k)}(\wtvtheta_j^{(t)}; \vtheta^{(c)})}{\partial \wtvw_{j}},  \qquad
 \left.\frac{\partial \widetilde{\ell}_{j}}{\partial \wtb_{j}}\right|_{\wtvtheta_{j}^{(t)}} \!=\!\frac{\partial \widetilde{\ell}_{j}^{(1)}(\wtvtheta_j^{(t)}; \vtheta^{(c)})}{\partial \wtb_{j}}, \qquad \left.\frac{\partial \widetilde{\ell}_{j}}{\partial \wte_{j}}\right|_{\wtvtheta_{j}^{(t)}} \! =\!\frac{\partial \widetilde{\ell}_{j}^{(2)}(\wtvtheta_j^{(t)}; \vtheta^{(c)})}{\partial \wte_{j}},
\eqa
where these partial derivatives are derivatives $\frac{\partial \widetilde{\ell}_{j}^{(1)}}{\partial \wtvw_{j}}$, $\frac{\partial \widetilde{\ell}_{j}^{(2)}}{\partial \wtvw_{j}}$, $\frac{\partial \widetilde{\ell}_{j}^{(1)}}{\partial \wtb_{j}}$,$\frac{\partial \widetilde{\ell}_{j}^{(2)}}{\partial \wte_{j}}$ and
\bqa
&& S_{\wtvw_{j}}^{(t)}=E\left[-\frac{\partial^{2} \widetilde{\ell}_{j}^{(1)}(\wtvtheta_j^{(t)}; \vtheta^{(c)})}{\partial \wtvw_{j} \partial \wtvw_{j}^{\top}}\right] + E\left[\left.-\frac{\partial^{2} \widetilde{\ell}_{j}^{(2)}(\wtvtheta_j^{(t)}; \vtheta^{(c)})}{\partial \wtvw_{j} \partial \wtvw_{j}^{\top}}\right| g_{i j}=1 \right] ;\\
&& S_{\wtvw_{j} \wtb_{j}}^{(t)}=E\left[-\frac{\partial^{2} \widetilde{\ell}_{j}^{(1)}(\wtvtheta_j^{(t)}; \vtheta^{(c)})}{\partial \wtvw_{j} \partial \wtb_{j}}\right] ; \quad
S_{\wtb_{j}}^{(t)}=E\left[-\frac{\partial^{2} \widetilde{\ell}_{j}^{(1)}(\wtvtheta_j^{(t)}; \vtheta^{(c)},)}{\partial \wtb_{j}^{2}}\right] ; \quad
S_{\wtb_{j} \wte_{j}}^{(t)}=0 ; \\
&& S_{\wtvw_{j} \wte_{j}}^{(t)}=E\left[\left.-\frac{\partial^{2} \widetilde{\ell}_{j}^{(2)}(\wtvtheta_j^{(t)}; \vtheta^{(c)})}{\partial \wtvw_{j}  \partial \wte_{j}}\right| g_{i j}=1 \right]; \quad
S_{\wte_{j}}^{(t)} = E\left[\left.-\frac{\partial^{2} \widetilde{\ell}_{j}^{(2)}(\wtvtheta_j^{(t)}; \vtheta^{(c)})}{\partial \wte_{j}^2}\right| g_{i j}=1 \right],
\eqa
evaluated at $\wtvtheta^{(t)}$ and $\vtheta^{(c)}$ during outer loop of iterations and at $\wtvtheta^{(t, k)}$ and $\vtheta^{(c)}$ during inner loop of iterations.

\section{Convergence analysis}
\label{section:convergence}

In this section, we discuss the convergence behavior of the algorithm analytically. Our algorithm contains two components: logistic regression part and Gamma regression part. If these two components' parameter estimation were independent of each other, then it is the situation of standard Generalized Linear Model (GLM) in each part. In our model, the two components' estimation can not be separated but the two components in the log likelihood are additive. Hence, the convergence behavior in one component standard GLM case is still relevant. {\modify In this section, we first talk} about the convergence behavior in the standard case as this case applies to the situation when we hold the $\wtvtheta$ fixed while estimating $\vtheta$ or vice versa.

In the standard GLM setting, most of the parameter estimation converges pretty fast. There are also abnormal behavior that could happen such as when estimated model component gets out of the valid range of the distribution. For example, in the Gamma regression, if the canonical link (negative inverse) is used, the estimated mean may become negative every now and then even though the distribution requires positive mean. \cite{Haberman:1977}, \cite{MLE-existence-LogisticReg-Binomial:1981}, and \cite{MLE-existence-LogisticReg-Multinomial:1984} presented conditions for the existence of MLE in logistic regression models. They proved in the non-trivial case that if there is overlap in the convex cones generated by the covariate values from different classes, the maximum likelihood estimates of the regression parameters exist and are unique. On the other hand, if the convex cones generated by the covariates in different classes have complete separation or quasi complete separation, then the maximum likelihood estimates do not exist or unbounded. Generally, they recommended to insert a stopping rule if complete separation is found or restart the iterative algorithm with standardized observations (to have mean 0 and variance 1) if quasi complete separation is found. \cite{MLE-existence-LogisticReg-Multinomial:1984} also recommended a procedure to check the conditions. For quasi complete separation, the estimation process diverges at least at some points. This makes the estimated probability of belonging to the correct class grows to one. Therefore, \cite{MLE-existence-LogisticReg-Multinomial:1984} recommended to check the maximum predicted probability $p^{(t)}$ for each data point at $t^{th}$ iteration. If the maximum probability is close to 1 and is bigger than previous iterations' maximum probability, there are two possibilities that this could happen. %\cite{MLE-existence-LogisticReg-Multinomial:1984} recommended
They suggested
to initially print a warning but continue the iteration because the data point is likely to be an outlier observation in its own class and there is overlap in the two convex cones. In this case, the MLE exists and is unique so the algorithm should be allowed to continue. The other possibility is that there is quasi complete separation in the data. In this case, the process should be stopped and rerun with the observation vectors standardized with zero mean and unit variance.

\cite{MLE-existence-LogisticReg-Multinomial:1984} also stated that the  difficulties  associated  with  complete  and  quasi complete  separation  are  small
sample  problems.  With  large  sample  size,  the  probability  of  observing  a  set  of
separated  data  points  is close  to  zero. Complete
separation  may  occur  with  any  type  of  data  but  it  is  unlikely  that  quasi complete
separation will occur with  truly  continuous  data.

\cite{convergence-GLM:2011} illustrated the non-convergence problem with Poisson regression example. The author proposed a simple solution and implemented it in the R $glm2$ package. Specifically, if the iteratively reweighted least squares (IRLS) procedure produces either an infinite deviance or predicted values which fall within invalid range, then the amount of update of the parameter estimates is repeatedly halved until the update no longer shows the behavior. Moreover, it made a further step-halving, which checks the updated deviance is making a reduction compared to that in the previous iteration. If it did not show the reduction, it triggers the step-halving to make the algorithm monotonically reduces the deviance.

Based on the aforementioned studies illustrating the standard case of GLM convergence behavior, we {\modify  analytically present} the convergence behavior of our alternating ZIG regression.
For further discussion, we first specify our convex cones. Recall, $\wtvtheta$ serves as data when we estimate $\vtheta_i$. In our context, the convex cones {\modify are} defined through $\wtvtheta$ while we estimate $\vtheta_i$ and through $\vtheta$ while we estimate $\wtvtheta_j$. That is, for estimating $\vtheta_i$, the convex cones are
\bqan
\label{two-convex-cones}
G_{\wtvtheta}^{(i)}=\left\{\sum_{\substack{j=1 \\ y_{i j} > 0}}^{n} k_{j} \wtvtheta_j \mid k_{j}>0\right\}, \quad F_{\wtvtheta}^{(i)}=\left\{\sum_{\substack{j=1 \\ y_{i j} = 0}}^{n} k_{j} \wtvtheta_j \mid k_{j}>0\right\} .
\eqan
For estimating the $\wtvtheta_j$, the convex cones are denoted as $G_{\vtheta}^{(i)}$ and $F_{\vtheta}^{(i)}$, respectively.

Firstly, we consider the case that either complete separation or quasi complete separation exists in the data. We discuss the estimation of $\vtheta_i$ while holding $\wtvtheta$ fixed.

According to the theorem, suppose %the condition $\Pi$ is not satisfied. Then
 there exists complete separation or quasi separation in $G_{\wtvtheta}^{(i)}$ and $F_{\wtvtheta}^{(i)}$. That is, $G_{\wtvtheta}^{(i)} \cap F_{\wtvtheta}^{(i)} = \varnothing$, $G_{\wtvtheta}^{(i)} \neq R^{d+2}$, and $F_{\wtvtheta}^{(i)} \neq R^{d+2}$. Also, suppose $G_{\wtvtheta}^{(i)} \neq \varnothing$ and $F_{\wtvtheta}^{(i)} \neq \varnothing$. Then there exists a vector direction $\vc_i$ such that $\wtvw_j^\top \vc_i \geq 0$ for $y_{i j}>0$ and $\wtvw_j^\top \vc_i \leq 0$ for $y_{i j}=0$. (Note: This vector can be taken to be the perpendicular direction to a vector that lies in between $G_{\wtvtheta}^{(i)}$ and $F_{\wtvtheta}^{(i)}$ but does not belong to either $G_{\wtvtheta}^{(i)}$ and $F_{\wtvtheta}^{(i)}$. )
Let $\ell_i^{(1)}(k)$ be the log likelihood of the Bernoulli part when the model parameters are updated toward direction $\vc_i$ by $k$ unit. That is, the original log likelihood $\ell_i^{(1)}$ and the updated $\ell_i^{(1)}(k)$ are as follows:
\begin{equation}
\begin{gathered}
\ell_i^{(1)}=\sum_{j=1}^n\left\{I\left(y_{i j}=0\right) \cdot \log \left(1-p_{i j}\right)+I\left(y_{i j}>0\right) \cdot \log p_{i j}\right\} \\
\text { where } p_{i j}=\frac{\exp \left(\eta_{i j}\right)}{1+\exp \left(\eta_{i j}\right)}, \quad \eta_{i j}=\vw_i^\top \wtvw_j+b_i+\wtb_j \\
\ell_i^{(1)}(k)=\sum_{j=1}^n\left\{I\left(y_{i j}=0\right) \cdot \log \left(1-p_{i j}(k)\right)+I\left(y_{i j}>0\right) \cdot \log p_{i j}(k)\right\} \\
\text { where } p_{i j}(k)=\frac{\exp \left(\eta_{i j}+k \wtvw_j^{\top} \vc_i\right)}{1+\exp \left(\eta_{i j}+k \wtvw_j^{\top} \vc_i\right)}, \quad \eta_{i j}=\vw_i^{\top} \wtvw_j+b_i+\wtb_j
\end{gathered}
\end{equation}
\bit
\item For $y_{i j}>0, \wtvw_j^{\top} \vc_i \geq 0$. Hence, as $k$ increases to $\infty$, $p_{i j}(k)$ increases toward $1$. This implies $\sum_{j=1}^n I\left(y_{i j}>0\right) \cdot \log p_{i j}(k)$ increases toward 0.
\item For $y_{i j}=0$, $\wtvw_j^{\top} \vc_i \leq 0$. Hence, as $k$ increases to $\infty$, $p_{i j}(k)$ decreases toward $0$. This implies $\sum_{j=1}^n I\left(y_{i j}=0\right) \cdot \log \left(1-p_{i j}(k)\right)$ increases toward $0$.
\eit
Putting the two pieces together, we know that as $k$ increases, $\ell_i^{(1)}(k)$ increases for any given $\vw_i, b_i$. Therefore, the maximum can not be reached until $k$ is $\infty$. This means the MLE does not exist or the solution set $\{\hat{\vtheta}_i \}$ is unbounded for the current value of $\wtvtheta$. This perspective can be also seen by looking at the partial derivative of $\ell_i^{(1)}(k)$ with respect to $k$. Note that
\bqan
\frac{\partial \ell_i^{(1)}(k)}{\partial k} =\sum_{j=1}^n\left\{I\left(y_{i j}=0\right) \cdot\left(-p_{i j}(k)\right) \wtvw_j^{\top} \vc_i\right\}+\sum_{j=1}^n \left\{I\left(y_{i j}>0\right)\left(1-p_{i j}(k)\right) \wtvw_j^{\top} \vc_i\right\}.
\label{formula:ell_i1_k }
\eqan
Since $\wtvw_j^\top \vc_i \leq 0$ when $y_{i j}=0$, we know the first term is non-negative. Similarly, $\wtvw_j^\top \vc_i \geq 0$ when $y_{i j}>0$ implies the second term is non-negative. Therefore, the partial derivative of $\ell_i^{(1)}(k)$ is non-negative. This indicates that the gradient of $\ell_i^{(1)}$ is positive unless $\wtvw_j^\top \vc_i = 0$. Therefore, there is no solution for $\ell_i^{(1)} = 0$ except the trivial solution $\vc_i = 0$ in Binomial regression alone. Of course, this is not exactly our case because we still have the Gamma regression component to be considered together.

Now consider the $\ell_i^{(2)}$ component with its first and second order derivatives with respect to $k$.
\bqa
\ell_i^{(2)}(k)
%&=&\sum_{j=1}^n I(y_{i j}\!>\!0)\left\{o_{i j}\!+\!v_{i j}\left[\log y_{i j}\!-\!\left(\tau_{i j}\!+\!k \wtvw_j^{\top} \vc_i\right)\right]\!-\!v_{i j} y_{i j} \exp \left(\!-\!\left(\tau_{i j}\!+\!k \wtvw_j^{\top} \vc_i\right)\right)\right\} \\
%&=&
=\sum_{j=1}^n I(y_{i j}\!>\!0)\left\{o_{i j}+v_{i j}\left[\log \frac{y_{i j}}{\exp \left(\tau_{i j}+k \wtvw_j^{\top} \vc_i\right)}\right]-v_{i j} \frac{y_{i j}}{\exp \left(\tau_{i j}+k \wtvw_j^{\top} \vc_i\right)}\right\},
\eqa
where $o_{i j}=- \Gamma \left(v_{i j}\right)+v_{i j} \log v_{i j}-\log y_{i j}$.
 Note that
%$\frac{\partial(\log t-t)}{\partial t}=\frac{1}{t}-1 \text { is decreasing in t }$
\bqa
\frac{\partial \ell_i^{(2)}(k)}{\partial k}
%&=& \sum_{j=1}^n g_{i j} v_{i j}\left\{-\wtvw_j^{\top} \vc_j-y_{i j} \exp \left[-\left(\tau_{i j}+k \wtvw_j^{\top} \vc_j\right)\right] \cdot\left(-\wtvw_j^{\top} \vc_j\right)\right\} \\
%&=&
=\sum_{j=1}^n I(y_{i j}\!>\!0) v_{i j}\wtvw_j^{\top} \vc_i\left[\frac{y_{i j}}{\exp \left(\tau_{i j}+k \wtvw_j^{\top} \vc_i\right)}-1\right]=
%\sum_{j=1}^n g_{i j} \sqrt{v_{i j}}\wtvw_j^{\top} \vc_i\left[\frac{y_{i j}- \exp \left(\tau_{i j}+k \wtvw_j^{\top} \vc_i\right)}{\exp \left(\tau_{i j}+k \wtvw_j^{\top} \vc_i\right)/ \sqrt{v_{i j}} }\right]
\sum_{j=1}^n I(y_{i j}\!>\!0) \wtvw_j^{\top} \vc_i   \sqrt{v_{i j}} u_{ij}(k),
\eqa
where $u_{ij}(k) = \frac{y_{i j}- \exp \left(\tau_{i j}+k \wtvw_j^{\top} \vc_i\right)}{\exp \left(\tau_{i j}+k \wtvw_j^{\top} \vc_i\right)/ \sqrt{v_{i j}} } $ is the standardized Gamma random variable that has mean 0 and variance 1 because the term $\exp \left(\tau_{i j}+k \wtvw_j^{\top} \vc_i\right)$ is the mean of $y_{i j}$ and $v_{i j}$ is the shape parameter. Further,
\bqa
\frac{\partial^2 \ell_i^{(2)}(k)}{\partial k^2}
%&=&\sum_{j=1}^n g_{i j} v_{i j} \wtvw_j^{\top} \vc_j\left[y_{i j} \exp \left(-\tau_{i j}-k \wtvw_j^{\top} \vc_j\right) \cdot\left(-\wtvw_j^{\top} \vc_j\right)\right]\\
%&=&
= -\sum_{j=1}^n I(y_{i j}\!>\!0) v_{i j}\left(\wtvw_j^{\top} \vc_i\right)^2 \cdot \frac{y_{i j}}{\exp \left(\tau_{i j}+k \wtvw_j^{\top} \vc_i\right)}<0.
\eqa
The negativity of the second order derivative implies that $\ell_i^{(2)}(k)$ is concave.  %Hence, $\ell_i^{(2)}(k)$ is maximized when $y_{i j} = \exp \left(\tau_{i j}+k \wtvw_j^{\top} \vc_i\right)$.
%The first order partial derivative can be split into two summations: one is over the observed $y_{i j}$ that are less than its mean, and the other one is over the observed $y_{i j}$ that are greater than its mean.
%, i.e.,
%\bqan
%\label{formual:ZIG-loglink-ll2-1}
%\hspace{-0.3in}\frac{\partial \ell_i^{(2)}(k)}{\partial k} &=& \sum_{j=1}^n g_{i j} v_{i j}\wtvw_j^{\top} \vc_i\!\left[\!\frac{y_{i j}}{\exp \left(\tau_{i j}\!+\!k \wtvw_j^{\top} \vc_i\right)}\!-\!1\!\right]\! I\left(y_{i j} \!>\! \exp \left(\tau_{i j}\!+\!k \wtvw_j^{\top} \vc_i\right)\right) \\
%\label{formual:ZIG-loglink-ll2-2}
%&+& \sum_{j=1}^n g_{i j} v_{i j}\wtvw_j^{\top} \vc_i\!\left[\!\frac{y_{i j}}{\exp \left(\tau_{i j}\!+\!k \wtvw_j^{\top} \vc_i\right)}\!-\!1\!\right]\! I\left(y_{i j} \!<\! \exp \left(\tau_{i j}\!+\!k \wtvw_j^{\top} \vc_i\right)\right).
%\eqan
%The summation term in (\ref{formual:ZIG-loglink-ll2-1}) is positive while that in (\ref{formual:ZIG-loglink-ll2-2}) is negative.

Gathering the partial derivatives from both $\ell_i^{(1)}$ and $\ell_i^{(2)}$ together, we get the partial derivative of $\ell_i$
{\small
\bqan \nonumber
&&\frac{\partial \ell_i(k)}{\partial k}= \frac{\partial \ell_i^{(1)}(k)}{\partial k} + \frac{\partial \ell_i^{(2)}(k)}{\partial k} \\
%&=&\sum_{j=1}^n\!\left\{I\!\left(y_{i j}\!=\!0\right) \cdot\left(-p_{i j}(k)\right) \wtvw_j^{\top} \vc_i\right\}+\sum_{j=1}^n\! \left\{I\!\left(y_{i j}\!>\!0\right)\left[1\!-\!p_{i j}(k)\!+\!\sqrt{v_{i j}} u_{ij}(k)\right] \wtvw_j^{\top} \vc_i\right\}. \qquad \quad \label{first_deriv}\\
&=&\sum_{j=1}^n\!- I\!\left(y_{i j}\!=\!0\right) p_{i j}(k) \wtvw_j^{\top} \vc_i \!+\!\sum_{j=1}^n\! I\!\left(y_{i j}\!>\!0\right)(1\!-\!p_{i j}(k)) \wtvw_j^{\top} \vc_i \!+\!\sum_{j=1}^n \! I\!\left(y_{i j}\!>\!0\right)\! \sqrt{v_{i j}} u_{ij}(k) \wtvw_j^{\top} \vc_i. \qquad \quad \label{first_deriv}
\eqan
}
From the above discussion, the first two terms in (\ref{first_deriv}) corresponding to $\frac{\partial \ell_i^{(1)}(k)}{\partial k}$ are greater than 0. The $u_{ij}(k)$ is centered at zero and has variance 1.
 %and $\frac{\partial \ell_i^{(2)}(k)}{\partial k}$ has one term positive and the other term negative.
 Therefore, in order for the MLE to exist, %the term in (\ref{formual:ZIG-loglink-ll2-2})
 the sum of $\sum_{i=1}^n I\!\left(y_{i j}\!>\!0\right) \sqrt{v_{i j}} u_{ij}(k) \wtvw_j^{\top} \vc_i I(u_{ij}(k)<0)$
 must cancel the total value of
 $\frac{\partial \ell_i^{(1)}(k)}{\partial k}$ and the positive term $\sum_{i=1}^n I\!\left(y_{i j}\!>\!0\right) \sqrt{v_{i j}} u_{ij}(k)\wtvw_j^{\top} \vc_i I(u_{ij}(k)>0)$. If the shape parameter $v_{ij}$ of Gamma distribution is small,  the distribution is highly skewed with a long right tail. In this case, there are more observations having values less than its mean. However, small shape parameter also makes $\sqrt{v_{i j}} u_{ij}(k)$ small such that the $\frac{\partial \ell_i^{(1)}(k)}{\partial k}$ may dominate and hence the entire partial derivative $\frac{\partial \ell_i(k)}{\partial k}$ is greater than zero.
 When the shape parameter is large, the Gamma random variables are approximately normally distributed. In this case, the observations are symmetrically located on either side of its mean. This {\modify makes} the number of observations satisfying $u_{ij}<0$ and $ u_{ij}>0$
 %term (\ref{formual:ZIG-loglink-ll2-1}) and (\ref{formual:ZIG-loglink-ll2-2})
 roughly equal. Consequently, the $\frac{\partial \ell_i^{(2)}(k)}{\partial k}$ is close to zero. Then there {\modify are no} extra values left to neutralize $\frac{\partial \ell_i^{(1)}(k)}{\partial k}$. As a result, regardless of whether the shape parameter $v_{ij}$ is large or small,  when complete separation or quasi complete separation holds, it is highly likely that the MLE does not exist. When the shape parameter $v_{ij}$ is intermediate such that the Gamma distribution is still skewed, the more values of Gamma observations less than its mean might allow $\sum_{i=1}^n I\!\left(y_{i j}\!>\!0\right) \sqrt{v_{i j}} u_{ij}(k) \wtvw_j^{\top} \vc_i I(u_{ij}(k)<0)$ to cancel all other positive terms. This is the case that there could be a solution for $\wtvtheta_i$.

Next, consider the case that there is overlap in the two convex cones  $G_{\wtvtheta}^{(i)}$ and $F_{\wtvtheta}^{(i)}$ %when we estimate the $\vtheta_i$
(i.e., there is neither complete separation nor quasi complete separation in the data).
Recall that the ZIG model is a two-parts model with Bernoulli part and Gamma part. The log likelihood function $\ell_i^{(1)}$ corresponding to the Bernoulli part %(\ref{formula:ZIG-loglink-ll1})
 can be shown to be strictly concave in $\vtheta_i = (\vw_i^\top, b_i)^\top$. This is because the first component $g_{i j}\eta_{i j} = g_{ij} (\vw_i^\top \wtvw_j + b_i + \wtb_j)$ in $\ell_i^{(1)}$ is an affine function of $\vtheta_i$ (see equation (\ref{li1}) for the expression of $\ell_i^{(1)}$),  which is both convex and concave, if we hold $\wtvw_j$ and $\wtb_j$ fixed. The second component $-\log\left\{1+\exp(\eta_{i j})\right\}$ is strictly concave as its second derivative is less than 0 as shown below
\bqa
%&& \frac{\partial }{\partial x}[-\log\left\{1+\exp(x)\right\}] = -\frac{\exp(x)}{1+\exp(x)}, \\
%&&
 \frac{\partial^2 }{\partial x^2}[-\log\left\{1+\exp(x)\right\}] = -\frac{\exp(x)}{\left[1+\exp(x)\right]^2} < 0.
\eqa
Thus, the log likelihood corresponding to the Bernoulli part $\ell_i^{(1)}$ is strictly concave. Now, consider the log likelihood corresponding to the Gamma part $\ell_i^{(2)}$ in (\ref{formula:ZIG-loglink-ll2}). We only need to consider the summation over the last two terms $-\nu_{i j}\tau_{i j}$ and $-\nu_{i j}y_{i j}\exp(-\tau_{i j})$ because the other terms do not involve the regression parameters, where $\tau_{i j}=\vw_i^\top \wtvw_j + e_i + \wte_j$. Note that $-\tau_{i j}$ is an affine function of $\vtheta_i$, which is both convex and concave and $\exp(g(x))$ is convex if $g(x)$ is convex (see \citealt{BoydConvex:2004}). This leads to the term $-\nu_{i j}y_{i j}\exp(-\tau_{i j})$ being strictly concave in $\vtheta_i$. Hence, we have the $\ell_i^{(2)}$ strictly concave in $\vtheta_i= (\vw_i^\top, b_i, e_i)^\top$.  Combining the two concave components $\ell_i^{(1)}$ and $\ell_i^{(2)}$, we know that the entire log likelihood for the $i^{th}$ row $\ell_i$ is a strictly concave function with respect to the parameter being estimated for any row $i$. Additionally, there is overlap in the two convex cones. Therefore, for any direction in the overlapping area of the convex cones, updating the parameter along that direction will lead to the two components of $\frac{\partial \ell_i^{(1)}(k)}{\partial k}$ in formula (\ref{formula:ell_i1_k }) be of opposite sign of each other.  In this case, $-\ell_i$ has an unique minimum when there is neither complete separation nor quasi complete separation in $G_{\wtvtheta}^{(i)}$ and $F_{\wtvtheta}^{(i)}$. Similar arguments apply when we estimate $\wtvtheta_j$ while holding $\vtheta$ fixed. That is, an unique MLE of $\wtvtheta_j$ exists when there is neither complete separation nor quasi complete separation in $G_{\vtheta}^{(j)}$ and $F_{\vtheta}^{(j)}$. This means, the alternating procedure will find the MLE of $\wtvtheta_j$ when $\vtheta$ is fixed and will find the MLE of $\vtheta_i$ when $\wtvtheta$ is fixed in this non-separation scenario.

{\modify
Next we consider the convergence behavior for estimation of $\vtheta$ without fixing the value of $\wtvtheta$.
For estimating $\vtheta$, consider the complete data $\vx = (\vy, \wtvtheta)$. The entire $\wtvtheta$ matrix is missing.
%In regular generalized linear models with response variable $y$ and $p$ observed covariates from $n$ observations,
Due to too many missing values, a logical approach is to regard the $\wtvtheta$ as randomly drawn from a distribution which has relatively few parameters. Assume the distribution of $\vx$ is in the exponential family.
%This could help us understand the convergence behavior of ZIG.

Let $\tilde{f}(\vx| \vtheta)$ be the unconditional density of the complete data $\vx= (\vy, \wtvtheta)$ and $K(\vx| \vy, \vtheta)$ be the conditional density
given $\vy$. Denote the marginal density of $\vy$ given $\vtheta$ as $g(\vy|\vtheta)$.
In the next few paragraphs we explain that the alternating updates in ZIG leads ultimately to a value of $\vtheta$ that maximizes
$\check{\ell}(\vtheta) = \log g(\vy| \vtheta)$.

For exponential families, the unconditional density $\tilde{f}(\vx|\vtheta)$ and conditional density $K(\vx| \vy, \vtheta) = \tilde{f}(\vx| \vtheta)/g(\vy| \vtheta)$ both have the same natural parameter $\vtheta$ and the same sufficient statistic $\vt(\vx)$ except that they are defined over different sample spaces $\Omega(\vx)$ versus $\Omega(\vy)$. We can write $ \tilde{f}(\vx| \vtheta)$ and $K(\vx|\vy, \vtheta)$ in general exponential family format as
\bqa  \tilde{f}(\vx| \vtheta) = \exp(\vtheta \vt(\vx)^T +c(\vx) )/a(\vtheta), \\
K(\vx|\vy, \vtheta) = \exp(\vtheta \vt(\vx)^T +c(\vx) )/b(\vtheta|\vy),
\eqa
where
$$
a(\vtheta) = \int_{\Omega(\vx)}  \exp(\vtheta \vt(\vx)^T +c(\vx) ) d\vx, \quad
b(\vtheta|\vy)= \int_{\Omega(\vy)}  \exp(\vtheta \vt(\vx)^T +c(\vx) ) d\vx.$$
Then $\check{\ell}(\vtheta) = -\log a(\vtheta) + \log b(\vtheta|\vy)$.
The first and second order derivatives of $\check{\ell}(\vtheta) $ is
\bqa
\frac{\partial \check{\ell}(\vtheta) }{\partial \vtheta} =
-\frac{\partial \log a(\vtheta)}{ \partial \vtheta} + \frac{\partial \log b(\vtheta|\vy)} {\partial \vtheta} =E(\vt(\vx)| \vy, \vtheta ) -
E(\vt(\vx)| \vtheta), \\
\frac{\partial^2 \check{\ell}(\vtheta) }{\partial \vtheta \partial \vtheta^T} =
-\frac{\partial^2 \log a(\vtheta)}{ \partial \vtheta \partial \vtheta^T} + \frac{\partial^2 \log b(\vtheta|\vy)} {\partial \vtheta \partial \vtheta^T} =
E[\Var(\vt(\vx)|\vy, \vtheta)|\vtheta ]
- \Var(\vt(\vx)|\vtheta),
\eqa
where $ E(\vt(\vx)| \vtheta] )$ and $\Var(\vt(\vx)|\vtheta)$  are the expectation and variance under the complete data likelihood from $\vy$ and $\wtvtheta$. And
$ E(\vt(\vx)|\vy, \vtheta] )$ and $\Var(\vt(\vx)|\vy, \vtheta)$  are the conditional expectation and variance of the sufficient statistics. $E[\Var(\vt(\vx)|\vy, \vtheta)|\vtheta ] $
  is the expected value of the conditional covariance matrix when $\vy$ has sampling density
  $g(\vy|\vtheta)$.  %
 %
 %and
 %is covariance matrix of the sufficient statistic $\vt$ under the complete data likelihood from $\vy$ and $\wtvtheta$,
 %
 %
The last equality of both equations assumes the order of expectation and derivative can be exchanged. The previous equation means the derivative of the log likelihood is the difference between the conditional and unconditional expectation of the sufficient statistics.

%The maximum likelihood estimator $\widehat{\vtheta}$ of $\vtheta$ is the solution of $\partial \check{\ell}( \vtheta)/\partial \vtheta =0$ if the solution is in the interior of the parameter space. Hence,
%\bqan
%\partial \log a(\vtheta)/ \partial \vtheta = \partial \log b(\vtheta|\vy)/\partial \vtheta. \label{eqa_em}
%\eqan
%Assuming that the order of expectation and derivative can exchange, then (\ref{eqa_em}) implies
%$$ E(\vt(\vx)| \vy, \vtheta )=
%E(\vt(\vx)| \vtheta). $$
%That is, the
%maximum likelihood estimator $\hat{\vtheta}$ of $\vtheta$ is the solution
%to the equations
%$$ E(\vt(\vy, \wtvtheta)| \vy, \vtheta)=
%E(\vt(\vy, \wtvtheta)| \vtheta). $$
Meanwhile, the updating equation based on Fisher scoring algorithm
can be written as %(see page 31 of \cite{Dempster77})
\bqa
\widehat{\vtheta}^{(k+1)} = \widehat{\vtheta}^{(k)} +
\left\{\Var(\vt(\vx)|\widehat{\vtheta}^{(k)}) -
E[\Var(\vt(\vx)|\vy, \widehat{\vtheta}^{(k)})|\widehat{\vtheta}^{(k)} ] \right\}^{-1} \left. \frac{\partial \check{\ell}(\vtheta) }{\partial \vtheta}\right|_{\vtheta = \widehat{\vtheta}^{(k)}}^T
  %[E(\vt(\vx)|\vy, \widehat{\vtheta}^{(k)}) - E(\vt(\vx)|\widehat{\vtheta}^{(k)}] )
  \\
  =\widehat{\vtheta}^{(k)} +
\Var^{-1}(E(\vt(\vx)|\vy, \widehat{\vtheta}^{(k)}) |\widehat{\vtheta}^{(k)})
[E(\vt(\vx)|\vy, \widehat{\vtheta}^{(k)}) -
   E(\vt(\vx)|\widehat{\vtheta}^{(k)}] ),
  \eqa
  where
  in the limit, $\widehat{\vtheta}^{(k+1)} = \widehat{\vtheta}^{(k)} = \vtheta^*$, for some $\vtheta^*$, which leads to $E(\vt(\vx)|\vy, \vtheta^{*}) -
   E(\vt(\vx)| \vtheta^* )$ or $\partial \check{\ell}(\vtheta)/\partial\vtheta =0 $ at $\vtheta^*$.

The complete data log likelihood based on
the joint distribution
of $\vx= (\vy, \wtvtheta)$ can be written as
$\log\tilde{f}(\vx| \vtheta) = \ell(\vtheta; \wtvtheta) + \log(P_{\wtvtheta}(\wtvtheta |\vtheta))$,
where $\ell(\vtheta; \wtvtheta)$ is defined in the end of section 2, and $P_{\wtvtheta}(\wtvtheta|\vtheta))$ is the probability density
function of $\wtvtheta$ given $\vtheta$. We know $P_{\wtvtheta}(\wtvtheta|\vtheta)$ is also
a member of the exponential family. Assume its sufficient statistics
are linear in $\wtvtheta$. Given $\wtvtheta$, the observed data likelihood $\ell(\vtheta; \wtvtheta)$ based on $\vy$ contains the bernoulli and gamma parts, in which the sufficient statistic for $\vtheta$ is linear in $\vy$.  Then
the sufficient statistics for the complete data problem are linear in
 the data $\vy$ and  $\wtvtheta$.
 In this case, calculating
$E(\vt(\vx)| \vy, \vtheta^{(k)})$
%so  that  the  estimation  step  in  the  EM  algorithm
is  equivalent to  a  procedure  which  first
fills  in  the individual data points for $\wtvtheta$ and then  computes
the  sufficient statistics using filled-in values. With the filled in value for $\wtvtheta$, the computation of the estimator for $\vtheta$ follows the usual
maximum likelihood principle. This results in iterative update of $\vtheta$ and $\wtvtheta$ back and forth.
Essentially, the problem is  a transformation  from an
assumed  parameter-vector  to another parameter-vector  that  maximized  the conditional expected likelihood.

One  of  the difficulties in the ZIG model  is  that  the  parametrization allows arbitrary orthogonal transformations on both $\vw$ and $\wtvw$ without
affecting the value of  the likelihood.
%ridges in the actual
%likelihood $g(y |\vtheta)$.
%When  there  are ridges in the complete data likelhood, the parameters  in the %complete-data model  are not  identifiable.
Even  for cases  where  the likelihood  for the  complete-data $(\vy, \wtvtheta)$
problem is concave, the likelihood  for the ZIG may not  be concave. Consequently, multiple solutions of the likelihood  equations can exist.
An example is a ridge of solutions
corresponding to orthogonal transformations of the parameters.
%The nonuniqueness of
%the specific basis

For complete-data problems in exponential family models with canonical links, the Fisher scoring algorithm is equivalent to the Newton-Raphson algorithm which has quadratic rate of convergence. This advantage is due to the fact that the second
derivative of the log-likelihood does not depend on the data.
In these cases, the Fisher scoring algorithm has quadratic rate of convergence when the starting values are near a maximum.  However, we do not have the complete data $\vy$ and $\wtvtheta$ when estimating $\vtheta$. Fisher scoring algorithms often fail to have quadratic convergence in incomplete-data problems  since the second derivative often does depend upon the data. Further, the scoring algorithm does not have the property of always increasing the likelihood. It could in some cases move toward a local maximum if the choice of starting values is poor.

}

In summary, we conclude that our alternating ZIG regression has an unique MLE in each side of the regressions {\modify when either $\vtheta$ or $\wtvtheta$ stay fixed} and the algorithm converges when there is overlap in data. The data refer to $\wtvtheta$ while we estimate $\vtheta_i$ and refer to $\vtheta$ while we estimate $\wtvtheta_j$. %More specifically,
When there is overlap in data, both $\ell_i^{(1)}$ and $\ell_i^{(2)}$ are concave functions with unique maximum.
%and the two terms in the first order partial derivative in $\ell_i^{(1)}$ and $\ell_i^{(2)}$ are of opposite sign.
%The case for Gamma regression part will always have opposite signs which allow a maximum to exist.
However, when there is complete separation or quasi complete separation, the alternating ZIG regression will fail to converge with high chance. This is because the first order partial derivative of $\ell_i^{(1)}$ is non-negative and $\ell_i^{(1)}$ increases with $k$. Even though the $\ell_i^{(2)}$ component is a well-behaved concave curve, the entire log likelihood, unfortunately, may not have MLE exist or the solution set may be unbounded especially when the Gamma observations have large or too small shape parameter. {\modify The overall convergence behaviour for estimating $\vtheta$ without holding $\wtvtheta$ fixed can treat $\wtvtheta$ as missing data. The alternating update ultimately find the maximum likelihood estimate of $\vtheta$ based on the sampling distribution of the observed matrix $\vy$ if the solution is in the interior of the parameter space. This requires the joint distribution of $\vy$ and $\wtvtheta$ to be in the exponential family with sufficient statistic that is linear in $\vy$ and $\wtvtheta$.}

\section{Adjusting parameter update with learning rate}
\label{section:lr_adjustment}
In the convergence analysis section, our discussion is based on holding either $\vtheta$ or $\wtvtheta$ fixed while estimating the other one. The Fisher scoring algorithm is a modified version of Newton's method. In general, the Newton's method solves $ g'(x)=0$ by numerical approximation. This algorithm starts with an initial value $x_0$ and compute a sequence of points via $x_{n+1}=x_n - g''(x_n)^{-1}g'(x_n)$. The Newton's method converges fast because the distance between the estimate and its true value shrinks quickly such that the distance in next step of iteration is asymptotically equivalent to the squared distance in previous iteration. That is, the Newton's method has quadratic convergence order (cf. page 29 of \citealt{computational-statistics:2013,Osborne:1992}). This convergence order holds when the initial value of the iteration is in the neighborhood of the true value and the third derivative $g'''(x)$ is continuous and $g''(x)$ is non-zero.

The Fisher scoring algorithm is slightly different from the Newton's method. In the Fisher scoring algorithm, we replace the $- g''(x_n)$ by its expected value. This algorithm is asymptotically equivalent to the Newton's method and therefore enjoys the same asymptotic property such as consistency of the estimate of the parameter. As sample size gets large, the convergence order increases (\citealt{Osborne:1992}). It has some advantages over the Newton's method in that the expected value of the Hessian matrix is positive definite which guarantees the update is uphill toward the direction of maximizing the log likelihood function assuming that the model is correct and the covariates are true explanatory variables.

As the Fisher scoring algorithm is assuming $\wtvtheta$ is the true value when we estimate $\vtheta$ or vice versa, there could be complications when the parameter being fixed is not equal to the true parameter value. To see this point, note that our updating equations are all written in the context of using the Fisher scoring algorithm, which relies on the expectation of the Hessian matrix using correct distribution at the true parameter value. In particular, the algorithm uses $\wtvtheta$ as fixed value while estimating $\vtheta$. The resulting estimate of $\vtheta$ determines the distribution because the distribution is a function of $\vtheta$ and $\wtvtheta$. When the parameter being fixed %(such as $\wtvtheta$)
at a value far from its true value during the intermediate steps, the distribution is wrong even though it is in the right family. The consequence of using a wrong distribution to compute the expectation of the Hessian matrix could lead to a sequence of parameter updates that converges to a limiting value unequal to the true parameter (\citealt{Osborne:1992}). In this case, the algorithm might diverge.
Our simulation study in later section confirms this point.

To avoid this parameter update divergence problem, we introduce learning rate adjustment so that the change in parameter estimate is scaled by $lr/t^{1/4}$, where $lr$ is a small constant learning rate such as 0.1 or 0.01, and $t$ is the iteration number. That is, the general updating formula is $\vtheta_{i}^{(t+1)} = \vtheta_{i}^{(t)}+ \frac{lr}{t^{1/4}} \cdot S_{\vtheta_{i}^{(t)}}^{-1} \cdot U_{\vtheta_{i}^{(t)}}$ {\modify and the} adjustment is applied to both inner loop and outer loop iterations.
{\modify The algorithm follows the same work flow as those listed in Algorithm \ref{algorithm:alternating-regression} except that the learning rate adjustment $\frac{lr}{t^{1/4}}$ is applied to the parameter update as given in Algorithm \ref{algorithm:alternating-ZIG-regression}}. How we decide to use this learning rate adjustment comes from modifying the popularly used adaptive moment estimation (Adam) and the stochastic gradient descent. In the stochastic gradient descent, the parameter
 update has learning rate adjustment to make small moves so that it compensate the random nature of selecting only one observation to compute the gradient. Specifically, given parameters $w^{(t)}$ and a gradient function evaluated at one randomly selected observation $g^{(t)}$, the update is based on formula $w^{(t+1)} = w^{(t)} - lr \cdot a(t)  g^{(t)}$, where $a(t)$ satisfies $\sum_{t=1}^{\infty} a(t) = \infty$ and $\sum_{t=1}^{\infty} a^2(t) <\infty$. The $a(t)$ corresponds to our $t^{-1/4}  S_{\vtheta_{i}^{(t)}}^{-1}$. However, our parameter update does not use just one $\wtvtheta$. Instead, all $\wtvtheta_1, \ldots, \wtvtheta_n$ were used in computing the gradient and the information matrix.
Given parameters
$w^{(t)}$ and a loss function $L^{(t)}$ at the $t^{th}$ training iteration,  the Adam update takes the form $w^{(t+1)} = w^{(t)} - lr \cdot \hat{m}_{w}/( \sqrt{\hat{v}_{w}}+ \epsilon)$,
where $\hat{m}_{w}$ and  $\hat{v}_{w}$ are the exponential moving average of the gradients and the second moments of the gradients in the past iterations, respectively, and
$ \epsilon $ is a small scalar (e.g.
$ 10^{-8}$) used to prevent division by 0. Our use of the Fisher information $ S_{\vtheta_{i}^{(t)}}^{-1}$ should provide a better mechanism than Adam's exponential moving average of second moments to achieve the effect of
 increasing the learning rate for sparser parameters and decreasing the learning rate for ones that are less sparse. This is because Adam only uses the diagonal entries of  $S_{\vtheta_{i}^{(t)}}$ and ignores the covariance between the estimated parameters existing in the off diagonal entries. \cite{Reddi.etal} and \cite{Shi.etal} both pointed out that Adam may not converge to optimal solutions even for some simple convex problems although it is overwhelmingly popular in machine learning applications.

\begin{algorithm}
\caption{ {\modify SA-ZIG regression}}
\label{algorithm:alternating-regression}
\begin{algorithmic}
\Require $S_{\vtheta_{i}^{(t)}}^{-1}$ and $S_{\wtvtheta_{j}^{(t)}}^{-1}$ exist.
\State

\State $t \gets 0$, overall $Loss^{(t)} $ = total negative log likelihood $\gets  \infty %10^{20}
$
\State $converged = False$
\While{$t \leq maxit$}
%\While{$\sum_{i=1}^{n} || \vtheta_{i}^{(t+1)} - \vtheta_{i}^{(t)} || \ge \epsilon$ or $\sum_{i=1}^{n} || \wtvtheta_{i}^{(t+1)} - \wtvtheta_{i}^{(t)} || \ge \epsilon$}
%
\While{$converged = False$}
\State $i \gets 1$
\While{$i \leq n$}
%\State retrieve $i$th row of data %from SQLite database.
\State do n\_epoch update of $U_{\vtheta_i^{(t)}}$ and $S_{\vtheta_i^{(t)}}$ and $\vtheta_i^{(t+1)}$ using $i^{th}$ row of data. \Comment{see Algorithm \ref{algorithm:epoch-update} or \ref{algorithm:alternating-ZIG-regression}}
% \State $i \gets i+1$
% \EndWhile
% \State $j \gets 1$
% \While{$j \leq n$}
\State do n\_epoch update of $U_{\wtvtheta_i^{(t)}}$ and $S_{\wtvtheta_i^{(t)}}$ and $\wtvtheta_i^{(t+1)}$ using $i^{th}$ column of data.
\State $i \gets i+1$
\EndWhile
\State
\For{$k=1, \ldots, n$}
\State retrieve $k$th row of data %from SQLite database.
\State recompute $U_{\vtheta_k^{(t+1)}}$,  $U_{\wtvtheta_k^{(t+1)}}$ and their $L_2$ norms using $\vtheta^{(t+1)}$ and $\wtvtheta^{(t+1)}$ values.
\State recompute $loss_k$ and $\widetilde{loss}_k$ for $k^{th}$ row and column respectively using $\vtheta^{(t+1)}$ and $\wtvtheta^{(t+1)}$ values. % , where loss is negative log likelihood.
\EndFor
\State compute the overall $Loss(\vtheta^{(t+1)}, \wtvtheta^{(t+1)})$.
% \State check if the Relative change in Loss (\ref{formula:relative-change-loss}) is less than $\epsilon$.
\State check if the relative change in Loss is less than a predefined threshold $\epsilon$.
% \State check if the condition $converged = \mid Loss^{(t+1)} - Loss^{(t)} \mid / (|Loss^{(t+1)}| + 0.1) < \epsilon$
% \State is True or False.
\EndWhile
\State $t \gets t+1$
\EndWhile
% \State Stabilized $\vtheta$ and $\wtvtheta$ lead to globally or locally optimized value of the loss function of the Alternating Tweedie regression.
\end{algorithmic}
\end{algorithm}

\begin{algorithm}
\caption{Epoch update in inner loop of Algorithm \ref{algorithm:alternating-regression} {\modify without learning rate adjustment} }
\label{algorithm:epoch-update}
\begin{algorithmic}
\For{\texttt{epoch $\in$ 1,\ldots, n\_epoch}}
\State compute $U_{\vtheta_i^{(t)}}$ and $S_{\vtheta_i^{(t)}}$.
\State update $\vtheta_i^{(t+1)}$ using current value of \{ $\wtvtheta_{j}^{(t)}, j\!=\!1, \ldots, n$ \} and $\vtheta_i^{(t)}$ based on formulae in (\ref{formula:ZIG-loglink-update-epoch})
%\State based on formulae in (\ref{formula:ZIG-loglink-update-epoch})
% \State based on Fisher scoring update.
\EndFor

\For{\texttt{epoch $\in$ 1,\ldots, n\_epoch}}
\State compute $U_{\wtvtheta_i^{(t)}}$ and $S_{\wtvtheta_i^{(t)}}$.
\State update $\wtvtheta_i^{(t+1)}$ using current value of \{ $\vtheta_{j}^{(t)}, j\!=\!1, \ldots, n$ \}, and $\wtvtheta_i^{(t)}$
\State based on formulae in (\ref{ZIG-loglink-update-tilde-epoch})
% \State based on Fisher scoring update.
\EndFor
\end{algorithmic}
\end{algorithm}

\begin{algorithm}
\caption{Updating equations in {\modify SA-ZIG} regression with learning rate adjustment}
\label{algorithm:alternating-ZIG-regression}
\begin{algorithmic}
\For{$i=1, \ldots, n$}
\For{$k=1, \ldots, E$}
\State $\vtheta_{i}^{(t, k+1)} = \vtheta_{i}^{(t, k)}+ \frac{lr}{t^{1/4}} \cdot S_{\vtheta_{i}^{(t, k)}}^{-1} \cdot U_{\vtheta_{i}^{(t, k)}}$
\State $\wtvtheta_{i}^{(t, k+1)} = \wtvtheta_{i}^{(t, k)}+ \frac{lr}{t^{1/4}} \cdot S_{\wtvtheta_{i}^{(t, k)}}^{-1} \cdot U_{\wtvtheta_{i}^{(t, k)}}$
\EndFor
\State $\vtheta_i^{(t)} \gets \vtheta_i^{(t, E)}$
\State $\wtvtheta_i^{(t)} \gets \wtvtheta_i^{(t, E)}$
\State $\vtheta_{i}^{(t+1)} = \vtheta_{i}^{(t)}+ \frac{lr}{t^{1/4}} \cdot S_{\vtheta_{i}^{(t)}}^{-1} \cdot U_{\vtheta_{i}^{(t)}}$
\State $\wtvtheta_{i}^{(t+1)} = \wtvtheta_{i}^{(t)}+ \frac{lr}{t^{1/4}} \cdot S_{\wtvtheta_{i}^{(t)}}^{-1} \cdot U_{\wtvtheta_{i}^{(t)}}$
\EndFor
\end{algorithmic}
\end{algorithm}
Using the learning rate adjustment makes smaller steps in each update before changing directions. This learning rate adjustment turns out to be crucial. The simulation study in next section {\modify examines} the effect of the learning rate adjustment.

\section{{\revise Numerical studies} }

\subsection{A simulation study with ZIG using log link}
\label{simulation}

In this section, we present a simulation study to assess the performance of the alternating ZIG regression. We found through some experiments that data generation have to be very careful because the mean of the Gamma distribution was given by $\mu_{i j} = \exp(\vw_i^\top \wtvw_j + e_i + \wte_j)$. When we randomly generate the $\vw_i$'s and $\wtvw_j$'s independently from Uniform distribution with each of them having dimension 50, their dot product could easily become so large that exponentiated value $\mu_{i j}$ and the variance of the distribution ($\varpropto \mu_{i j}^2$) become infinity or undefined. Keeping this point in mind, we generate the data as follows:
\bit
\item The $w_{i k}$'s were generated independently from Uniform(-0.25, 0.25) with seed 99, $i=1, \ldots, 300$ and $k=1, \ldots, 50$. Set $\vw_i=(w_{i1},\ldots, w_{i50})'$.
\item The $\wtw_{i k}=w_{i k}$, $i=1, \ldots, 300$ and $k=1, \ldots, 50$. That is, $\wtvw_i = \vw_i$.
\item The $b_i$'s and $\wtb_i$'s were generated independently from Uniform(0, 0.05) with seed 97 and 96, respectively, $i=1, \ldots, 300$.
\item The $e_i$'s and $\wte_i$'s were generated independently from Uniform(0.1, 0.35) with seed 1 and 2, respectively, $i=1, \ldots, 300$.
\item The success probability $p_{i j}$ of positive observations was calculated as $p_{i j} = (1+\exp(-\eta_{i j}))^{-1}$, where $\eta_{i j}$ is based on the logit formula (\ref{formula:ZIG-logitlink-eta-ij}).
\item Generate $B_{i j}$ independently from Bernoulli distribution with success probability $p_{i j}$, $i=1, \ldots, 300$ and $j=i, \ldots, 300$.
\item If $B_{i j} = 0$, set $Y_{i j} = 0$. Otherwise, generate $Y_{i j}$ from Gamma distribution with mean $\mu_{i j} = \exp(\tau_{i j})$ and shape parameter $\nu_{i j} = 4$, where $\tau_{i j}$ is computed based on formula (\ref{formula:ZIG-loglink-tau-ij}) for $i=1, \ldots, 300$ and $j=1, \ldots, 300$.
\eit
The matrix containing $Y_{i j}$, $i,j=1, \ldots, 300$, is the data we used for the alternating ZIG regression. With the generated data matrix $\vY$, we applied our algorithm to the data with two different sets of initial values for the unknown parameters. One setting is to check whether the algorithm would perform well when the parameter estimation is less difficult when part of the parameters were initialized with true values. The other setting demands more accurate estimation in both $\vw_i$ and $\wtvw_j$ in the right direction.

In Setting 1, we set all the parameters' initial values equal to their true values except for $\wtw_{i k}, i=1, \ldots, 300, k=1, \ldots, 50$. The initial value of $\wtw_{i k}$'s were randomly generated from Uniform(-0.25, 0.25) with seed number 98. Even though initial values of $\vw, b, \wtb, e$ and $\wte$ were set to be equal to the true values, the algorithm was not aware of this and these parameters were still estimated along with $\wtvw$.

In Setting 2, we generated the initial values for all the parameters randomly as follows:
\bit
\item $\vw$'s initial values were generated independently from Uniform(-0.25, 0.25) with dimension 300 by 50 and seed 102.
\item $\wtvw$'s initial values were generated independently from Uniform(-0.25, 0.25) with dimension 300 by 50 and seed 103.
\item $b_i$'s and $\wtb_j$'s initial values were generated independently from Uniform(0, 0.05) with seed 104 and 105, respectively, $i,j=1, \ldots, 300$.
\item $e_i$'s and $\wte_j$'s were independently generated from Uniform(0.1, 0.35) with seed 106 and 107, respectively, $i,j=1, \ldots, 300$.
\eit

We consider both update with learning rate adjustment {\modify (Algorithm \ref{algorithm:alternating-ZIG-regression})} and without learning rate adjustment {\modify (Algorithm \ref{algorithm:alternating-regression})}.

% data generation
% w = (np.random.RandomState(99).rand(V, d) - 0.5) * 0.5 # V * d
% w_til = w
% # w_til = (np.random.RandomState(98).rand(V, d) - 0.5) # V * d
% b = np.random.RandomState(97).rand(V,1)*0.05
% b_til = np.random.RandomState(96).rand(V,1)*0.05
% e = np.random.RandomState(1).rand(V,1)*0.25 + 0.1
% e_til = np.random.RandomState(2).rand(V,1)*0.25 + 0.1

% initialization
% w = (np.random.RandomState(102).rand(V, d) - 0.5) * 0.5 # V * d
% # w_til = w
% w_til = (np.random.RandomState(103).rand(V, d) - 0.5) *0.5 # V * d
% b = np.random.RandomState(104).rand(V,1)*0.05
% b_til = np.random.RandomState(105).rand(V,1)*0.05
% e = np.random.RandomState(106).rand(V,1)*0.25 + 0.1
% e_til = np.random.RandomState(107).rand(V,1)*0.25 + 0.1

\begin{figure}[H]
    \centering
    \includegraphics[width=0.65\textwidth]{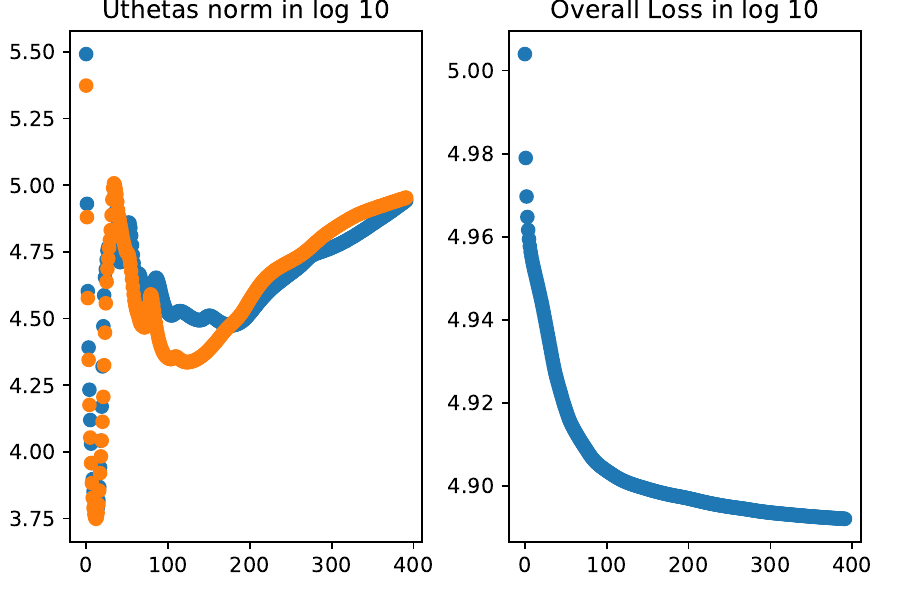}
    \includegraphics[width=0.65\textwidth]{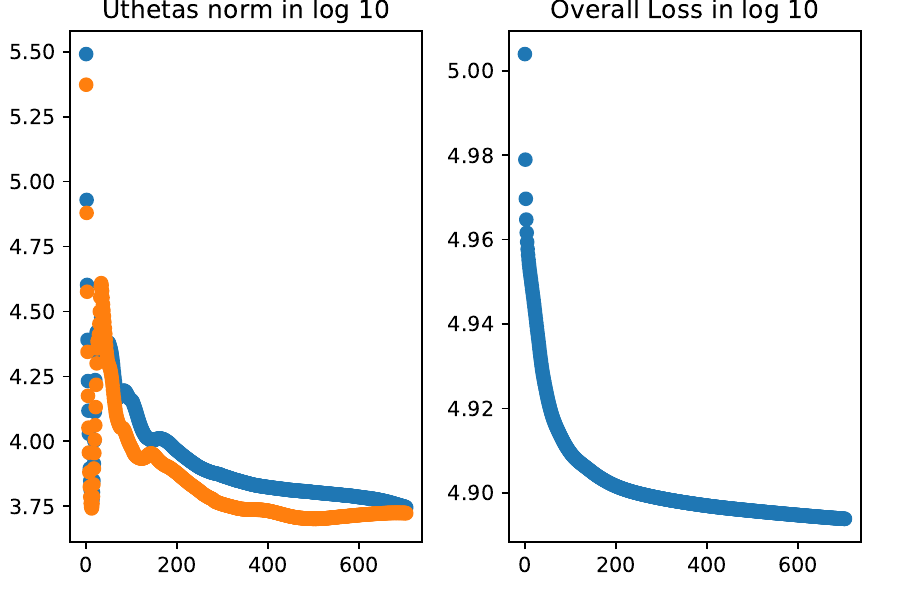}
    \caption{\small $L_2$ norm of the score functions $\vU_{\vtheta}$ and $\vU_{\wtvtheta}$ (Uthetas norm), and overall loss of the ZIG model in log$_{10}$ scale. Top two panels: without learning rate adjustment;  bottom two panels: with learning rate adjustment. All parameters were initialized with true parameter values except for $\wtvw$. %The left panels give the norm of the score vectors and the right panels give overall loss in $\log_{10}$ scale over iteration.
     In the top panels, the norms of the score vectors reduce drastically in early iterations but increase over later iterations even though the overall loss is consistently reduced. In the bottom panels, the norms of the score vectors %reduce quickly to a certain level but bounces back.
     first show similar pattern as the top panel but consistently reduce in later iterations. The overall loss shows the desired reducing trend throughout all iterations. %{\modified change title Utheta to expression}
     }
    \label{fig:ZIG-loglink-Utheta-loss-true-parameters}
\end{figure}

The results of Setting 1 are presented in Table \ref{table:ZIG-loglink-true-parameters} and Figure \ref{fig:ZIG-loglink-Utheta-loss-true-parameters}. We can see in Figure \ref{fig:ZIG-loglink-Utheta-loss-true-parameters} that even though the norm of the score vectors went up after initial reduction, the loss function (i.e., the negative log-likelihood) consistently reduces as the iteration number increases. In the last 5 iterations, the algorithm with no learning rate adjustment {\modify (Algorithm \ref{algorithm:alternating-regression})} reached lower value in overall loss compared to that with learning rate adjustment {\modify (Algorithm \ref{algorithm:alternating-ZIG-regression})}. However, the norm of the score vectors from the algorithm with learning rate adjustment are much smaller than those from the algorithm with no learning rate adjustment (see Table \ref{table:ZIG-loglink-true-parameters}).

\begin{table}[ht]
%\centering
%\begin{center}
\begin{tabular}{rrrr|rrr}
  \hline
  & \multicolumn{3}{c|}{No learning rate adjustment} & \multicolumn{3}{c}{With learning rate adjustment} \\ \cline{2-7}
Iteration & $||U_{\vtheta}||_{L_2}$ & $||U_{\wtvtheta}||_{L_2}$ & Overall Loss & $||U_{\vtheta}||_{L_2}$ & $||U_{\wtvtheta}||_{L_2}$ & Overall Loss \\
  \hline
-5 & 85805.79 & 88896.77 & 77998.77 & 5596.06 & 5298.31 & 78313.71 \\
  -4 & 86123.05 & 89012.56 & 77996.44 & 5596.00 & 5303.81 & 78312.30 \\
  -3 & 86614.46 & 89236.36 & 77994.11 & 5581.83 & 5294.08 & 78310.92 \\
  -2 & 87309.84 & 89758.44 & 77991.79 & 5579.14 & 5293.48 & 78309.59 \\
  -1 & 87574.27 & 89699.66 & 77989.41 & 5556.29 & 5286.12 & 78308.15 \\
   \hline
\end{tabular}
%\end{center}
\caption{Result of the last 5 iterations for the $L_2$ norm of the score vectors and overall loss in ZIG model with log link. The parameters were initialized with true parameters except $\wtvw$ which was randomly initialized.  The negative iteration number means counting from the end. The algorithm with no learning rate adjustment did not reduce the score vectors' norms as fast as the algorithm with learning rate adjustment even though it achieved slightly lower loss.
}
\label{table:ZIG-loglink-true-parameters}
\end{table}

\begin{figure}[H]
\setlength{\tabcolsep}{-2pt}
\renewcommand{\arraystretch}{0.3}
\begin{tabular}{cccc}
    \includegraphics[width=0.25\linewidth]{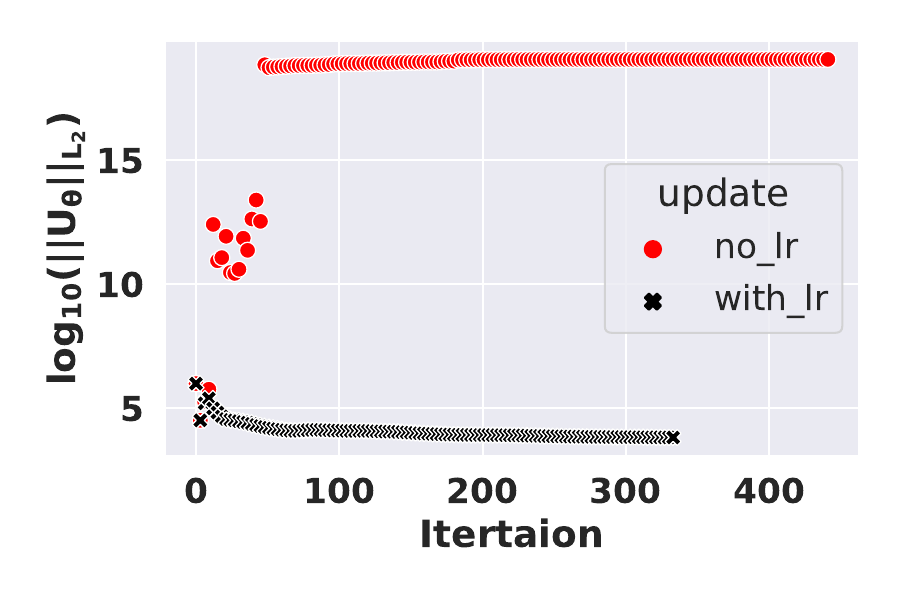} &
    \includegraphics[width=0.25\linewidth]{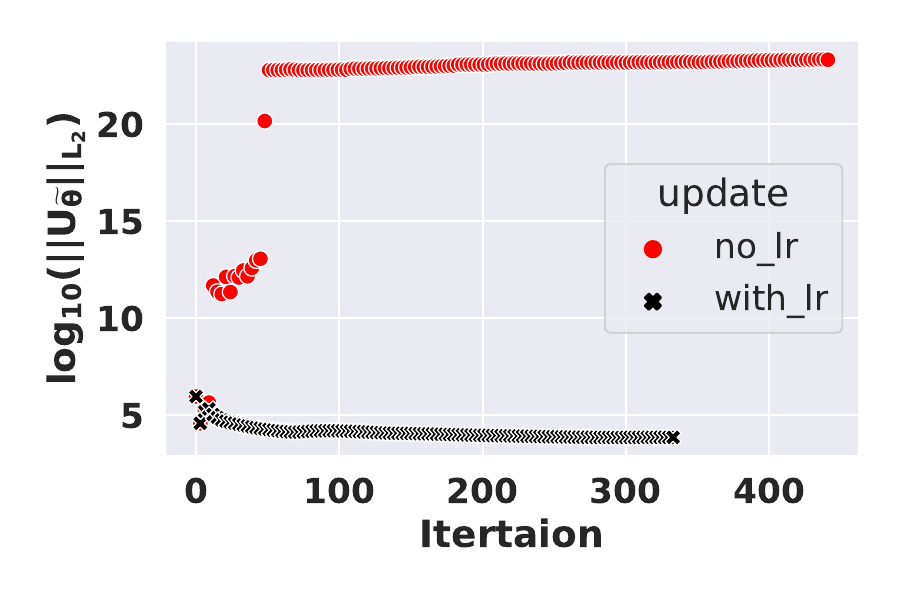} &
    \includegraphics[width=0.25\linewidth]{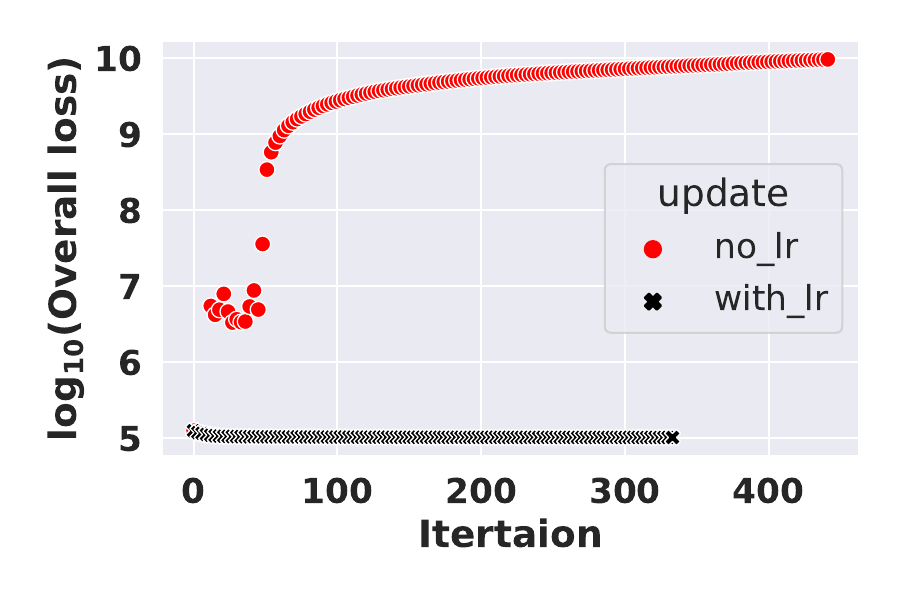} &
    \includegraphics[width=0.25\linewidth]{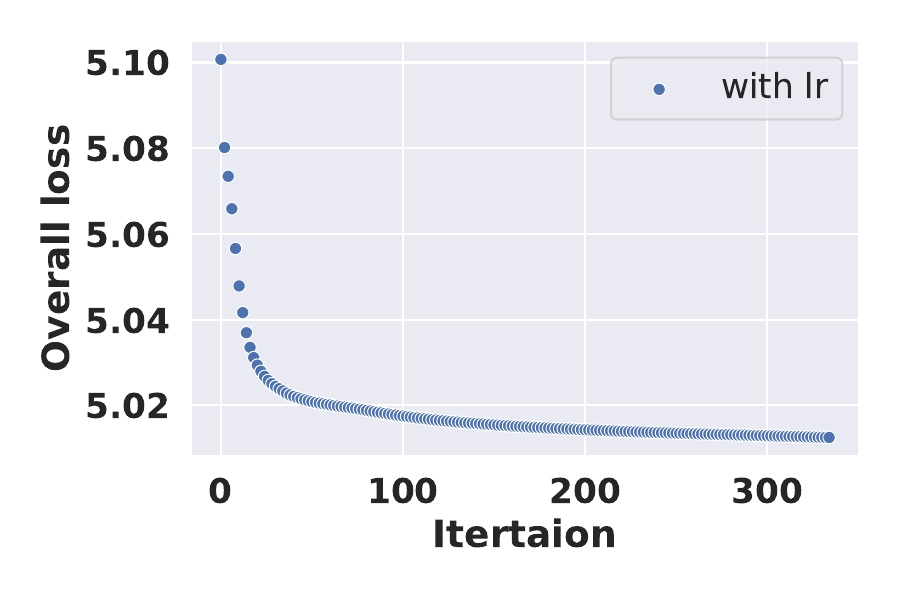} \\
    \includegraphics[width=0.25\linewidth]{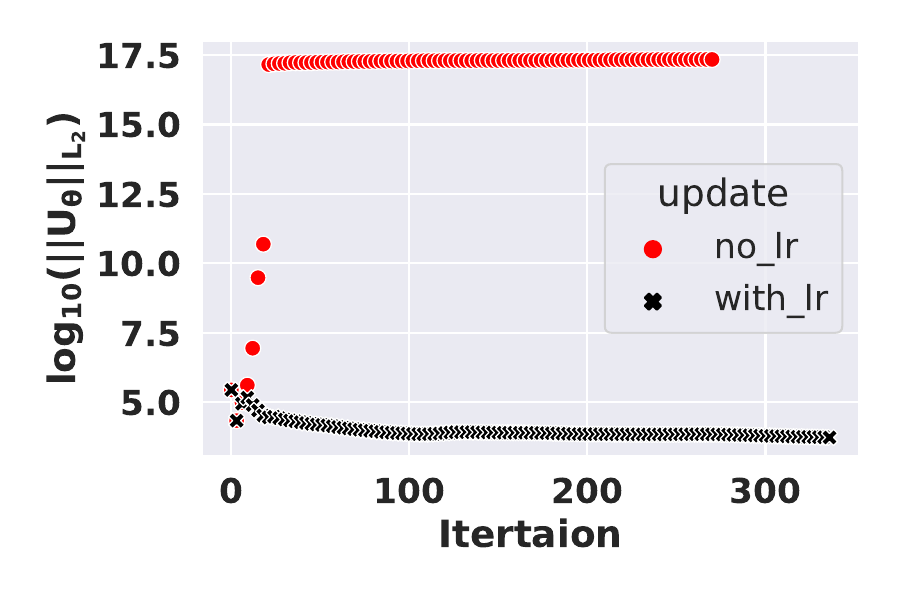} &
    \includegraphics[width=0.25\linewidth]{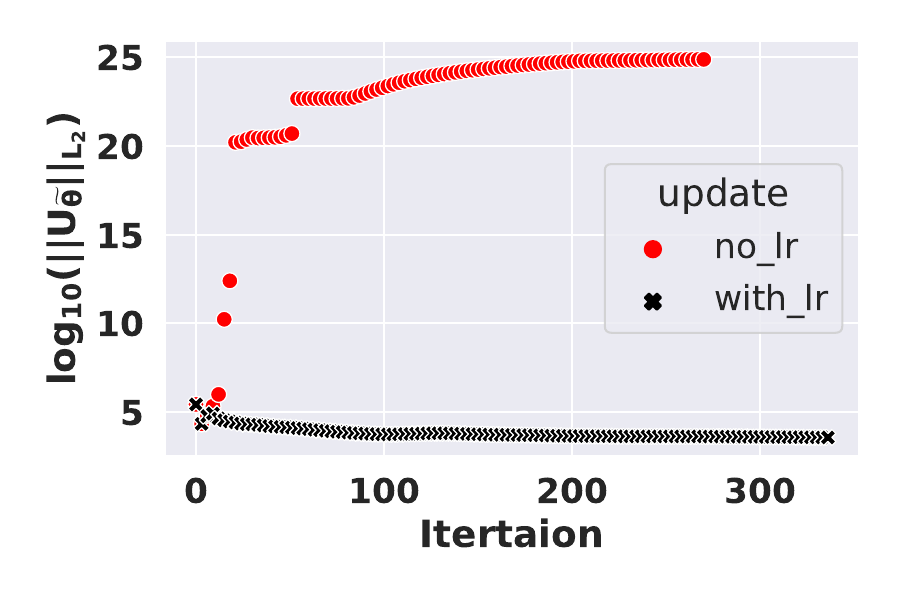} &
    \includegraphics[width=0.25\linewidth]{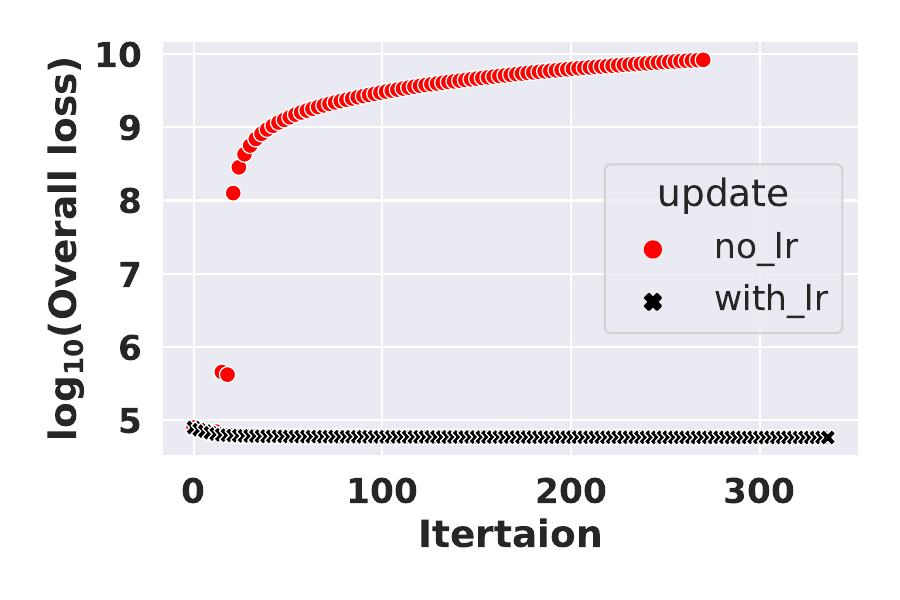} &
    \includegraphics[width=0.25\linewidth]{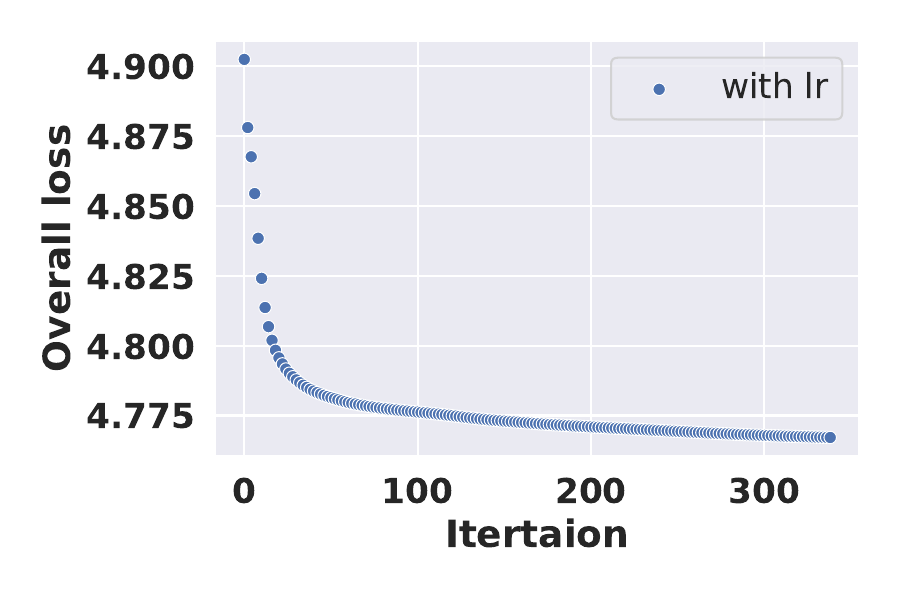} \\
    \includegraphics[width=0.25\linewidth]{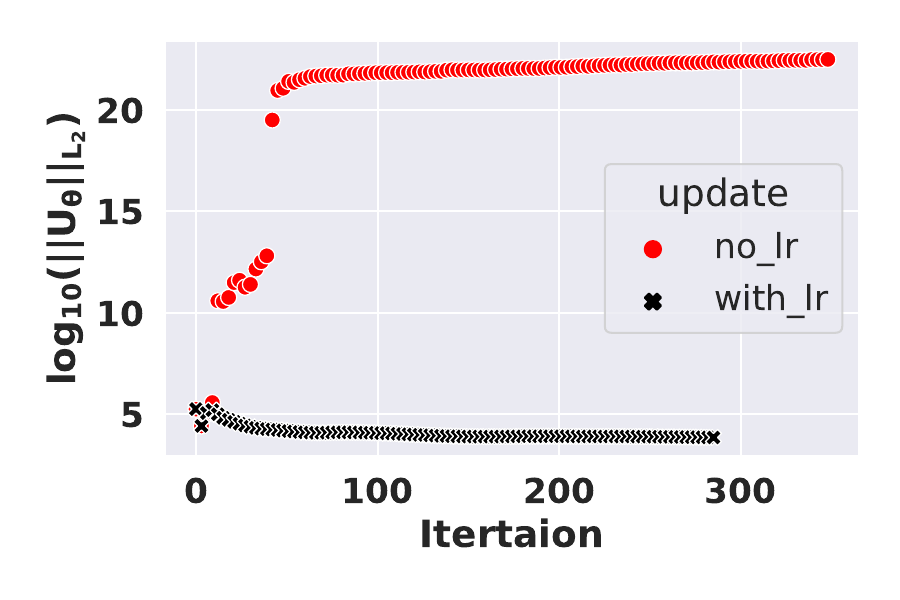} &
    \includegraphics[width=0.25\linewidth]{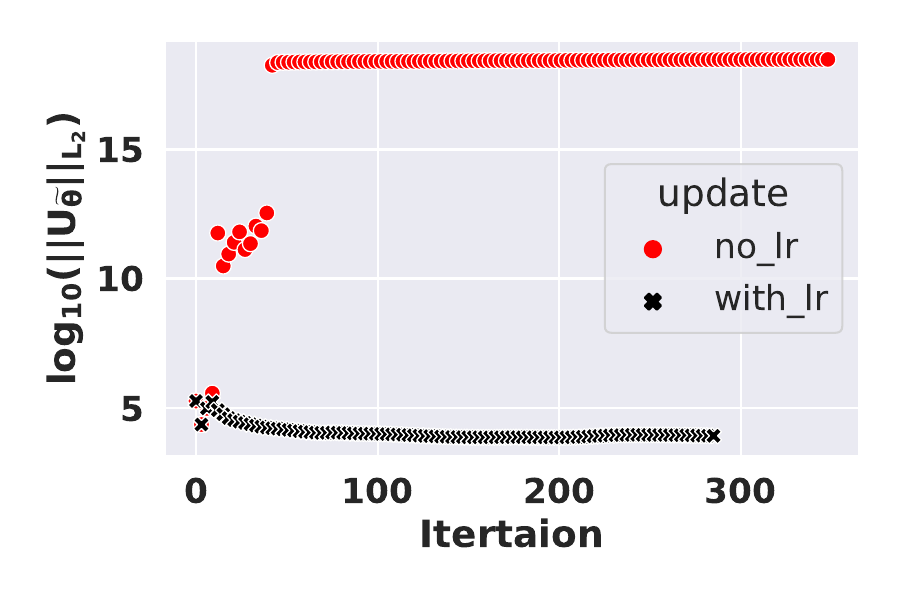} &
    \includegraphics[width=0.25\linewidth]{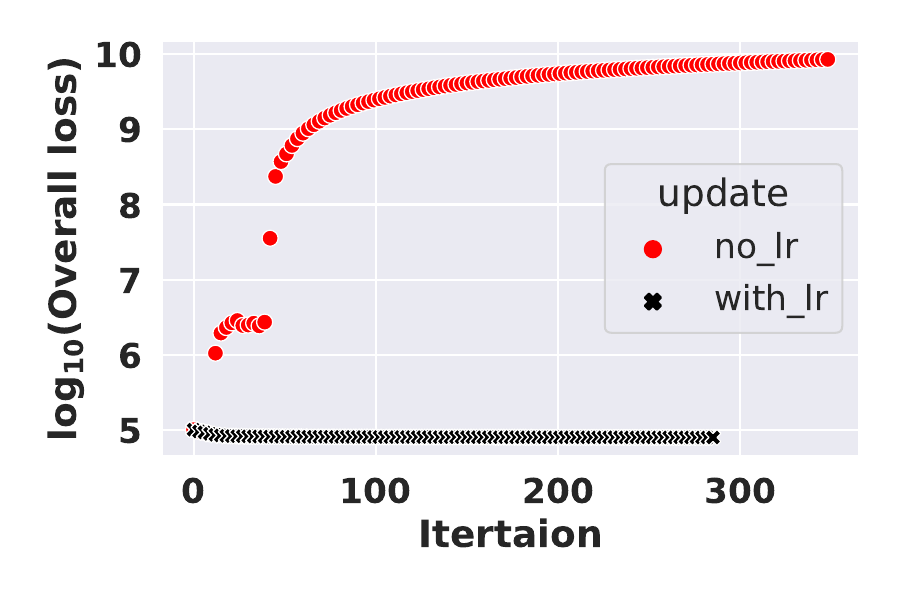} &
    \includegraphics[width=0.25\linewidth]{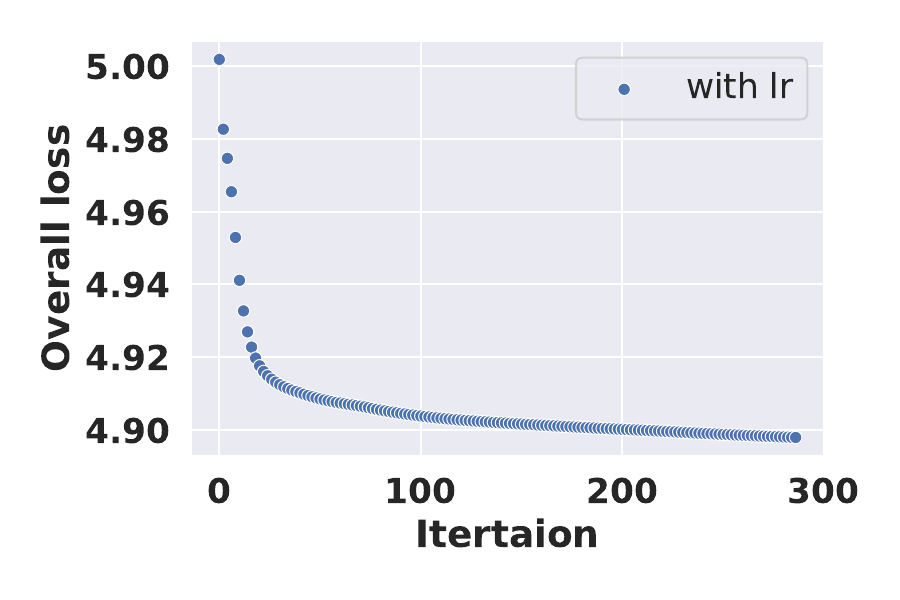} \\
    \includegraphics[width=0.25\linewidth]{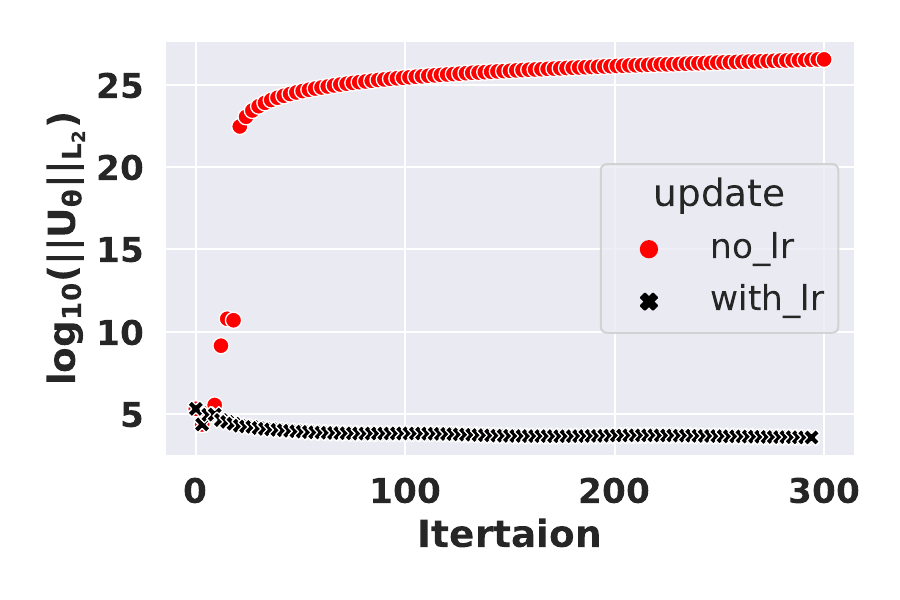} &
    \includegraphics[width=0.25\linewidth]{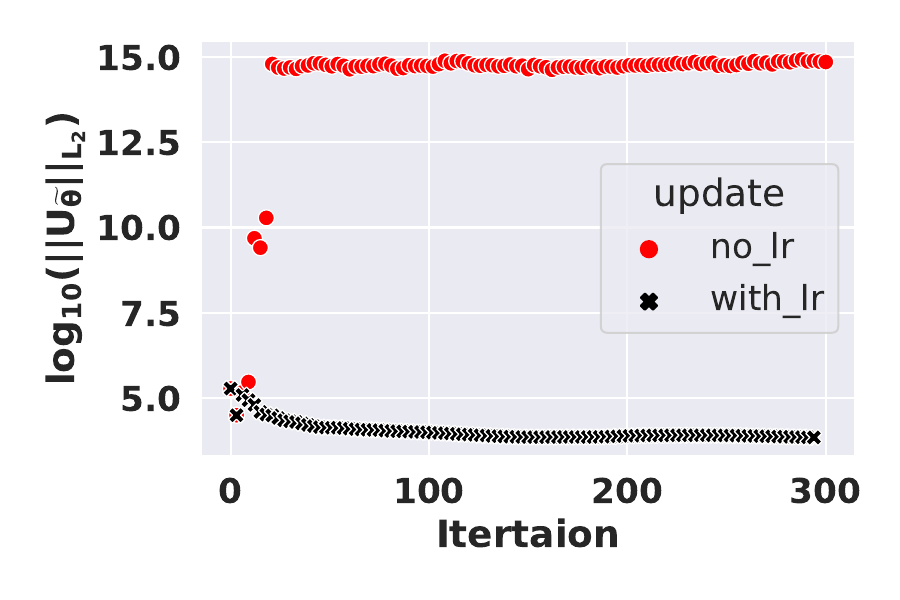} &
    \includegraphics[width=0.25\linewidth]{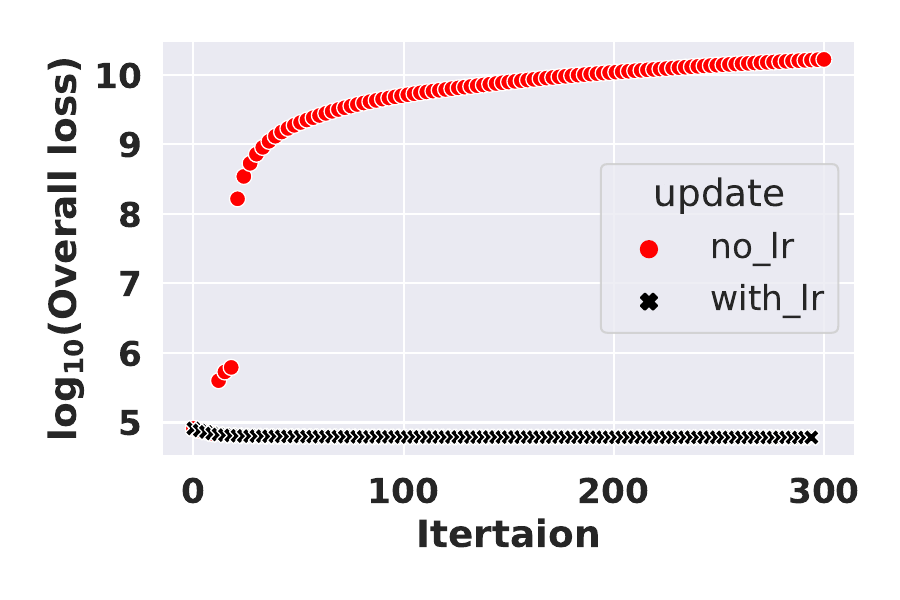} &
    \includegraphics[width=0.25\linewidth]{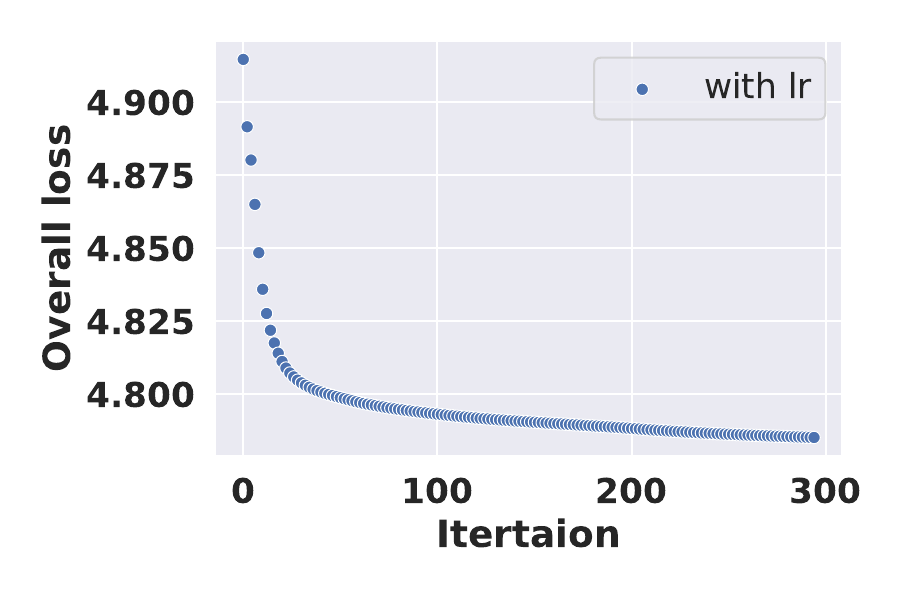} \\
    \includegraphics[width=0.25\linewidth]{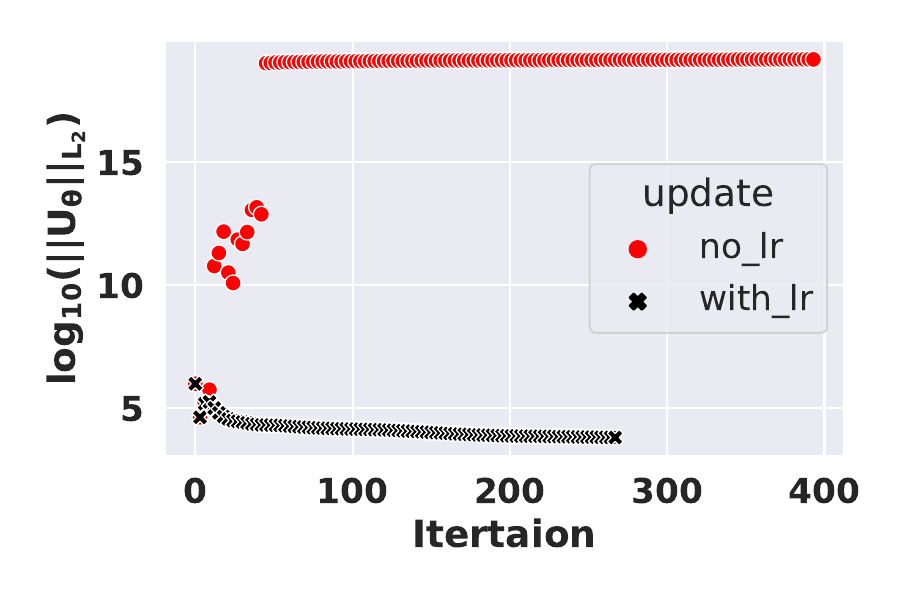} &
    \includegraphics[width=0.25\linewidth]{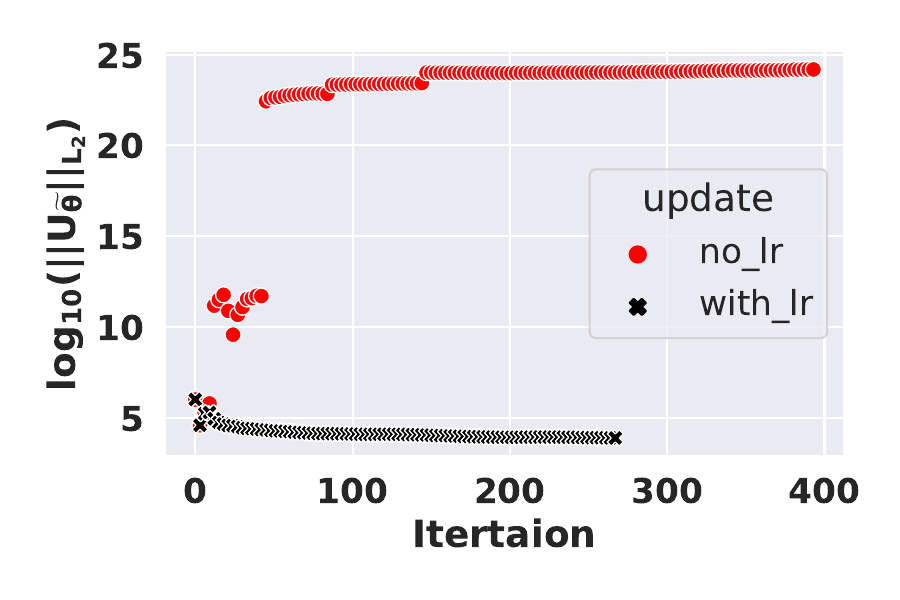} &
    \includegraphics[width=0.25\linewidth]{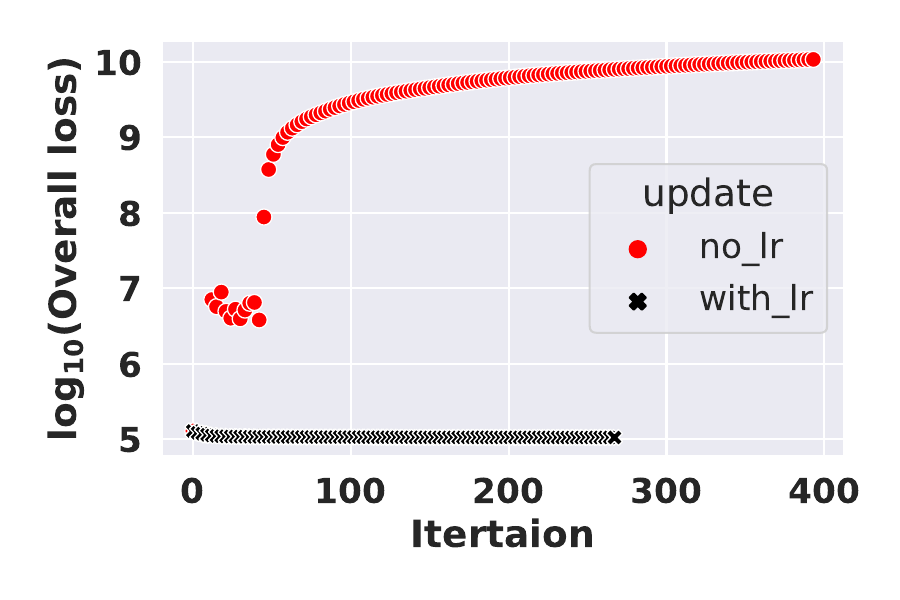} &
    \includegraphics[width=0.25\linewidth]{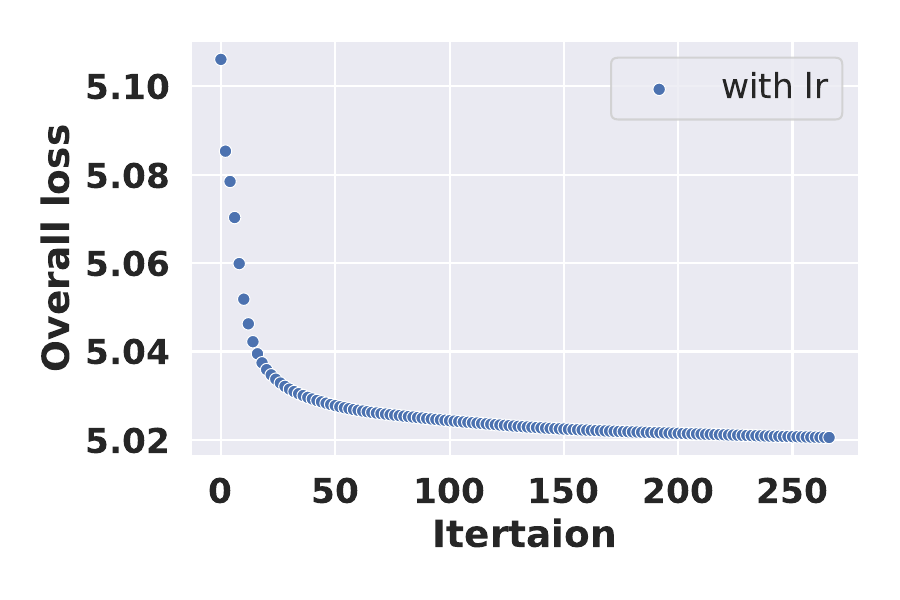} \\
    \includegraphics[width=0.25\linewidth]{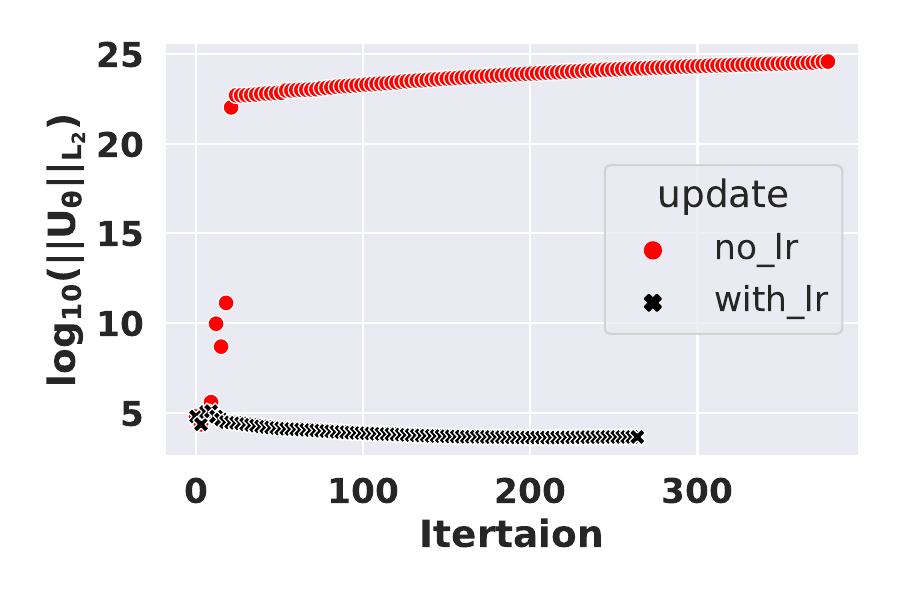} &
    \includegraphics[width=0.25\linewidth]{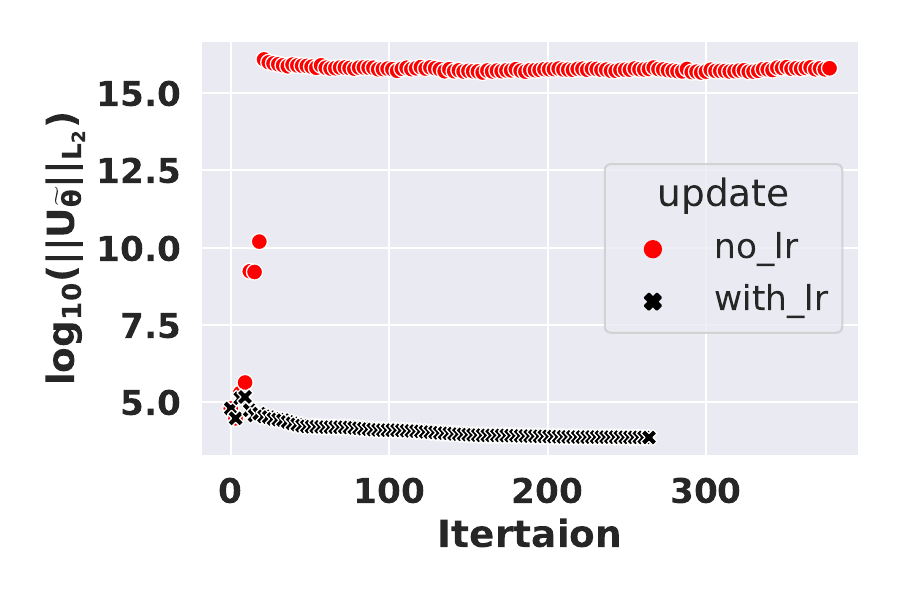} &
    \includegraphics[width=0.25\linewidth]{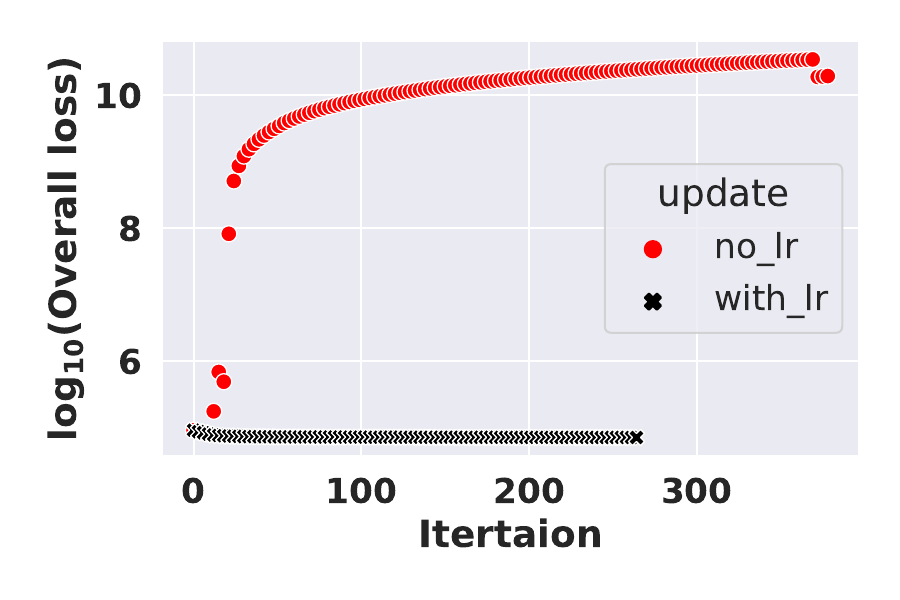} &
    \includegraphics[width=0.25\linewidth]{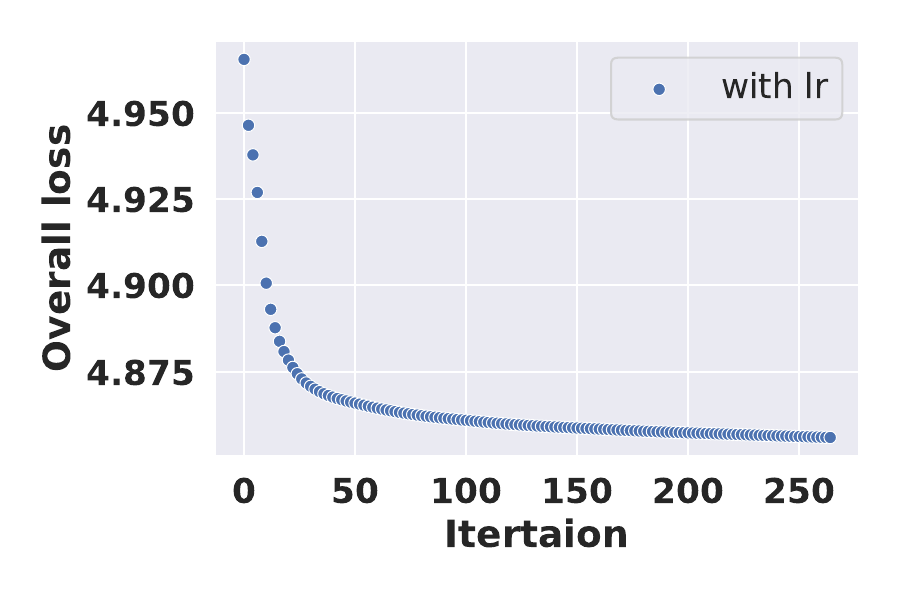} \\
    \includegraphics[width=0.25\linewidth]{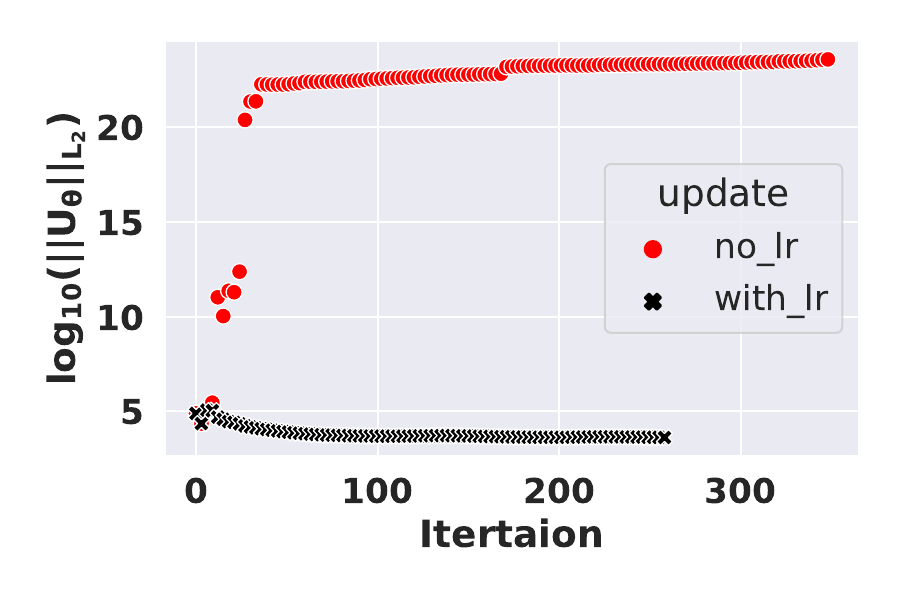} &
    \includegraphics[width=0.25\linewidth]{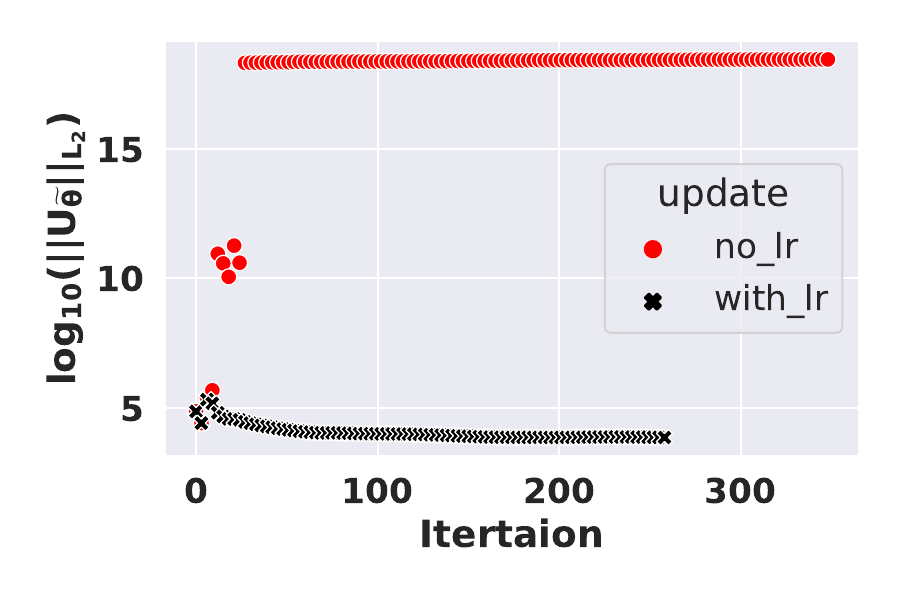} &
    \includegraphics[width=0.25\linewidth]{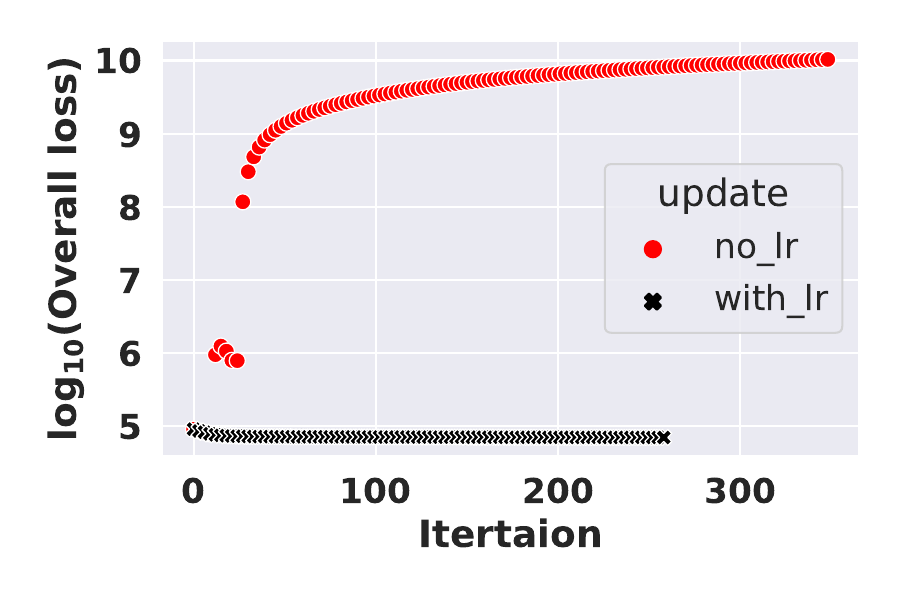} &
    \includegraphics[width=0.25\linewidth]{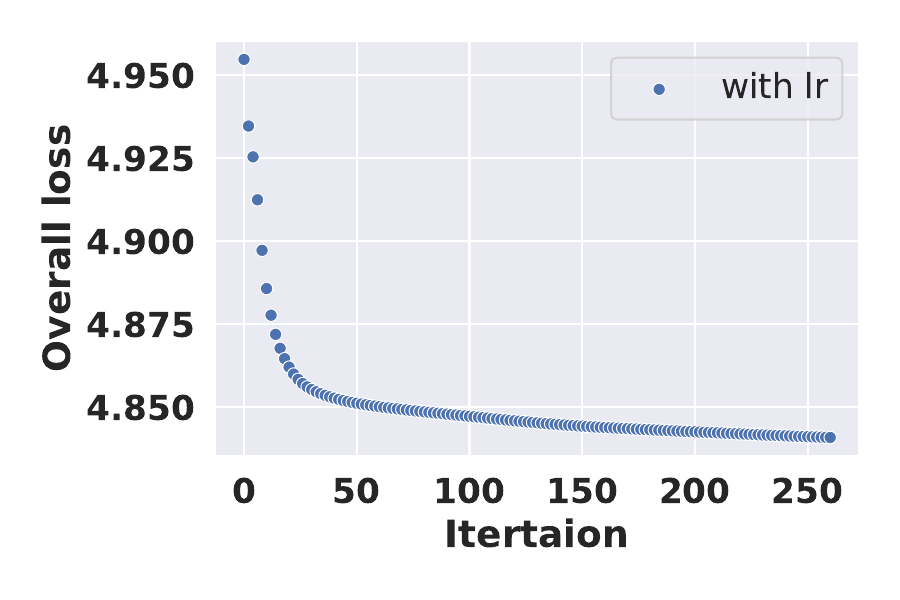} \\
    \includegraphics[width=0.25\linewidth]{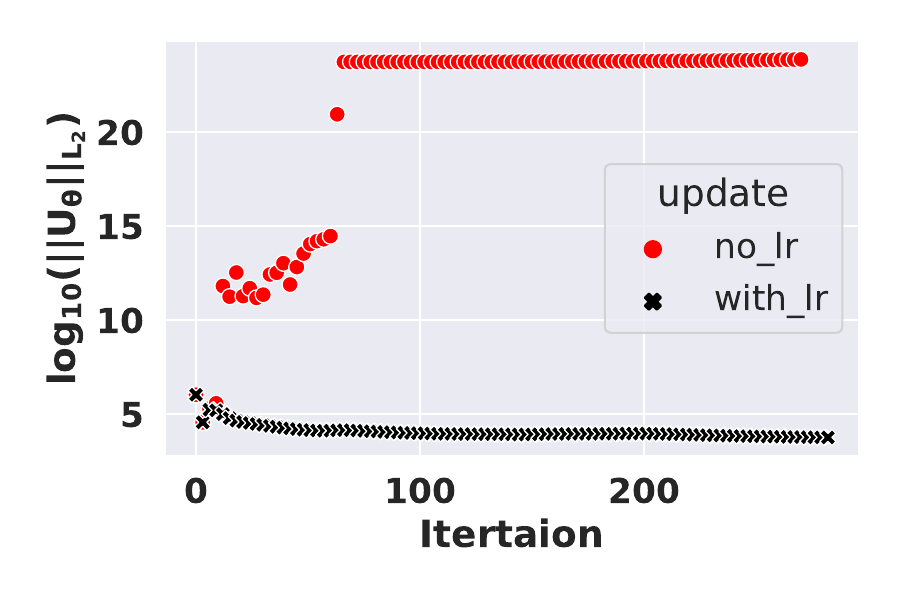} &
    \includegraphics[width=0.25\linewidth]{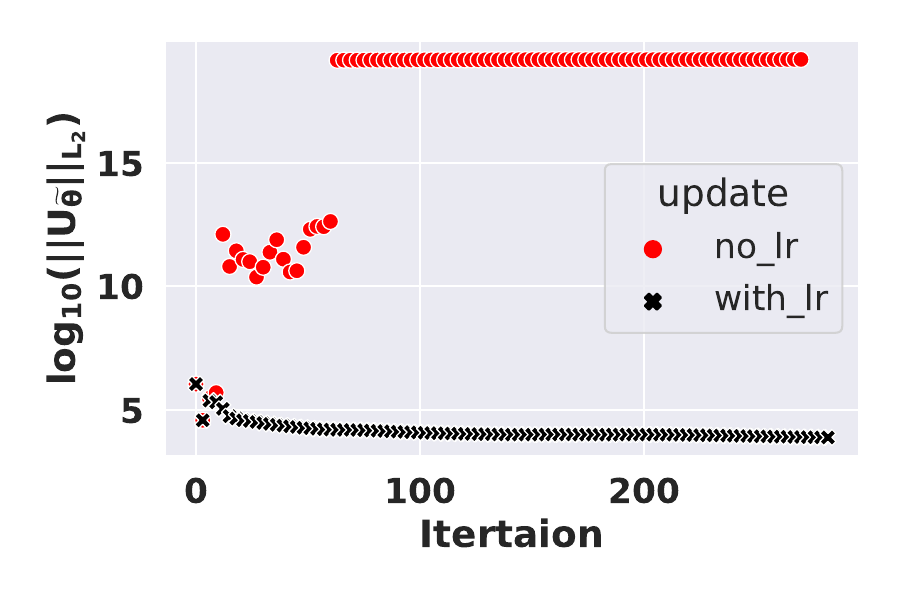} &
    \includegraphics[width=0.25\linewidth]{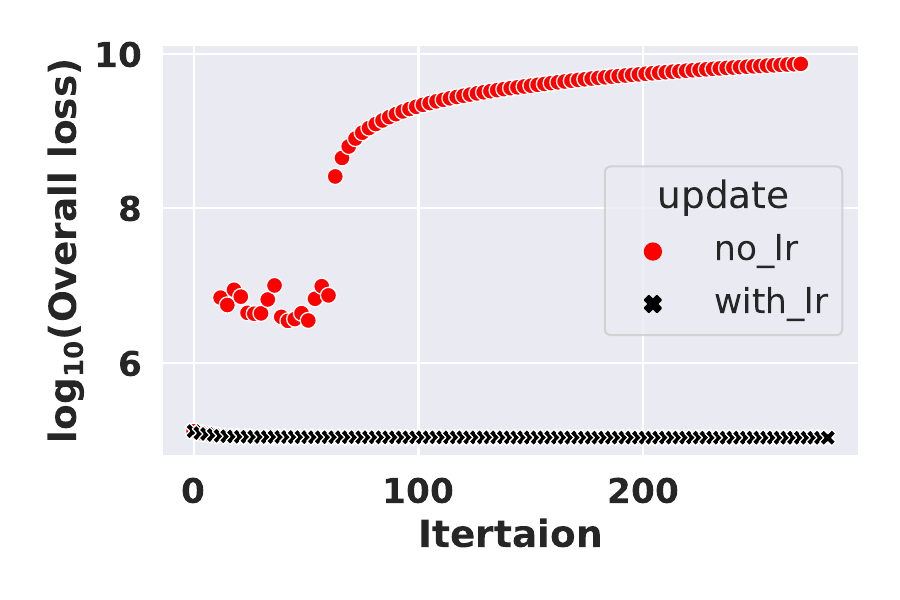} &
    \includegraphics[width=0.25\linewidth]{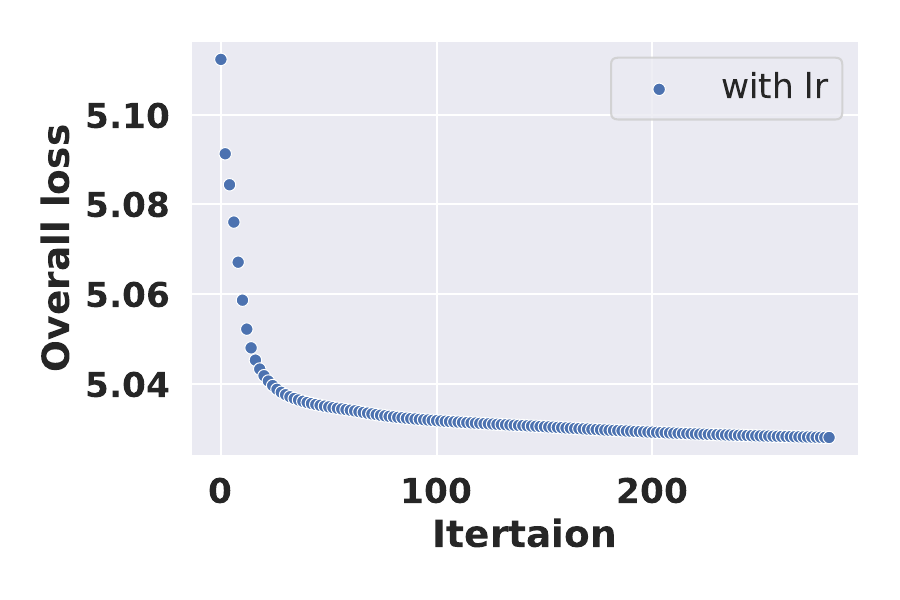} \\
\end{tabular}
%\vspace{-0.2in}
\caption{\small Comparing performance of the alternating ZIG regression using log link with or without learning rate adjustment over 8 simulated datasets. Each row is for one dataset.
First column: $\log_{10}(L_2$ norm of $U_{\vtheta})$; Second column: $\log_{10}(L_2$ norm of $U_{\wtvtheta})$; Third column: $\log_{10}($overall loss$)$; Fourth column: Overall loss of the algorithm with learning rate adjustment.
}
\label{fig:MLE_ZIG_loglink_overall_simul_plots}
\end{figure}

The results for Setting 2 are given in Figure \ref{fig:MLE_ZIG_loglink_overall_simul_plots}.
%The first and the second columns in Figure \ref{fig:MLE_ZIG_loglink_overall_simul_plots} show the $L_2$ norm of the score vectors. The third  column exhibits the overall loss of two algorithms (with or without learning rate adjustment).
The last column shows the algorithm with learning rate adjustment alone because the scale of the third column is too high to see the reducing trend. In this case, with the initial values of all parameters generated randomly from Uniform distribution, the uncertainty in the parameters caused the algorithm without learning rate to breakdown. We see that the algorithm with no learning rate adjustment {\modify using updates in Algorithm \ref{algorithm:epoch-update}} struggles to reduce the norms and headed toward wrong directions throughout the optimization process. On the other hand, the algorithm with learning rate adjustment had a little fluctuation at the beginning but continued toward the right direction by reducing both the norm of the score vectors and the loss function as the iterations proceed. Note that the dataset was identical for these two applications and both algorithms started with the same initial values. The only difference between the two algorithms is that learning rate adjustment is used in one algorithm but not in the other algorithm. Hence, the failure of the algorithm with no learning rate adjustment
is due to update being too big which leads to the algorithm going toward wrong directions.
{\modify The numerical study confirms that the alternating updates are capable to find the maximum likelihood estimate and learning rate adjustment in Algorithm \ref{algorithm:alternating-ZIG-regression} is an important component of the algorithm.}
%Application of the proposed method will be presented in a separate paper.

{\revise

\subsection{An application of SA-ZIG on word embedding }
In this section, we demonstrate the application of SA-ZIG using log link on a small dataset. This dataset comprises news articles sourced from the Reuters news archive. Using a Python script to simulate user clicks, we downloaded approximately 2,000 articles from the Business category, spanning from 2019 to the summer of 2021. These articles are stored in a SQLite database. From this corpus, we created a small vocabulary consisting of the $V=300$ most frequently used words. We train a $d=20$-dimensional dense vector representation for each of the 300 words using the downloaded business news.

To apply the SA-ZIG model, we first obtain weighted word-word co-occurrence count computed as follows:

\bqa
Y_{ij} =\left\{\begin{array}{cc}
    \sum_{\mbox{s $\in$ all sentences}} 1/d_{ij}(s),  & \mbox{if $d_{ij}(s) \le$ k} \\
    0, & \text{otherwise}
\end{array} \right.
\eqa
where $d_{ij}(s)$ = separation between word $i$ and word $j$ in sentence $s$, $k$ is pre-determined window size 10.
%The sparse entries of the co-occurrence counts are stored in a sqlite database for model training.

 Figure \ref{fig_hist_app1} shows the histogram of the counts from the first 12 rows. The counts for each row are highly skewed and has a lot of zeros.It can be seen that zero inflated Gamma may be suitable for the data.

 \begin{figure}
 \includegraphics[width=\linewidth]{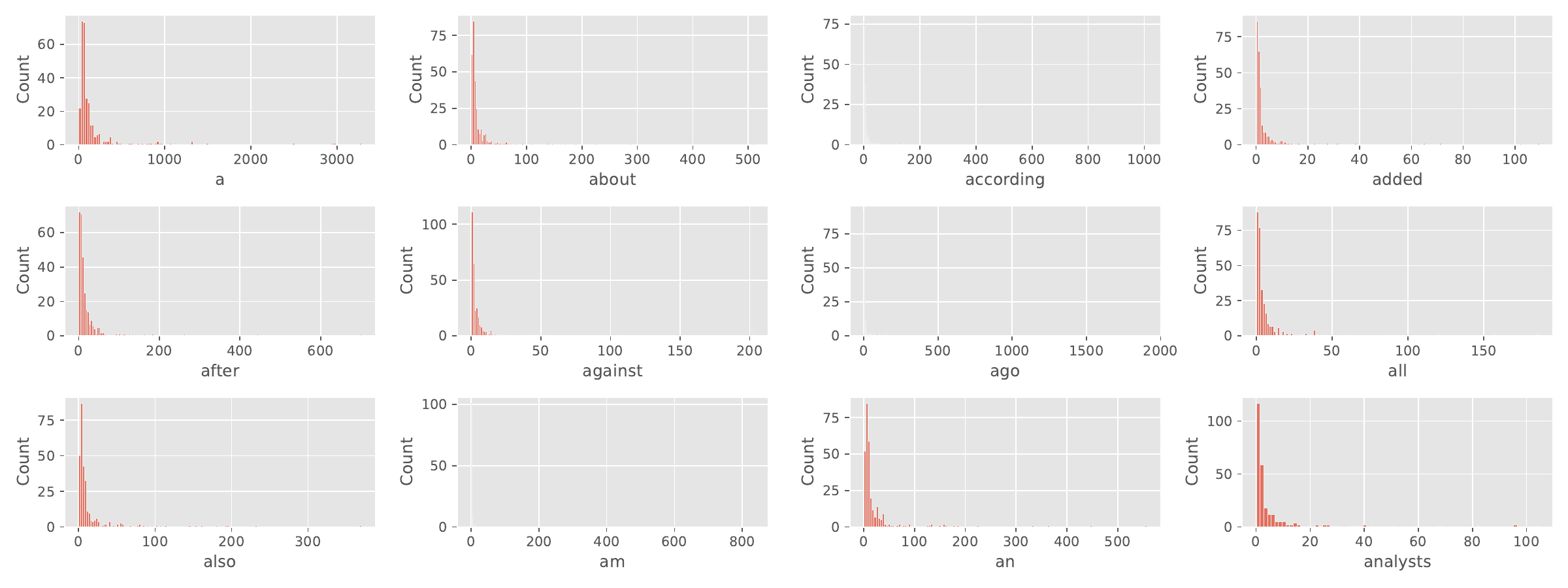}
 \caption{Histogram of weighted count from each of the first 12 rows of the data matrix}
 \label{fig_hist_app1}
\end{figure}

The co-occurrence matrix was sparse format was fed into the SA-ZIG model with learning
rate adjustment. The parameters are initialized based on the following description:

\bit
\item The entries in $V\times d$ matrix $\vw$ were independently generated from uniform distribution between $-0.5/(Vd)$ and  $0.5/(Vd)$ with seed 99.
    \item The entries in $V\times d$ matrix $\wtvw$ were generated from uniform distribution between $-0.5/(Vd)$ and  $0.5/(Vd)$ with seed 98.
        \item The $V$ bias terms in $\vb$ were independently generated from uniform (-0.1, 0.1) with seed 97.
            \item The $V$ bias terms in $\wtvb$ were independently generated from uniform (-0.1, 0.1) with seed 96.
\item The $V$ bias terms in $e_1, \ldots, e_V$ were independently generated from uniform (0.1, 0.6) with seed 1.
    \item The bias terms in $\wte_1, \ldots, \wte_V$ were independently generated from uniform (0.1, 0.6) with seed 2.
    \item The learning rate at $t^{th}$ iteration is set to $lr/(t^{1/4})$ as given in Algorithm \ref{algorithm:alternating-ZIG-regression} with $lr=0.5$.
    \eit

We let the model parameter update run 60 iterations in the outler loop and 20 epochs in the inner loop. These numbers are used because the GloVe model training in \cite{pennington2014glove} run 50 iterations  for vector dimensions $\le$ 300.
We want to see how the estimation proceeds with the iteration number. At 60$^{th}$ iteration, the algorithm has not converged yet. This can be seen from the loss curve in Figure \ref{loss_and_cor_of_vectors}, which still sharply moves downward.
The cosine similarity between word pairs are shown in the heatmap on the right panel of Figure \ref{loss_and_cor_of_vectors}.

\begin{figure}
\begin{tabular}{cc}
\includegraphics[width=0.5\linewidth]{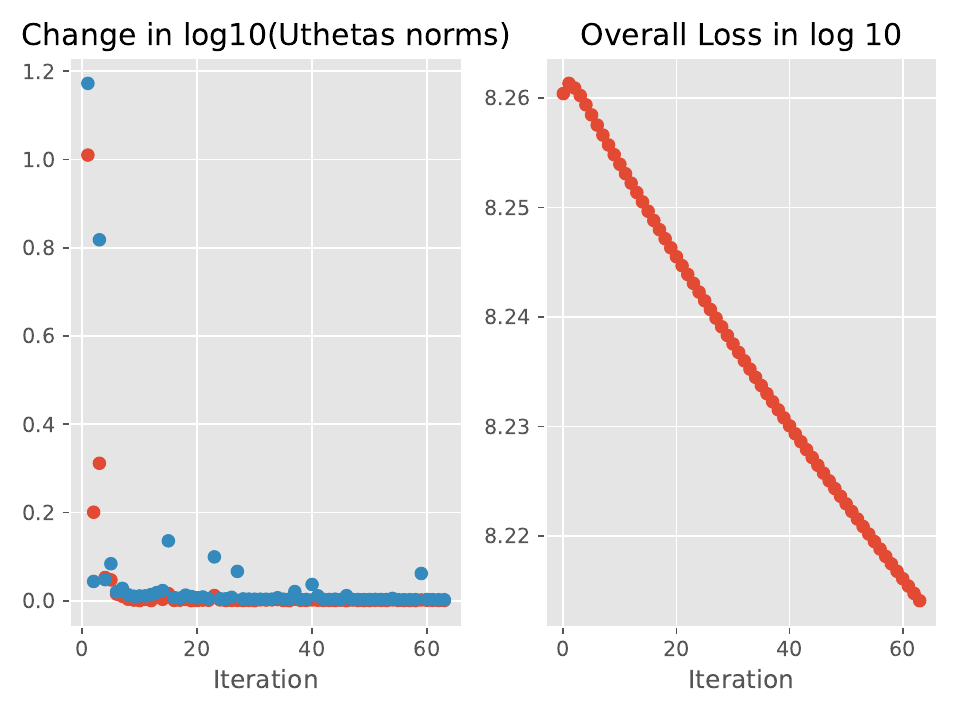}
\includegraphics[width=0.5\linewidth]{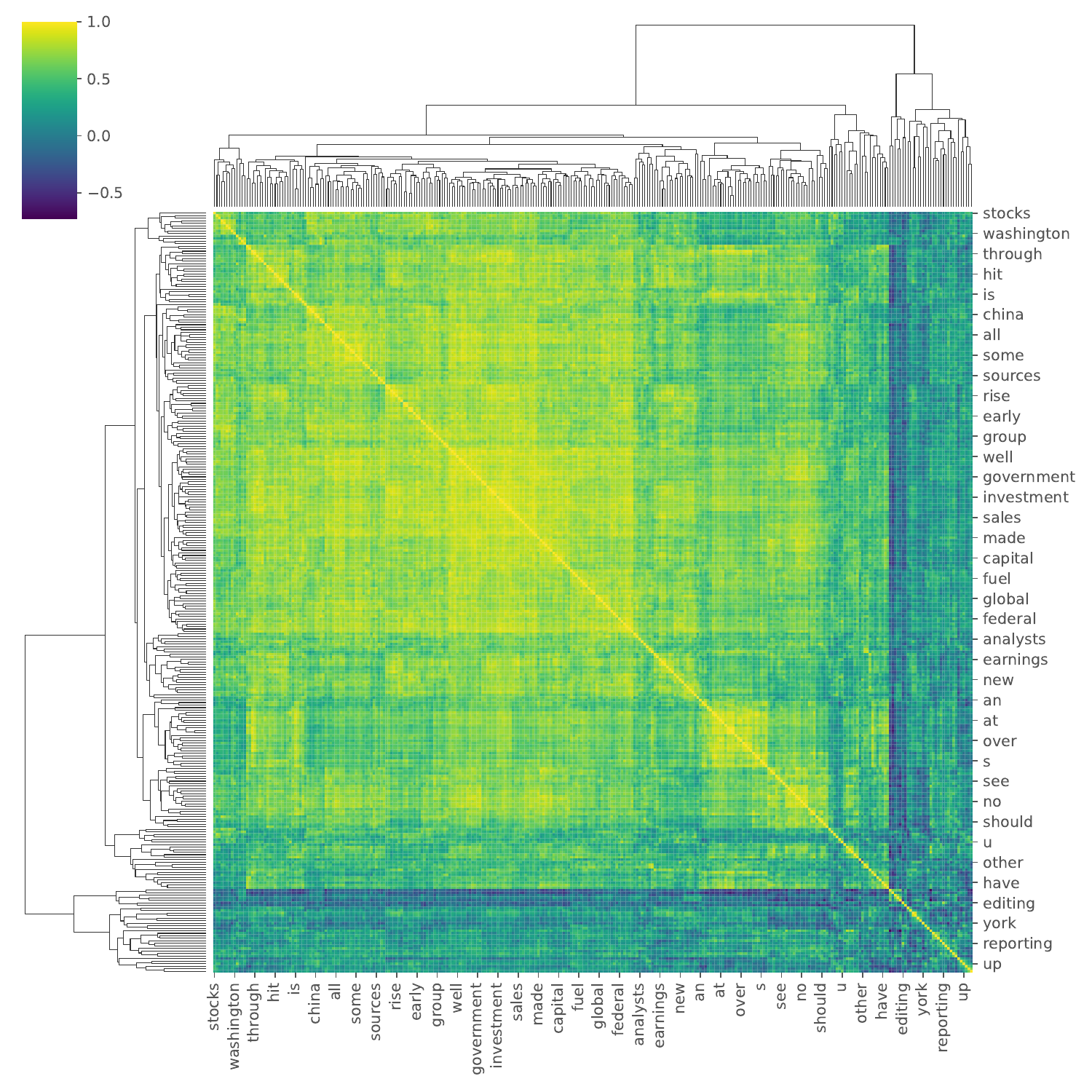}
\end{tabular}
\caption{Left: The absolute value of change in the log$_{10}$ norms of $\vU_{\vtheta}$ and $\vU_{\wtvtheta}$ from successive iterations and overall loss curve. Right: Cosine similarity of word vector representations shown as heatmap.}
\label{loss_and_cor_of_vectors}
\end{figure}

To visualize the relationships between different words, we first project the learned vector representations to first 10 principal components (PCs). These 10 PCs cumulatively explained 97\% of variations in the word vectors.  We then conduct further dimension reduction with T-Distributed Stochastic Neighbor Embedding (t-SNE) to 2-dimensional space. The t-SNE
minimizes the divergence between two distributions: a distribution that measures pairwise similarities of the input objects and a distribution that measures pairwise similarities of the corresponding low-dimensional points in the embedding. The words on t-SNE coordinates are shown in Figure \ref{tsne_on_pca}. Each point on the plot represents a word in the low dimensional space. The closer two words are on the plot, the more similar they are in meaning or usage.
 For example, the top five most similar words to `capital' from the 300 words  are `firm',	`management', `fund',	`board', 	`financial'. These are based on the cosine similarity. The five most similar words to `stocks' from the 300 words are 	`companies',	`bitcoin',	`banks',	`prices'  and	`some'. The top five most similar words to `finance' are	`technology',	`energy',	`trading',	`consumer',	`analysts'. Words `monday', `tuesday', `wednesday' are clustered together.
 This scatter plot captures some semantic similarities.

\begin{figure}
\includegraphics[width=\linewidth]{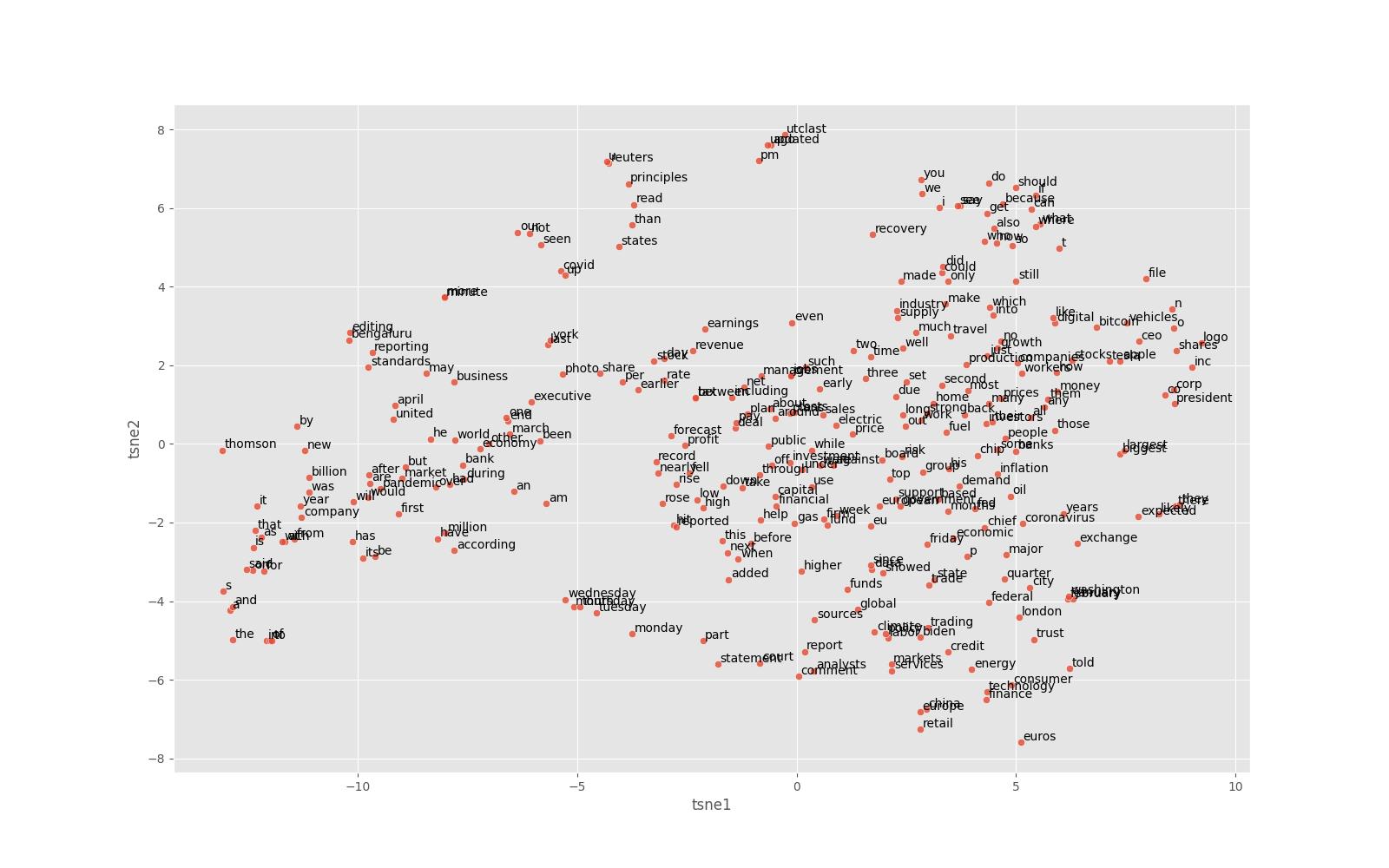}
\caption{Plot of word vector representations using t-SNE on first 10 principal components.}
\label{tsne_on_pca}
\end{figure}
}

\section{\modify Conclusion and discussion}
\label{section:conclusion}
{\modify

In summary, we presented the shared parameter alternating zero-inflated Gamma (SA-ZIG) regression model in this paper. The SA-ZIG model is designed for highly skewed non-negative matrix data. It uses a logit link to model the zero versus positive observations. For the Gamma part, we considered two link functions: the canonical link and the log link, and derived updating formulas for both.

We proposed an algorithm that alternately updates the parameters $\vtheta$ and $\wtvtheta$  while holding one of them fixed. The Fisher scoring algorithm, with or without learning rate adjustment, was employed in each step of the alternating update. Numerical studies indicate that learning rate adjustment is crucial in SA-ZIG regression. Without it, the algorithm may fail to find the optimal direction.

After model estimation, the matrix is factorized into the product of a left matrix and a right matrix. The rows of the left matrix and the columns of the right matrix provide vector representations for the rows and columns, respectively. These estimated row and column vector representations can then be used to assess the relevance of items and make recommendations in downstream analysis.

The SA-ZIG model is inherently similar to factor analysis. In factor analysis, both the loading matrix and the coefficient vector are unknown. The key difference between SA-ZIG and factor analysis is that SA-ZIG uses a large coefficient matrix, whereas factor analysis uses a single vector. Additionally, SA-ZIG assumes a two-stage Bernoulli-Gamma model, while factor analysis assumes a normal distribution.

In both models, likelihood-based estimation can determine convergence behavior by linking the complete data likelihood with the conditional likelihood. For factor analysis, the normal distribution and the linearity of the sufficient statistic in the observed data allow the use of ALS to estimate both the loading matrix and the coefficient vector (\cite{Dempster77}). However, SA-ZIG cannot use ALS because the variance of the Gamma distribution is not constant.

In both SA-ZIG and factor analysis, the unobserved row (or column) vector representation and the factor loading matrix can be treated as missing data, assuming the data is missing completely at random. For missing data analysis, the well-known Expectation-Maximization (EM) algorithm can be used to estimate parameters. The EM algorithm has the advantageous property that successive updates always move towards maximizing the log likelihood. It works well when the proportion of missing data is small, but it is notoriously slow when a large amount of data is missing.

For SA-ZIG, the alternating scheme with the Fisher scoring algorithm offers the benefit of a quadratic rate of convergence if the true parameters and their estimates lie within the interior of the parameter space. However, in real applications, the estimation process might diverge at either stage of the alternating scheme because the Fisher scoring update does not always guarantee an upward direction, especially in cases of complete or quasi-complete separation. Additionally, the algorithm may struggle to find the optimal solution due to the non-identifiability of the row and column matrices under orthogonal transformations, leading to a ridge of solutions. The learning rate adjustment in Algorithm \ref{algorithm:alternating-ZIG-regression} helps by making small moves during successive updates in later stage of the algorithm and thereby is more likely to find a solution.

Future research on similar problems could explore alternative distributions beyond Gamma. Tweedie and Weibull distributions, for instance, are capable of modeling both symmetric and skewed data through varying parameters, each with its own associated link functions. However, new algorithms and convergence analyses would need to be developed specifically for these distributions. In practical applications, the most suitable distribution for the observed data is often uncertain, making diagnostic procedures an important area for further investigation.

}

%% file: ZIG_main.bbl
\begin{thebibliography}{}

\bibitem[\protect\citeauthoryear{Albert and Anderson}{Albert and
  Anderson}{1984}]{MLE-existence-LogisticReg-Multinomial:1984}
Albert, A. and J.~A. Anderson (1984).
\newblock On the existence of maximum likelihood estimates in logistic
  regression models.
\newblock {\em Biometrika\/}~{\em 71\/}(1), 1--10.

\bibitem[\protect\citeauthoryear{Albert, Follmann, and Barnhart}{Albert
  et~al.}{1997}]{Albert-et-al:1997}
Albert, P.~S., D.~A. Follmann, and H.~X. Barnhart (1997).
\newblock A generalized estimating equation approach for modeling random length
  binary vector data.
\newblock {\em Biometrics\/}~{\em 53\/}(3), 1116--1124.

\bibitem[\protect\citeauthoryear{Boyd and Vandenberghe}{Boyd and
  Vandenberghe}{2004}]{BoydConvex:2004}
Boyd, S. and L.~Vandenberghe (2004).
\newblock {\em Convex optimization}.
\newblock Cambridge university press.

\bibitem[\protect\citeauthoryear{Dempster, Laird, and Rubin}{Dempster
  et~al.}{1977}]{Dempster77}
Dempster, A.~P., N.~M. Laird, and D.~B. Rubin (1977).
\newblock Maximum likelihood from incomplete data via the em algorithm.
\newblock {\em Journal of the Royal Statistical Society: Series B
  (Methodological)\/}~{\em 39\/}(1), 1--22.

\bibitem[\protect\citeauthoryear{Du}{Du}{2007}]{Du:2007}
Du, J. (2007).
\newblock Which estimator of the dispersion parameter for the gamma family
  generalized linear models is to be chosen?
\newblock {\em Master Thesis, Dalarna University.
  http://www.statistics.du.se/essays/D07C.Du.pdf\/}.

\bibitem[\protect\citeauthoryear{Edelman and Jeong}{Edelman and
  Jeong}{2022}]{ref6}
Edelman, A. and S.~Jeong (2022).
\newblock Fifty three matrix factorizations: A systematic approach.
\newblock {\em arXiv eprint math.NA 2104.08669,
  https://doi.org/10.48550/arXiv.2104.08669\/}.

\bibitem[\protect\citeauthoryear{Fern\'{a}ndez-Tob\'{i}as, Cantador, Tomeo,
  Anelli, and Di~Noia}{Fern\'{a}ndez-Tob\'{i}as
  et~al.}{2019}]{Fernandez-Tobias-et_al2019}
Fern\'{a}ndez-Tob\'{i}as, I., I.~Cantador, P.~Tomeo, V.~W. Anelli, and
  T.~Di~Noia (2019).
\newblock Addressing the user cold start with cross-domain collaborative
  filtering: exploiting item metadata in matrix factorization.
\newblock {\em User Modeling and User-Adapted Interaction\/}~{\em 29},
  443--486.

\bibitem[\protect\citeauthoryear{Gan, Liu, Li, and Zhang}{Gan
  et~al.}{2021}]{ref5}
Gan, J., T.~Liu, L.~Li, and J.~Zhang (2021, 07).
\newblock {Non-negative Matrix Factorization: A Survey}.
\newblock {\em The Computer Journal\/}~{\em 64\/}(7), 1080--1092.

\bibitem[\protect\citeauthoryear{Givens and Hoeting}{Givens and
  Hoeting}{2013}]{computational-statistics:2013}
Givens, G.~H. and J.~A. Hoeting (2013).
\newblock {\em Computational Statistics, Second Edition\/} (2 ed.).
\newblock John Wiley \& Sons, Inc.

\bibitem[\protect\citeauthoryear{Haberman}{Haberman}{1977}]{Haberman:1977}
Haberman, S. (1977).
\newblock {\em The Analysis of Frequency Data}.
\newblock Midway reprint. University of Chicago Press.

\bibitem[\protect\citeauthoryear{Have, Kunselman, Pulkstenis, and Landis}{Have
  et~al.}{1998}]{Ten-et-al:1998}
Have, T. R.~T., A.~R. Kunselman, E.~P. Pulkstenis, and J.~R. Landis (1998).
\newblock Mixed effects logistic regression models for longitudinal binary
  response data with informative drop-out.
\newblock {\em Biometrics\/}~{\em 54\/}(1), 367--383.

\bibitem[\protect\citeauthoryear{Koren}{Koren}{2008}]{Koren}
Koren, Y. (2008).
\newblock Factorization meets the neighborhood: a multifaceted collaborative
  filtering model.
\newblock In {\em Proceedings of the 14th ACM SIGKDD International Conference
  on Knowledge Discovery and Data Mining}, KDD '08, New York, NY, USA, pp.\
  426–434. Association for Computing Machinery.

\bibitem[\protect\citeauthoryear{Marschner}{Marschner}{2011}]{convergence-GLM:2011}
Marschner, I.~C. (2011).
\newblock {glm2}: Fitting generalized linear models with convergence problems.
\newblock {\em The R Journal\/}~{\em 3}, 12--15.

\bibitem[\protect\citeauthoryear{Mikolov, Chen, Corrado, and Dean}{Mikolov
  et~al.}{2013}]{mikolov2013A}
Mikolov, T., K.~Chen, G.~S. Corrado, and J.~Dean (2013).
\newblock Efficient estimation of word representations in vector space.
\newblock In {\em International Conference on Learning Representations}.

\bibitem[\protect\citeauthoryear{Mikolov, Sutskever, Chen, Corrado, and
  Dean}{Mikolov et~al.}{2013}]{mikolov2013B}
Mikolov, T., I.~Sutskever, K.~Chen, G.~S. Corrado, and J.~Dean (2013).
\newblock Distributed representations of words and phrases and their
  compositionality.
\newblock In C.~Burges, L.~Bottou, M.~Welling, Z.~Ghahramani, and K.~Weinberger
  (Eds.), {\em Advances in Neural Information Processing Systems}, Volume~26.
  Curran Associates, Inc.

\bibitem[\protect\citeauthoryear{Mills}{Mills}{2013}]{Mills:2013}
Mills, E.~D. (2013).
\newblock Adjusting for covariates in zero-inflated gamma and zero-inflated
  log-normal models for semicontinuous data.
\newblock {\em PhD Dissertation, University of Iowa. 10.17077/etd.7v3bafbd\/}.

\bibitem[\protect\citeauthoryear{Moulton, Curriero, and Barroso}{Moulton
  et~al.}{2002}]{Moulton-et-al:2002}
Moulton, L.~H., F.~C. Curriero, and P.~F. Barroso (2002).
\newblock Mixture models for quantitative hiv rna data.
\newblock {\em Statistical Methods in Medical Research\/}~{\em 11\/}(4),
  317--325.

\bibitem[\protect\citeauthoryear{Nguyen, Wang, and Kalousis}{Nguyen
  et~al.}{2016}]{Nguyen_et_al2016}
Nguyen, P., J.~Wang, and A.~Kalousis (2016).
\newblock Factorizing \protect{lambdaMART} for cold start recommendations.
\newblock {\em Machine Learning\/}~{\em 104}, 223--242.

\bibitem[\protect\citeauthoryear{Nobre, Carvalho, Griep, Fonseca, Melo, Santos,
  and Chor}{Nobre et~al.}{2017}]{Nobre-etal:2017}
Nobre, A.~A., M.~S. Carvalho, R.~H. Griep, M.~d. J. M.~d. Fonseca, E.~C.~P.
  Melo, I.~d.~S. Santos, and D.~Chor (2017, Jan.).
\newblock Multinomial model and zero-inflated gamma model to study time spent
  on leisure time physical activity: an example of elsa-brasil.
\newblock {\em Revista de Sa\'{u}de P\'{u}blica\/}~{\em 51}, 76.

\bibitem[\protect\citeauthoryear{Osborne}{Osborne}{1992}]{Osborne:1992}
Osborne, M.~R. (1992).
\newblock Fisher's method of scoring.
\newblock {\em International Statistical Review / Revue Internationale de
  Statistique\/}~{\em 60\/}(1), 99--117.

\bibitem[\protect\citeauthoryear{Panda and Ray}{Panda and
  Ray}{2022}]{Panda_Ray2022}
Panda, D.~K. and S.~Ray (2022).
\newblock Approaches and algorithms to mitigate cold start problems in
  recommender systems: A systematic literature review.
\newblock {\em Journal of Intelligent Information Systems\/}~{\em 59\/}(2),
  341--366.

\bibitem[\protect\citeauthoryear{Pennington, Socher, and Manning}{Pennington
  et~al.}{2014}]{pennington2014glove}
Pennington, J., R.~Socher, and C.~D. Manning (2014).
\newblock Glove: Global vectors for word representation.
\newblock In {\em Empirical Methods in Natural Language Processing (EMNLP)},
  pp.\  1532--1543.

\bibitem[\protect\citeauthoryear{Puthiya~Parambath and
  Chawla}{Puthiya~Parambath and Chawla}{2020}]{Puthiya_et_al2020}
Puthiya~Parambath, S.~A. and S.~Chawla (2020).
\newblock Simple and effective neural-free soft-cluster embeddings for item
  cold-start recommendations.
\newblock {\em Data Mining and Knowledge Discovery\/}~{\em 34}, 1560--1588.

\bibitem[\protect\citeauthoryear{Reddi, Kale, and Kumar}{Reddi
  et~al.}{2018}]{Reddi.etal}
Reddi, S.~J., S.~Kale, and S.~Kumar (2018).
\newblock On the convergence of adam and beyond.
\newblock In {\em International Conference on Learning Representations}.

\bibitem[\protect\citeauthoryear{Saberi-Movahed, Berahman, Sheikhpour, Li, and
  Pan}{Saberi-Movahed et~al.}{2024}]{saberimovahed2024}
Saberi-Movahed, F., K.~Berahman, R.~Sheikhpour, Y.~Li, and S.~Pan (2024).
\newblock Nonnegative matrix factorization in dimensionality reduction: A
  survey.
\newblock {\em arXiv:2405.03615, https://arxiv.org/abs/2405.03615\/}.

\bibitem[\protect\citeauthoryear{Shi, Li, Hong, and Sun}{Shi
  et~al.}{2021}]{Shi.etal}
Shi, N., D.~Li, M.~Hong, and R.~Sun (2021).
\newblock {RMS}prop converges with proper hyper-parameter.
\newblock In {\em International Conference on Learning Representations}.

\bibitem[\protect\citeauthoryear{Silvapulle}{Silvapulle}{1981}]{MLE-existence-LogisticReg-Binomial:1981}
Silvapulle, M.~J. (1981).
\newblock On the existence of maximum likelihood estimators for the binomial
  response models.
\newblock {\em Journal of the royal statistical society series
  b-methodological\/}~{\em 43}, 310--313.

\bibitem[\protect\citeauthoryear{Wang and Zhang}{Wang and
  Zhang}{2013}]{Wang_zhang2013}
Wang, Y.-X. and Y.-J. Zhang (2013).
\newblock Nonnegative matrix factorization: A comprehensive review.
\newblock {\em IEEE Transactions on Knowledge and Data Engineering\/}~{\em
  25\/}(6), 1336--1353.

\bibitem[\protect\citeauthoryear{Wei, Zhou, Grosmark, Ajabi, Sparks, Zhou,
  Brandon, Losonczy, and Paninski}{Wei et~al.}{2020}]{Wei-etal:2020}
Wei, X.-X., D.~Zhou, A.~D. Grosmark, Z.~Ajabi, F.~T. Sparks, P.~Zhou, M.~P.
  Brandon, A.~Losonczy, and L.~Paninski (2020).
\newblock A zero-inflated gamma model for post-deconvolved calcium imaging
  traces.
\newblock {\em Neural data science / analysis\/}~{\em 3\/}(2).

\bibitem[\protect\citeauthoryear{Wu and Carroll}{Wu and
  Carroll}{1988}]{Wu-Carroll:1988}
Wu, M.~C. and R.~J. Carroll (1988).
\newblock Estimation and comparison of changes in the presence of informative
  right censoring by modeling the censoring process.
\newblock {\em Biometrics\/}~{\em 44\/}(1), 175--188.

\bibitem[\protect\citeauthoryear{Zhang and Liu}{Zhang and
  Liu}{2015}]{Zhang_Liu2015}
Zhang, Z. and H.~Liu (2015).
\newblock Social recommendation model combining trust propagation and
  sequential behaviors.
\newblock {\em Applied Intelligence\/}~{\em 43}, 695--706.

\end{thebibliography}
